\documentclass{article}
\usepackage[utf8]{inputenc}
\usepackage[T1]{fontenc}
\usepackage{textcomp}
\usepackage{lmodern}
\usepackage{multirow}
\usepackage[margin={1 in}]{geometry} 
\usepackage{amsmath}
\usepackage{booktabs}
\usepackage{tikz}

\usepackage{graphicx, algorithm, algpseudocode}

\usepackage{subcaption}
\usepackage[normalem]{ulem} 
\usepackage{blindtext}
\usepackage{comment}
\usepackage{bbm}
\usepackage[authoryear]{natbib}
\usepackage{hyperref}
\usepackage{amsmath}
\usepackage{graphicx}
\usepackage{bbm}
\usepackage{enumitem}
\usepackage{amsfonts}
\usepackage{mathdots}

\usepackage{tcolorbox}
\usepackage{enumitem}
\usepackage{MnSymbol}
\usepackage{wasysym}
\usepackage{bm}
\usepackage{xurl}
\usepackage{float}
\usepackage{amsthm}
\newtheorem{lemma}{Lemma}
\newtheorem{theorem}{Theorem}
\newtheorem{remark}{Remark}
\newtheorem{definition}{Definition}

\title{Latent Noise Injection for Private and Statistically Aligned Synthetic Data Generation}
\author{Rex Shen \thanks{\texttt{rshen0@alumni.stanford.edu}} \\
  Stanford University \and Lu Tian \thanks{\texttt{lutian@stanford.edu}} \\
  Stanford University}
\date{June 19, 2025}

\begin{document}

\maketitle

\begin{abstract}

Synthetic Data Generation has become essential for scalable, privacy-preserving statistical analysis. While standard approaches based on generative models, such as Normalizing Flows, have been widely used, they often suffer from slow convergence in high-dimensional settings, frequently converging more slowly than the canonical $1/\sqrt{n}$ rate when approximating the true data distribution.

To overcome these limitations, we propose a Latent Noise Injection method using Masked Autoregressive Flows (MAF). Instead of directly sampling from the trained model, our method perturbs each data point in the latent space and maps it back to the data domain. This construction preserves a one to one correspondence between observed and synthetic data, enabling synthetic outputs that closely reflect the underlying distribution, particularly in challenging high-dimensional regimes where traditional sampling struggles.

Our procedure satisfies local $(\epsilon, \delta)$-differential privacy and introduces a single perturbation parameter to control the privacy-utility trade-off. Although estimators based on individual synthetic datasets may converge slowly, we show both theoretically and empirically that aggregating across $K$ studies in a meta analysis framework restores classical efficiency and yields consistent, reliable inference. We demonstrate that with a well-calibrated perturbation parameter, Latent Noise Injection achieves strong statistical alignment with the original data and robustness against membership inference attacks. These results position our method as a compelling alternative to conventional flow-based sampling for synthetic data sharing in decentralized and privacy-sensitive domains, such as biomedical research.

\end{abstract}

\section{Introduction}

Recently, the exponential growth of data has underscored the increasing need for synthetic data generation to replicate observed datasets. While extensive research has been conducted in this area, a fundamental challenge remains: generating high-quality synthetic samples that accurately preserve the statistical distributions and key properties of the original data. Ensuring the quality and fidelity of these synthetic samples helps address various challenges, such as protecting data confidentiality while enabling statistically valid analyses. In addition, synthetic data generation can reduce the need for extensive real-world experiments, thereby lowering costs and improving scalability.

In this work, we focus on Normalizing Flows as a powerful generative modeling approach for producing synthetic data. Among the variety of generative methods available, Normalizing Flows possess unique advantages. Unlike models such as Variational Autoencoders (VAEs) or Generative Adversarial Networks (GANs)~\citep{goodfellow2020generative, kingma2019introduction}, which rely on approximate reconstruction losses, Normalizing Flows leverage explicit likelihood-based training, enabling both density estimation and efficient sampling. More importantly, Normalizing Flows permit fast inversion and can transform complex data distributions into simple base distributions, typically standard Gaussian distributions, where adding random noise is often more natural. These properties make Normalizing Flows particularly well-suited for synthetic data generation, especially in multi-dimensional settings where capturing intricate dependencies is essential.

Among the family of Normalizing Flows, Masked Autoregressive Flows (MAFs) stand out for their balance of flexibility and tractability. MAF extends autoregressive models by stacking multiple transformations, each representing a learned, invertible map from latent variables to the data space. This increases model expressiveness while maintaining computational efficiency~\citep{papamakarios2017masked}. The architecture of MAF builds upon the Masked Autoencoder for Distribution Estimation (MADE)~\citep{germain2015made}, which enforces the autoregressive property by applying binary masks to a feedforward neural network. This property makes MAF—and Normalizing Flows more broadly—particularly well-suited for applications that require exact and tractable density evaluation, in contrast to VAEs and GANs, which lack closed-form likelihoods. Beyond generative modeling, density estimators like MAF have been successfully integrated into Bayesian workflows. They have been used as surrogates for posterior distributions~\citep{papamakarios2016fast}, as proposal distributions in importance sampling~\citep{paige2016inference}, and as key components in amortized inference frameworks for simulation-based models~\citep{greenberg2019automatic}. While MAF provides strong performance across a wide range of tasks, the broader family of Normalizing Flows includes several notable variants, each offering different trade-offs in expressiveness, computational efficiency, and applicability. For example, Continuous Normalizing Flows (CNFs)~\citep{chen2018neural, gao2024convergence, du2022flow} model data transformations as solutions to ordinary differential equations (ODEs), enabling highly flexible and adaptive mappings. However, their expressiveness comes at the cost of greater computational complexity due to the need to solve ODEs during training and sampling.

On the other hand, discrete-time flows such as RealNVP \citep{dinh2016density} and GLOW \citep{kingma2018glow} utilize coupling layers and invertible transformations to scale effectively to high-dimensional data, especially in image domains. The foundational NICE model \citep{dinh2014nice} introduced volume-preserving flows with additive coupling layers, paving the way for these architectures. In contrast to MAF, which is optimized for fast likelihood computation, Inverse Autoregressive Flows (IAF) \citep{kingma2016improved} are designed for fast sampling, making them particularly suitable for variational inference tasks. More expressive models such as Neural Spline Flows \citep{durkan2019neural} enhance flexibility by replacing affine transformations with monotonic rational-quadratic splines, allowing construction of richer yet still tractable mappings. Finally, Conditional Normalizing Flows \citep{ardizzone2019guided, winkler2019learning} extend standard flows by incorporating conditioning variables, supporting tasks such as class-conditional generation, structured prediction, and simulation-based inference. While we choose MAF as the basis for synthetic data generation in this paper due to the aforementioned considerations, we don't claim it is the best choice.  As the field continues to advance, hybrid architectures that combine the strengths of discrete-time flows (i.e., MAF) with continuous-time models (i.e., CNFs) represent a promising direction for developing new techniques for synthetic data generation.

To formalize a common setting in the literature, consider an observed dataset consisting of features and relevant outcomes from $n$ observations, denoted by $\mathbf{X} = \{X_1, \cdots, X_n\},$ where each $X_i \in \mathbb{R}^{d}$. The goal is to generate a synthetic dataset $\mathbf{\tilde{X}} = \{\tilde{X}_1, \cdots, \tilde{X}_m\}$, where $\tilde{X}_i \in \mathbb{R}^d$, such that
$$\tilde{X} \stackrel{d}{\sim} X$$ 
i.e., the synthetic data are drawn from a distribution similar to the underlying distribution of observed data. Importantly, the sample sizes $n$ and $m$ need not be equal.

The resulting synthetic dataset $\{\tilde{X}_1, \cdots, \tilde{X}_m\}$ can be used as a substitute for the original data in downstream statistical analyses without requiring access to original observations. This objective is closely aligned with the goals of Federated Learning (FL), where models are trained across distributed data sources without centralized data sharing on the individual level \citep{mcmahan2017communication}. However, FL typically involves iterative communication between local clients and a central server—transmitting gradients, parameter updates, or summary statistics across multiple rounds—which can impose substantial communication overhead, especially in large or bandwidth-constrained networks.

Moreover, FL workflows must be tailored to specific analyses. The summary statistics and updates required to fit a linear regression model, for example, differ from those of logistic regression. Consequently, FL generally requires the analytical objectives to be pre-specified, limiting its flexibility. In scenarios involving exploratory or sequential analyses—where subsequent steps depend on earlier results—FL becomes cumbersome and, in some cases, infeasible.

By contrast, our synthetic data strategy enables a one-shot solution: once the synthetic data is generated and shared, users can directly conduct statistical analyses without requiring repeated interactions or communication rounds. This approach offers significant flexibility and efficiency. However, it is important to acknowledge a key limitation: synthetic data, even when generated using state-of-the-art generative models, often fails to fully replicate results obtained from the original raw data. This shortfall arises from the intrinsic difficulty of nonparametrically estimating the joint distribution of a high-dimensional random vector with the typical sample sizes available in practice. A more subtle practical difficulty is that even if we can generate the synthetic data from a distribution that closely approximates that of $X,$ one may expect to reproduce a parameter estimate but certainly not the associated uncertainty measure such as the standard error due to the simple fact that the sample size of the synthetic data potentially can be arbitrarily big, which may yield artificially small standard errors of any sensible parameter estimates. This inflated confidence on parameter estimates from the synthetic data is due to the fact that users of the synthetic data neither account nor know how to account for the intrinsic randomness in estimating the generative model from observed data of finite sample size.

To mitigate this issue, it is desirable for synthetic data to resemble the observed data not only in distribution, but also in their actual values, i.e.,
$$\tilde{X} \approx X.$$
At the same time, $\tilde{X}$ must not be too close to $X$, as this would compromise privacy. Ideally, we aim to generate synthetic data that preserves analytical utility while satisfying formal privacy guarantees, such as local differential privacy (Local DP) \citep{ji2014differential, dwork2006calibrating, yang2024local}, which provides a quantifiable bound on privacy loss and supports controlled data sharing.

Accordingly, the main contribution of our work is a method for generating synthetic data that:
\begin{enumerate}
\item is sampled from a distribution closely matching that of the original data $X$;
\item remains sufficiently close to the observed data to reproduce results of statistical analysis based on original data;
\item introduces sufficient perturbation to satisfy the differential privacy guarantees.
\end{enumerate}

Specifically, we propose a procedure that diverges from traditional resampling methods by instead generating synthetic data through a latent-space transformation. Specifically, we map the observed data into a latent space, inject random perturbations, and then map the perturbed representations back to the data space. This approach induces a one to one correspondence between the original and synthetic datasets when $m = n$, ensuring that each synthetic observation reflects a controlled deviation from a specific real data point.

Another appealing feature of our method is its applicability to meta analysis for synthesizing evidence across multiple studies. In particular, traditional meta analysis does not require individual level data; instead, study-specific estimators are extracted and aggregated to form a combined estimate of the parameter of interest. However, similar to Federated Learning, meta analysis is only feasible when the relevant summary statistics in the form of parameter estimates, CIs, and related inferential quantities are available from all participating studies. These summary statistics are typically obtained either from published results or by request from data custodians. As a result, meta analysis cannot be conducted in an exploratory or flexible manner for arbitrary parameters without access to individual level data.

In practice, especially in fields such as healthcare and the social sciences, individual level data often cannot be shared due to privacy concerns~\citep{riley2010meta, shen2022data}. To address this challenge, our method facilitates the generation and sharing of privacy-preserving synthetic datasets from multiple studies, thereby enabling meta analysis for any parameter of interest. Crucially, our data generation strategy is designed to preserve the statistical integrity of each study, ensuring that downstream aggregation yields valid inference. Specifically, although standard Normalizing Flow models may converge slowly to the true distribution, our Latent Noise Injection method enhances fidelity at the study level. When synthetic datasets from multiple studies are combined, the resulting aggregate estimates exhibit greater stability than those derived from individual synthetic datasets.  For example, one may expect that its result (1) is robust to sharing synthetic data of poor quality from a few studies and (2) may average out the difference between synthetic data and original data across all studies of interest.  This makes our approach particularly advantageous for multi-site meta analyses.

Taken together, our method offers a practical middle ground—reducing the communication burden relative to Federated Learning while enabling access to statistically valid, privacy-preserving samples. This makes it especially suitable for decentralized scientific collaboration in sensitive domains, such as healthcare and finance. Moreover, future work may explore hybrid frameworks that integrate synthetic data generation with formal differential privacy guarantees or adaptive strategies that balance fidelity and privacy depending on the intended downstream analysis.
\begin{comment}
This framework aligns with growing interest in federated and decentralized inference, where preserving privacy while enabling robust statistical combination of results is paramount. Our work demonstrates that carefully structured synthetic data can play a central role in this setting—not only for privacy protection but also for inferential quality across studies.
\end{comment}

The structure of this paper is as follows. We begin by introducing Normalizing Flows, outlining their mathematical foundations and their role in generative modeling. We then describe our proposal for generating high-quality synthetic samples based on Normalizing Flows. We also present theoretical guarantees showing that our method can produce inference results equivalent to those based on the original data, while simultaneously providing privacy protection under the framework of Local DP~\citep{ji2014differential}. Next, we examine the operating characteristics of our method through extensive simulation studies and applications to real-world datasets. Finally, we conclude by discussing the broader implications of our findings and outline future research directions, including integrating transfer learning to further improve synthetic data quality, and extending our framework to more general meta analytic settings.

\section{Method}

In the following, we first provide a detailed overview of the Normalizing Flow framework and highlight the Masked Autoregressive Flow (MAF) architecture, which underpins our proposed methodology (Section \ref{sec:Norm_Flow}). Then, we introduce meta analysis (Section \ref{sec:Meta_Analysis}).  Lastly, we present our main contribution in proposing a new synthetic data generation method with its applications in meta analysis (Section \ref{sec:NewProposal}).

\subsection{Normalizing Flows}
\label{sec:Norm_Flow}

Normalizing Flows have emerged as a compelling framework for generative modeling due to their ability to combine expressive power with exact likelihood computation. Normalizing Flows construct complex data distributions from a simple base distribution by composing a series of invertible transformations.  The architecture of Normalizing Flows enables efficient training via maximum likelihood and tractable sampling, while preserving the distributional structure of the data.

In the context of privacy-preserving synthetic data generation, these properties are particularly advantageous. The invertibility ensures a one to one correspondence between a real observation and its ``projection'' into a latent space that follows a simple base distribution allowing controlled perturbations. Moreover, the explicit likelihood evaluation permits rigorous evaluation of distributional fidelity. Thus, Normalizing Flows provide an ideal foundation for generating high-fidelity synthetic data that balance statistical utility and privacy guarantees.

The general framework of the Normalizing Flow revolves around learning models in multi-dimensional continuous spaces. The objective of Normalizing Flows is to learn a bijective function $f: \mathbb{R}^d \rightarrow \mathbb{R}^d$ based on our observed data $\textbf{X} =\{X_1, \cdots, X_n \}$ consisting of $n$ independently identically distributed (i.i.d.) copies of a random vector $X\in \mathbb{R}^d$, such that $X = f(Z),$  where $Z$ follows a simple base distribution of the same dimension of $X$ (e.g., a standard multivariate Gaussian). The rational of training $f$ is as follows:  given a simple base distribution $p_Z$ on a latent variable space and a bijection $f(\cdot),$  the change of variables formula defines a distribution on $X=f(Z)$ by
$$p_X(X) = p_Z\left\{f^{-1}(X)\right\} \left| \det \left(\nabla f^{-1}(X)\right) \right|, $$ 
or equivalently
\begin{equation}\log\{p_X(X)\} = \log\left[p_Z\left\{f^{-1}(X)\right\}\right] + \log\left\{\left|\det\left\{\nabla f^{-1}(X)\right\} \right|\right\},  \label{eq:loglikNF}
\end{equation}
where $p_Z(\cdot)$ is the density function of $Z$. One may want to train $f$ to maximize the observed log-likelihood function 
$$l(f)=\sum_{i=1}^n \log\{p_X(X_i)\}.$$
However, directly computing the Jacobian in (\ref{eq:loglikNF}) for large $d$ is expensive and even intractable. % Second, it is restricted to bijective functions, which again appear impractical for modeling arbitrary distributions.
On the other hand, if we carefully design a parametrization of $f$ to make the computation of likelihood function simple and efficient, then the bijective function $f$ can be learned accordingly. Specifically, Normalizing Flows can be formed by composing \( k \) invertible ``simple'' transformations \( \{ f_j \}_{j=1}^{k} \) allowing a fast computation of the Jacobian and
\[
    X = f_k \circ f_{k-1} \circ \dots \circ f_1 (Z) = f(Z).
\]
Using the change-of-variables formula, the log-likelihood at observation $X$ can be computed as:
\begin{equation} \label{eq:Discrete_Flow}
     \log \{p_X(X)\} = \log[p_Z\{f^{-1}(X)\}] + \sum_{j=1}^{k} \log \left| \det \left[\nabla f_j^{-1} \{f_j^{-1}\circ\cdots \circ f_{k-1}^{-1}\circ f_k^{-1}(X)\}\right] \right|.   
\end{equation}

In the next subsection, we provide a detailed overview of Normalizing Flows with a particular focus on the Masked Autoregressive Flow (MAF) architecture, which is a special type of Normalizing Flow and permits fast computation of the log-likelihood function in (\ref{eq:Discrete_Flow}). The MAF forms the backbone of our approach.

\subsubsection{Masked Autoregressive Flow (MAF) Methodology}

MAF \citep{papamakarios2017masked} is a generative model designed to learn complex data distributions through a sequence of invertible transformations. It trains the Normalizing Flow by leveraging the autoregressive structure to parameterize a flexible transformation of a base probability distribution, typically a standard Gaussian distribution, to match a complex target distribution $p_X.$  

The MAF Model is constructed using multiple autoregressive transformations, implemented via the Masked Autoencoder for Distribution Estimation (MADE). The key components of the MAF model include:

\begin{itemize}
        \item \textbf{Base Distribution:} The latent variable typically follows an isotropic Gaussian, \( Z \sim N(0, \mathbf{I}_d) \), and transformations are applied to map it to a more complex target distribution, $P_X,$ where $\mathbf{I}_d$ is a $d$ by $d$ identity matrix. In a typical application, the target distribution $P_X$ is unknown, but one observes $n$ i.i.d. samples $X_1, X_2, \cdots, X_n$ from $P_X.$
        \item \textbf{MADE Layers:} The backbone of MAF is the MADE network, which is a masked feedforward map enforcing autoregressive dependencies. Specifically, suppose that $X=(x_1, \cdots, x_d)'$ and $Z=(z_1, \cdots, z_d)'$. The MADE network determines a one to one correspondence between $X$ and $Z$ as
        $$ z_i=\frac{x_i-\mu_i(x_1, \cdots, x_{i-1})}{\sigma_i(x_1, \cdots, x_{i-1})}, i=1, 2,  \cdots, d,$$
         where $\mu_i(\cdot)$ and $\sigma_i(\cdot)$ are individual neural networks with $(x_1, \cdots, x_{i-1})$ as input. 
    \item  \textbf{Permutation Layers:} Since MAF relies on autoregressive factorization, the ordering of variables influences the model's expressiveness. The Permutation Layer reorders inputs and re-trains a fresh MADE network. This step ensures training multiple invertible transformations to capture dependencies among individual components of $X$ in different orders. The final transport map is the composition of all resulting MADE networks.

\end{itemize}

\begin{figure}
    \centering
    \includegraphics[width=0.8\textwidth]{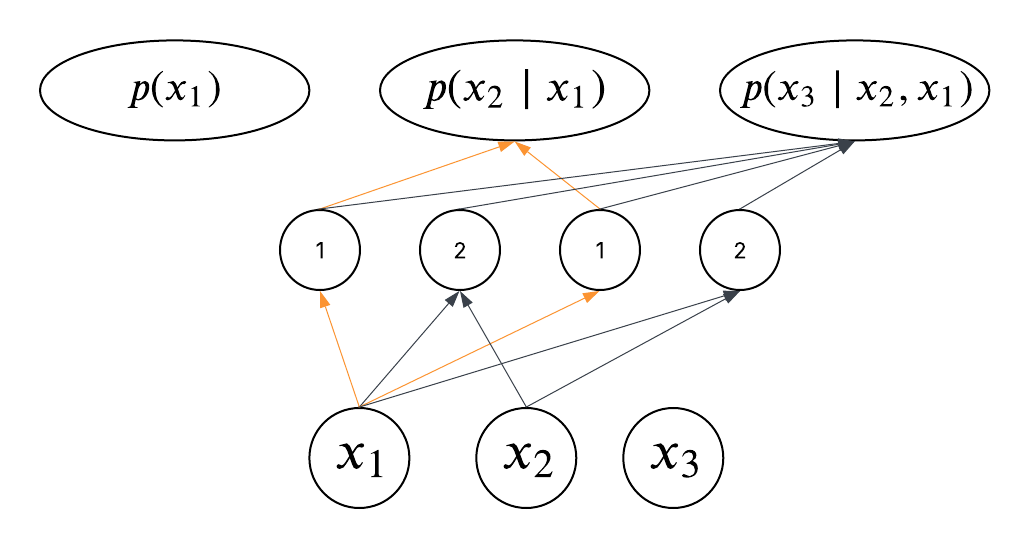}
    \caption{An autoregressive structure of a MADE layer corresponding to the factorization $p(x_1, x_2, x_3) = p(x_1) p(x_2|x_1) p(x_3|x_1, x_2)$.}
    \label{fig:Made_Diagram}
\end{figure}

 The key motivation in designing MAF is that the joint distribution of $X=(x_1,\cdots, x_d)'$ is determined by a set of conditional distributions:  $ x_i \mid x_1, \cdots, x_{i-1}, i=1, \cdots, d,$ as
 \[
    p_X(X) = \prod_{i=1}^{d} p(x_i \mid x_1, \cdots, x_{i-1}).
\]
 In MAF, these conditional distributions are modeled with Gaussian autoregressive working models:
\[
    x_i \mid  x_1, \cdots, x_{i-1} \sim  N\left\{\mu_i(x_1, \cdots, x_{i-1}), \sigma_i^2(x_1, \cdots, x_{i-1})\right\},
\]
where \( \mu_i (\cdot) \) and \( \sigma_i(\cdot)  \) are flexible neural networks. This transformation ensures that each output variable can be generated sequentially. Figure \ref{fig:Made_Diagram} illustrates the case with three variables $x_1$, $x_2$, and $x_3$, where $x_2$ depends only on $x_1$, and $x_3$ depends on both $x_2$ and $x_1$. The orange arrows represent the dependency paths relevant to modeling $p(x_2 \mid x_1)$, and black arrows represent the dependency paths relevant to modeling $p(x_3 \mid x_2, x_1).$ Note, each unit is assigned a number indicating which inputs it can depend on. 
This network ensures that the output for each variable \( x_i \) depends only on \( x_1, \cdots, x_{i-1} \), preserving the autoregressive structure. Moreover, unlike typical autoregressive models that require sequential updates, MAF benefits from MADE’s structure to compute the density at any data point in a single forward pass, making it highly suitable for GPU acceleration in training.

To see how MADE integrates into the Masked Autoregressive Flow (MAF), recall that Normalizing Flows transform a simple base distribution into a complex one using invertible mappings. In a simple MAF with a single MADE layer, the forward and inverse mappings are given by:
\begin{equation} \label{eq:Forward}
x_i = \mu_i(x_1,\cdots, x_{i-1}) + z_i \cdot \sigma_i(x_1, \cdots, x_{i-1}),
\end{equation}
and
\begin{equation}\label{eq:Inverse}
z_i = \frac{x_i - \mu_i(x_1, \cdots, x_{i-1})}{\sigma_i(x_1, \cdots, x_{i-1})}
\end{equation}
for $i = 1, \dots, d,$ respectively.  
Thus, the log-determinant of the Jacobian simplifies to
\[
    \log \left| \det \nabla f^{-1}(X) \right| = - \sum_{i=1}^{d} \log\{\sigma_i(x_1, \cdots, x_{i-1})\}.
\]

This formulation allows for efficient likelihood computation. Specifically, the log-likelihood function at $X$ can be evaluated as
$$\log\{p_X(X)\}=\log\{p_Z(Z)\}-\sum_{i=1}^{d} \log\{\sigma_i(x_1, \cdots, x_{i-1})\},$$
where $Z=f^{-1}(X)$ can be calculated via (\ref{eq:Inverse}).
Therefore, we can train $\{\mu_i(\cdot), \sigma_i(\cdot), i=1, \cdots, d\}$ to maximize the empirical log-likelihood:
\[
l(f) = \sum_{i=1}^{n} \log\{p_X(X_i)\},
\]
via the stochastic gradient descent algorithm. If there is more than one MADE layer, the log-likelihood function can still be efficiently calculated as 
$$\log\{p_X(X)\}=\log\{p_Z(Z)\}+\sum_{i=1}^k \log\left|\det\left(\nabla f_i^{-1}  \right)  \right|,$$
where $f=f_k\circ f_{k-1} \circ \cdots \circ f_1.$ In the following, we give a example of MAF with three MADE layers. With a given observation $X = (x_{1}, x_{2}, x_{3})',$ we proceed with the following steps to calculate $Z=f^{-1}(X)$ and the log-determinant of the Jacobian. 
\begin{enumerate}
    \item \textbf{First MAF Layer:}
    
    \begin{enumerate}
        \item We use an autoregressive transformation:
    \[
     (z_{11}, z_{12}, z_{13}) = \left(\frac{x_1 - \mu_{31}}{\sigma_{31}},  \frac{x_2 - \mu_{32}(x_1)}{\sigma_{32}(x_1)},  \frac{x_3 - \mu_{33}(x_1, x_2)}{\sigma_{33}(x_1, x_2)}\right).
    \]

    \item Output: $ Z_1=f_3^{-1}(x),$ where $Z_1=(z_{11}, z_{12}, z_{13})'.$
    \end{enumerate}

\item \textbf{Second MAF Layer:}

\begin{enumerate}
    
     \item Permutation: the variable order is shuffled. For instance,
    \[
     (1, 2, 3) \rightarrow (3, 1, 2).
    \]
    
    \item We use a new auto-regressive transformation on the permuted variables:
    \[
    (z_{23}, z_{21}, z_{22}) = \left(\frac{z_{13} - \mu_{23}}{\sigma_{23}},  \frac{z_{11} - \mu_{21}(z_{13})}{\sigma_{21}(z_{13})},  \frac{z_{12} - \mu_{22}(z_{13}, z_{11})}{\sigma_{22}(z_{13}, z_{11})}\right);
    \]

    \item Output: $ Z_2=f_2^{-1}(Z_1),$ where $Z_2=(z_{21}, z_{22}, z_{23})'.$
\end{enumerate}

\item \textbf{Third MAF Layer:}

\begin{enumerate}
     \item Permutation: the variable order is shuffled again. For instance,
    \[
     (1, 2, 3) \rightarrow (2, 3, 1).
    \]
    
    \item We use a new auto-regressive transformation on the permuted variables:
    \[
    (z_{32}, z_{33}, z_{31}) = \left(\frac{z_{22} - \mu_{12}}{\sigma_{12}},  \frac{z_{23} - \mu_{13}(z_{22})}{\sigma_{13}(z_{22})},  \frac{z_{21} - \mu_{11}(z_{22}, z_{23})}{\sigma_{11}(z_{22}, z_{23})}\right);
    \]

    \item Output: $ Z_3=f_1^{-1}(Z_2),$ where $Z_3=(z_{31}, z_{32}, z_{33})'.$
\end{enumerate}
    \item \textbf{The Final Output:}  
    $$Z=f^{-1}_1\circ f^{-1}_2\circ f^{-1}_3(X)=f^{-1}(X)=Z_3, $$
     where the transport map, $f=f_3\circ  f_2 \circ f_1,$ is characterized by $\mu_{ij}(\cdot)$ and $\sigma_{ij}(\cdot)$ with $ i=1,2,3$ being the index for the MAF layer and $j=1,2, 3$ being the index for the individual components of $X$ or $Z\in \mathbb{R}^d,$ and the log-determinant of the Jacobian is
    \begin{align*}
\log\left|\det\left( \nabla f^{-1}(X) \right)\right| 
&= -\left[\log(\sigma_{31}) + \log\left\{\sigma_{32}(x_1)\right\} + \log\left\{\sigma_{33}(x_1, x_2)\right\} \right] \\
&\quad -\left[\log(\sigma_{23}) + \log\left\{\sigma_{21}(z_{13})\right\} + \log\left\{\sigma_{22}(z_{13}, z_{11})\right\} \right] \\
&\quad -\left[\log(\sigma_{12}) + \log\left\{\sigma_{13}(z_{22})\right\} + \log\left\{\sigma_{11}(z_{22}, z_{23})\right\} \right].
\end{align*}

\begin{comment}
     \begin{align*} 
     \log\left|\det\{\nabla f^{-1}(X)\}\right|&-\left[\log(\sigma_{31})+\log\{\sigma_{32}(x_1)\}+\log(\sigma_{33}(x_1, x_2)  \right]\\
     &-\left[\log(\sigma_{23})+\log\{\sigma_{21}(z_{13})\}+\log\{\sigma_{22}(z_{13}, z_{11})\} \right]\\
     &-\left[\log(\sigma_{12})+\log\{\sigma_{13}(z_{22})\}+\log\{\sigma_{11}(z_{22}, z_{23})\} \right] 
     \end{align*}
\end{comment}

\end{enumerate}
Therefore, the log-likelihood function value at observation $X$ is 
\begin{align*}
&\log[p_Z\{f^{-1}(X)\}]+\log \left|\det\{\nabla f^{-1}(X)\}\right|\\
=& -3\log(\sqrt{2\pi})-\frac{z_{31}^2+z_{32}^2+z_{33}^2}{2}\\
&-\left[\log(\sigma_{31}) + \log\left\{\sigma_{32}(x_1)\right\} + \log\left\{\sigma_{33}(x_1, x_2)\right\} \right] \\
&-\left[\log(\sigma_{23}) + \log\left\{\sigma_{21}(z_{13})\right\} + \log\left\{\sigma_{22}(z_{13}, z_{11})\right\} \right] \\
&-\left[\log(\sigma_{12}) + \log\left\{\sigma_{13}(z_{22})\right\} + \log\left\{\sigma_{11}(z_{22}, z_{23})\right\} \right],
\end{align*}
which is very easy to compute.

So far, we have introduced MAF that applies a finite sequence of invertible maps to a random variable. However, there are other versions of Normalizing Flows as discussed in the Introduction. Specifically, there is a class of Continuous Normalizing Flows that realize the continuous limit as $k \rightarrow \infty$ in (\ref{eq:Discrete_Flow}).

\begin{remark}
Instead of a finite sequence of $k$ invertible maps from (\ref{eq:Discrete_Flow}), a Continuous Normalizing Flow (CNF) models an infinitesimal
transformation governed by an initial-value problem
\begin{equation}\label{eq:IV_Problem}
  \frac{dX_t}{dt}\;=\;v_{\theta}(t,X_t),\qquad 
  X_{0}=Z,\qquad t\in[0,1],
\end{equation}
where the \emph{velocity field} $v_{\theta}(t, x):\![0,1]\times \mathbb{R}^d\!\to\! \mathbb{R}^d$
is a neural network with parameters $\theta$ that can be estimated from the data (e.g.\ \citealp{chen2018neural, grathwohl2018ffjord}). The basic idea is to find an appropriate velocity field such that the terminal-time random variable approximates the target distribution $p_X,$ when the base distribution 
\[
  X_0=Z \;\sim\; N\bigl(0,\, \mathbf{I}_d\bigr).
\] 
The solution of this initial value problem generates a smoothing path between the base and target distribution. Combining the Ordinary Differential Equation with the instantaneous change-of-variables formula gives
\begin{equation} \label{eq:Continuous_Formula}
  \partial_t
  \begin{bmatrix}
      X_t\\
      \log p(X_t)
  \end{bmatrix}
  =
  \begin{bmatrix}
      v_{\theta}(t,X_t)\\
      -\operatorname{Tr}\left\{\nabla_x v_{\theta}(t,X_t)\right\}
  \end{bmatrix},                               
\end{equation}
where $Tr(\cdot)$ denotes the trace and $\nabla_x v_{\theta}(t,X_t)$ is the Jacobian Matrix of the velocity field with respect to $X_t$. Integrating (\ref{eq:IV_Problem}) and (\ref{eq:Continuous_Formula}) over time yields the exact log-density:
\begin{align}
%  X_1 &= X_0+\int_{0}^{1} v_{\theta}(t,X_t)\,dt,                     \\
  \log p_X(X) &=
    \log \{p_Z(Z)\} +\int_{0}^{1}\operatorname{Tr}\!\left\{\nabla_x v_{\theta}(t,X_t)\right\}\,dt.
\end{align}

Crucially, the sum of log-determinants in (\ref{eq:Discrete_Flow}) collapses to a single time-integral of a Jacobian Trace. There are several benefits of this approach. First, it drops the raw complexity of calculating the determinant from $O(d^3)$ in (\ref{eq:Discrete_Flow}) to $O(d^{2})$, or even $O(d)$ when a particular trace estimator (i.e., the Hutchinson Stochastic Trace Estimator) is employed \citep{grathwohl2018ffjord}. Second, it provides architectural freedom in that because (\ref{eq:Continuous_Formula}) involves only the trace of the Jacobian, any neural architecture can
    parameterize $v_{\theta}$ without sacrificing exact likelihoods, while every discrete layer must admit a tractable determinant. The corresponding analagous likelihood to be trained on is the one that maximizes the observed log-likelihood function
\[
  \sum_{i=1}^n \left[
    \log\{p_Z(X_{i0})\}+
    \int_{0}^{1}\operatorname{Tr}\!\left\{\nabla_x v_{\theta}(t,X_{it})\right\}\,dt
  \right],                                                
\]
whose gradients are efficiently obtained via the adjoint sensitivity method \citep{cao2002adjoint}. One advantage of considering CNF is the presence of theoretical guarantees on the convergence rate of the estimated transport map \citep{gao2024convergence}. 
\end{remark}

\subsection{Meta Analysis}
\label{sec:Meta_Analysis}

Having already introduced the methodology of Normalizing Flows in Section \ref{sec:Norm_Flow}, we now turn to meta analysis, which is a statistical methodology used to combine results from multiple independent studies or datasets. In particular, suppose that we have \( K \) datasets, denoted by \( \mathbf{X}_1, \dots, \mathbf{X}_K \), each providing a point estimator \( \hat{\theta}_k \) for an underlying parameter \( \theta_0 \) of interest. The goal of meta analysis is to synthesize these estimates into a more robust and generalizable consensus estimator \( \hat{\theta}_0 \).

In many real-world scenarios, especially those involving privacy-sensitive data (e.g., medical records or proprietary datasets), direct access to individual level data \( \mathbf{X}_k \) may not be feasible due to regulatory or ethical constraints. Fortunately, meta analysis can often still be conducted in such settings by leveraging summary statistics only. Specifically, if each site or study can share the pair \( (\hat{\theta}_k, \sigma_k^2) \)—that is, the point estimate and its associated variance—then classical statistical techniques such as fixed-effects or random-effects models (e.g., the DerSimonian-Laird Method by\citep{dersimonian1986meta}) can be employed to compute \( \hat{\theta}_0 \) in meta analysis.

There are two general classes of methods used for meta analysis: the fixed effects model and the random effects model. A fixed effects model assumes that all $\hat{\theta}_k$ estimates a common paramter $\theta_0$ in the sense that  
\begin{equation}
\hat{\theta}_k \sim N(\theta_0, \sigma_k^2), k=1, \cdots, K,  \label{eq:FixEffect}
\end{equation}
and we are interested in estimating $\theta_0$ based on $\{(\hat\theta_k, \sigma_k^2), k=1, \cdots, K\}.$  The corresponding estimator is the well-known inverse variance estimator
\begin{equation}
\hat{\theta}_F=\frac{\sum_{k=1}^K \sigma_k^{-2}\hat{\theta}_k }{\sum_{k=1}^K \sigma_k^{-2}}.
\label{eq:Est_Fixed}
\end{equation}
The variance of $\hat{\theta}_F$ is simply
$$ \sigma^2_F=\left(\sum_{k=1}^K \sigma_k^{-2}\right)^{-1}.$$ 
This estimator would be the optimal estimator with the smallest variance under (\ref{eq:FixEffect}). Oftentimes, the fixed effects model may be too restrictive and a natural extension allowing more between study heterogeneity is the random effects model, which assumes a hierarchical model 
$$\hat{\theta}_k \sim N(\theta_k, \sigma_k^2),$$
\begin{equation}\mbox{where}~~ \theta_k \sim N(\theta_0, \tau_0^2), k=1, \cdots, K.
\label{eq:RandomEffect}
\end{equation}
In other words, each estimator $\hat{\theta}_k$ estimates a study-specific population parameter $\theta_k,$ which can be different across studies. To facilitate the evidence synthesis, the random effects model assumes that all $\theta_k$'s are drawn from a common population $N(\theta_0, \tau_0^2)$ and the center of the population, $\theta_0,$ is the parameter of interest, which can be used to summarize information from all studies. $\tau_0^2$ characterizes the between study heterogeneity and is also a parameter of interest. The fixed effects model (\ref{eq:FixEffect}) is a special case of the random effects model (\ref{eq:RandomEffect}), as the random effects model degenerates into a fixed effects model when $\tau^2_0=0.$ To estimate $\theta_0$ in the random effects model (\ref{eq:RandomEffect}), one needs to notice that 
$$\hat{\theta}_k \sim N(\theta_0, \sigma_k^2+\tau_0^2), k=1, \cdots, K,$$
and an inverse variance estimator for $\theta_0$ would still be optimal.  However, the marginal variance of $\hat{\theta}_k$ is unknown, as it contains $\tau_0^2.$  Therefore, an estimator for $\theta_0$ can be constructed by first estimating $\tau_0^2$ via the following steps:  
\begin{enumerate}
\item obtain an initial estimator of $\theta_0$ as $\hat{\theta}_F$ given in (\ref{eq:Est_Fixed}).
\item estimate $\tau_0^2$ as 
    $$ \hat{\tau}^2=\max\left\{0,  \frac{\sum_{k=1}^K (\hat{\theta}_k-\hat{\theta}_F)^2/\sigma_k^2-(K-1)}{\sum_{k=1}^K \sigma_k^{-2}-(\sum_{k=1}^K \sigma_k^{-4})/(\sum_{k=1}^K \sigma_k^{-2})} \right\}. $$
\item update the initial estimator of $\theta_0$ as
\begin{equation}
\hat{\theta}_R=\frac{\sum_{k=1}^K (\sigma_k^{2}+\hat{\tau}^2)^{-1}\hat{\theta}_k }{\sum_{k=1}^K (\sigma_k^{2}+\hat{\tau}^2)^{-1}}.
\label{eq:Est_Random}
\end{equation}
\item approximate the variance of $\hat{\theta}_R$ by 
$$  \widehat{\sigma}^2_R=\left\{\sum_{k=1}^K (\sigma_k^{2}+\hat{\tau}^2)^{-1}\right\}^{-1},$$
based on which the Wald-Type confidence interval (CI) for $\theta_0$ can be constructed as 
 $$ [\hat{\theta}_R-1.96\hat{\sigma}_R, \hat{\theta}_R+1.96\hat{\sigma}_R].$$
\end{enumerate}
This estimator was originally proposed in \cite{dersimonian1986meta}. Since then, various estimators for $\theta_0$ have been proposed under the same random effects model framework, such as MLE, RMLE, Sidik-Jonkman Estimator, and Hunter-Schmidt Estimator, with different finite-sample performance \citep{harville1977maximum, inthout2014hartung, hunter2004methods, michael2019exact}. However, the DL Estimator \citep{dersimonian1986meta} remains to be one of the most popular choice in practice, and it is optimal when $K\rightarrow \infty.$ In summary, meta analysis can be conducted without access to individual level data, once relevant summary statistics from individual studies are available.  

However, challenges arise when the analyst has not yet decided on a specific parameter of interest and instead wishes to explore multiple candidate estimands. In such exploratory analyses—common in early-stage research or hypothesis generation—summary statistics are not always available. This brings the need for access to individual level data, which is typically required to flexibly define and estimate new parameters post hoc.

To reconcile this need for analytic flexibility with privacy protection, one potential solution is to share perturbed or synthetic data that mimics the statistical properties of the original datasets while preserving individual privacy. These synthetic datasets enable flexible exploratory analyses without compromising the confidentiality of the original individual records. We propose a novel method to create such perturbed datasets in Section \ref{sec:NewProposal} using the Normalizing Flows described in Section \ref{sec:Norm_Flow}.

\subsection{Our Contribution}
\label{sec:NewProposal}

% What is meta analysis. 

% Multiple datasets,  each dataset will give us $\hat{\theta}_k$ and we want combine them to generate a consensus estimator 
% $\hat{\theta}_0.$

% $\hat{\theta}_k$ is obtained from a dataset $\mathbf{X}_k, k=1, \cdots, K.$

% If $\mathbf{X}_k$ can not be shared, then the meta analysis still can be conducted if $(\hat{\theta}_k, \sigma_k^2), k=1, \cdots, K$ are shared.  However, if the user has not decided which parameter to estimate and want to explore an array of choices of possible parameters of interest in the context of meta analysis,  one would need access to individual level data. 

% For the purpose of privacy protection, we can not share the original data but only perturbed data. 

\subsubsection{Synthetic Data Generation}
\label{s:Single_Dataset}
In this subsection, we describe how to generate a synthetic dataset preserving the distribution of the observed data, while providing appropriate privacy protection via MAF.  Specifically,  given one observed dataset 
$$\mathbf{X} =\{X_1, X_2, \cdots, X_n\},$$ 
where $X_1, X_2, \cdots, X_n$ are i.i.d. copies of $X \sim P_X,$ which is an unknown distribution, the perturbation procedure is as follows. Let $f = f_k \circ f_{k - 1} \circ \dots \circ f_1$ be a composition of $K$ bijective and invertible functions $\{f_1, f_2, \cdots, f_k\}$ in a MAF, such that $X = f(Z),$ where $Z$ follows a standard Gaussian distribution, $N(0, \mathbf{I}_d),$ and $X \sim P_X.$ Then, perturbing an observed observation $X$ without changing its marginal distribution can be achieved via %\textcolor{orange}{Replace $f^{-1}(X)$ with X. $M(x)$ is an approriate random perturbation of ...} 
$$ 
f\left[\mathcal{M}\left\{ f^{-1}(X)\right\}  \right],$$ where $f^{-1}(\cdot)$ is the inverse function of $f(\cdot)$ and $\mathcal{M}(Z)$ is an appropriate random perturbation of a random variable $Z \sim N(0, \mathbf{I}_d).$ 

In practice, $f(\cdot)$ is unknown and needs to be estimated. This can be done with any type of Normalizing Flow. For this paper, we propose to use the MAF for efficient, higher quality training. There are several advantages of using MAF for synthetic data generation. First, the observations in generated synthetic data are expected to follow a distribution similar to $P_X.$  Second, unlike other synthetic data generation methods, observations in the perturbed dataset $\tilde{\mathbf{X}}$ have a one to one correspondence with data points in the original data, i.e, 
$$X_i \longleftrightarrow \tilde{X}_i,$$ 
where $\tilde{X}_i$ is the generated synthetic observation by perturbing $X_i,$ with a tuning parameter controlling the ``distance'' between $X_i$ and $\tilde{X}_i.$ Lastly, by transforming the original data into a latent space, where transformed observations approximately follow a standard Gaussian distribution, injecting independent Gaussian noise becomes a very natural way to introduce random perturbation. This perturbation approach accounts for the varying marginal distributions of different components of $X$ and their potential complex correlations. 
Therefore, we propose Algorithm \ref{ag:Perturb} (Latent Noise Injection) for synthetic data generation.
\begin{algorithm}[H]
\caption{Synthetic Data Generation (Latent Noise Injection)}
\label{ag:Perturb}
\begin{algorithmic}
\State train our flow on observed data $\mathbf{X}$, for which we denote trained functions as $\{\hat{f}_i\}_{i = 1}^K$. 
\For{ $i \in \{1, \cdots, n\}$  }
perturb sample $X_i\in \mathbf{X}$ as
\begin{equation}
\label{eq:Our_Procedure}
X_i \stackrel{\hat{f}^{-1}} \rightarrow f^{-1}(X_i) \rightarrow \sqrt{w} \hat{f}^{-1}(X_i) + \sqrt{1 - w} Z_i \stackrel{\hat{f}} \rightarrow \tilde{X}_i,
\end{equation}
where $\hat{f}=\hat{f}_k \circ \hat{f}_{k - 1} \dots \hat{f}_1,$ $\hat{f}^{-1}(\cdot)$ is the inverse function of $\hat{f}(\cdot)$ and $w \in (0, 1)$ is a tuning parameter controlling the privacy protection, and $Z_i \sim N(0, \mathbf{I}_d)$ is a Gaussian random variable independent of observed data $\mathbf{X}.$  
\EndFor
\State return the perturbed synthetic dataset 
$$\tilde{\mathbf{X}}=\{\tilde{X}_1, \cdots, \tilde{X}_n\}.$$
\end{algorithmic}
\end{algorithm}

  The proposed Latent Noise Injection method bridges different privacy-utility regimes via the perturbation parameter $w \in [0, 1]$. When $w = 1$, the generative distribution matches the empirical distribution of observed data. As $w \to 0$, however, the algorithm behaves like independent sampling from a distribution estimated by Normalizing Flows to approximate $P_X,$ as there is no direct one to one correspondence between original data and perturbed data. This resembles the classical generative sampling approach, where differences between analysis results based on synthetic and original data may remain substantial, as nonparametrically matching a target distribution can be challenging.

\subsubsection{Theoretical Properties of Perturbed Estimator \(\tilde{\theta}_w\) based on Synthetic Data }
\label{sec:Validity_Perturbed_Est_Single_Study}
Now, we analyze the asymptotic properties of estimators derived from generated synthetic data. To this end, we first investigate the convergence behavior of a general estimator \(\tilde{\theta}_{w}=\theta(\hat{P}_{w})\) obtained from our perturbed synthetic data, compared to its counterpart \(\hat{\theta}\) from observed data. Let 
$$\mathbf{X}=\{X_{1}, X_2, \cdots, X_{n}\} ~~\mbox{and}~~ \tilde{\mathbf{X}}=\{\tilde{X}_{1w}, \tilde{X}_{2w}, \cdots, \tilde{X}_{nw}\},$$ 
be observed and perturbed data, respectively, where $X_{i}, i=1, \cdots, n$ are $n$ i.i.d. copies of a $d$-dimensional random vector $X\sim P_X,$  $\tilde{X}_{iw}$ is perturbed $X_{i}$ with a transport map $\hat{f}$ and $w\in [0, 1]$. Let the empirical distribution of $\mathbf{X}$ and $\tilde{\mathbf{X}}$ be denoted by $\hat{P}$ and $\hat{P}_{w}$ respectively. Here, we assume that the transport map $\hat{f}$ is trained based on an independent dataset consisting of $O(n)$ i.i.d. observations following the distribution $P_X$ such that $\hat{f}^{-1}(X)$ approximately follows a standard Gaussian distribution $N(0, \mathbf{I}_d).$
In practice, $\hat{f}(\cdot)$ is oftentimes trained on $\mathbf{X}$ as well and induces a correlation between $\hat{f}(\cdot)$ and summary statistics of $\mathbf{X},$ which we ignore in our subsequent discussion.

In classical settings, an empirical parameter estimator converges to a population parameter $\theta$ with the root $n$ convergence rate, i.e., 
\[
\hat{\theta} - \theta = \theta(\hat{P})-\theta(P_X)=O_p\left(\frac{1}{\sqrt{n}}\right).
\]
One may hope that the estimator based on synthetic data, $\tilde{\theta}_{w}=\theta(\hat{P}_{w}),$ is close to $\hat{\theta},$ so that 
$$\tilde{\theta}_w-\theta\approx \hat{\theta}-\theta,$$
i.e., as an estimator of $\theta,$ $\tilde{\theta}_{w}$ is almost as good as $\hat{\theta}.$
Unfortunately, this is not true in general, as $\hat{P}_{w}$ may not be sufficiently close to either $P_X$ or its empirical counterpart $\hat{P}.$  Although we do not provide a formal theorem, existing theoretical results suggest that convergence rates for estimators under generative sampling without a linkage to original observations are typically slower than the classical \( O_p(n^{-1/2}) \) rate when the data dimension \( d > 2 \) \citep{tian2024enhancing, fournier2015rate}. In particular, under suitable regularity conditions, \citet{tian2024enhancing} show that the 1-Wasserstein Distance between $P_X$ and the distribution induced by the trained Normalizing Flow $P_{\hat{f}}=\hat{f}\left\{N(0, \mathbf{I}_d)\right\}$ is: 
\begin{equation}\label{eq:Non_Transfer_Distance}
E \left \{ \mathcal{W}_1(P_X, P_{\hat{f}}) \right \} = O\left(n^{-\frac{r}{2r + d}} \log^{5/2} n \right),  
\end{equation}
where the density function of $P_X$ has a continuous \(r\)th derivative.  Noting that $r/(2r+d)<1/2$ regardless of ``smoothness'' of $P_X,$ this convergence rate in Wasserstein Distance is slower than the regular root $n$ rate with an approximation error that may be propagated into the estimation of $\theta$, causing $\theta(P_X)-\theta(P_{\hat{f}})$ to converge at a rate slower than root $n$ as well.

\begin{remark}
Note, an intrinsic estimation error term \( \epsilon^z \) arises as an additional term in (\ref{eq:Non_Transfer_Distance}), but was omitted. It captures the excess risk in estimating the synthetic data distribution when the latent input \( Z \sim N(0, \mathbf{I}_d) \) is passed through the learned transport map \( \hat{f} \). We assume that $\epsilon^z$ is negligible compared to the overall generation error of the flow-based model because Lemma 39 of \citet{tian2024enhancing} suggests that $\epsilon^z = O_p(n^{-r_z/(2r_z + d)} \log^{5/2} n),$ where the base density function of $Z$ has continuous $r_z$th derivative. Since our base distribution is $N(0, \mathbf{I}_d),$ which is infinitely smooth, \( \epsilon^z \) is negligible.
\end{remark}

Noting that $\theta(\hat{P}_{0})$ with $w=0$ is a finite sample approximation to $\theta(P_{\hat{f}})$ with a precision of $O(n^{-1/2}),$ 
$$\tilde{\theta}_{w}-\theta=\left\{\theta(\hat{P}_{w})-\theta(P_{\hat{f}})\right\}+\left\{\theta(P_{\hat{f}})-\theta(P_X)\right\}$$ 
also cannot converge at the regular root $n$ rate for $w=0$.  Indeed, our simulation results suggest that for many $w$ away from 1,
\[
\tilde{\theta}_w - \theta \neq O_p\left(\frac{1}{\sqrt{n}}\right),
\]
and instead, the convergence is slower. Therefore, in general, one may not be able to replace $\hat{\theta}$ from the original observed data by $\tilde{\theta}_{w}$ from the synthetic data without losing efficiency. We also note that when $w\rightarrow 1,$  $\tilde{\theta}_{w}$ also approaches $\hat{\theta}.$ However, if $w$ is too close to 1, then the privacy protection would be lost as the injected noise level would be too small. We hope that when $w$ is sufficiently but not too close to 1, a good transport map trained via Normalizing Flows can still guarantee that the difference between $\tilde{\theta}_{w}$ and $\hat{\theta}$ is small, such that we may replace the original data with the synthetic data without losing asymptotic efficiency for the purpose of estimating $\theta$. The following Theorem \ref{thm:Meta_Analysis_Rate} ensures that this is possible.

\begin{theorem} \label{thm:Meta_Analysis_Rate}
Let \( \hat{\theta}=\theta(\hat{P}) \) be the estimator computed from the empirical distribution \( \hat{P} \) for a study, where $\hat{P}$ is the probability measure induced by $n$ i.i.d. observations in observed data $\mathbf{X}=\left\{X_1, \cdots, X_n\right\}.$  Similarly, \( \tilde{\theta}_w=\theta(\hat{P}_{w}) \) is the estimator computed from the synthetic distribution \( \hat{P}_{w} \) generated by our mechanism with perturbation weight \( w \).  Here, $\theta(\cdot)$ is a functional with the first order expansion
$$ \theta(\hat{\mathcal{P}})=\theta(\mathcal{P})+\frac{1}{n}\sum_{i=1}^n l(\mathcal{X}_i)+o_p\left(n^{-1}\right),$$
where $\hat{\mathcal{P}}$ is the probability measure induced by the empirical distribution of $n$ i.i.d. observations $\mathcal{X}_1, \cdots, \mathcal{X}_n \sim \mathcal{P}$. We assume the following:
\begin{enumerate}
\item $P_X$ has a continuously differentiable density function over a compact support on $\mathbb{R}^d.$
\item invertible transport map trained from the MAF, $\hat{f}(\cdot),$ converges to a deterministic invertible transport map $f(\cdot)$ in probability, where $Z=f^{-1}(X)\sim N(0, \mathbf{I}_d).$ Specifically, $\|\hat{f}-f\|_2=O_p(\delta_{1n})$ for $\delta_{1n}=o(1).$
\item $\hat{f}(\cdot)$  and $f(\cdot)$ are uniformly bi-Lipchitz Continuous on the support of $X.$  
\item Both $f$ and $\hat{f}$ are members of a class of functions $\mathcal{F}=\{f_\theta\},$ and the class of functions
$${\cal L}=\left\{l\left[\tilde{f}\left(\sqrt{w}\tilde{f}^{-1}(x)+\sqrt{1-w}z\right)\right]\mid w \in [0, 1], \|\tilde{f}-f\|_2\le \epsilon_n, \tilde{f}(\cdot)\in {\cal F}: \mathbb{R}^d \rightarrow \mathbb{R}^d \right\}$$ 
is a Donsker Class.
\item $l(\cdot)$ is continuously differentiable on $\mathbb{R}^d.$ 
\item $1-w=o(\delta_{2n})$ for $\delta_{2n}=o(1).$
\end{enumerate}
Then, we have  
\[
\theta(\hat{P}_{w}) - \theta(\hat{P}) = O_p\left(\delta_{1n}^2+\delta_{1n}\sqrt{\delta_{2n}}\right)+o_p(n^{-1/2}).
\]
\end{theorem}

Theorem \ref{thm:Meta_Analysis_Rate} provides a rate for the distance between perturbed estimator \( \tilde{\theta}_{w} \) and the non-private estimator \( \hat{\theta} \) as $n \rightarrow \infty$ and $w\rightarrow 1.$  More importantly, this convergence rate can be faster than the regular rate $O_p(n^{-1/2})$ under appropriate conditions. In such a case, both $\tilde{\theta}_{w}$ and $\hat{\theta}$ estimate $\theta$ with the same asymptotic precision. Specifically, if we let $w=1-O(n^{-2/d})$ and $\|\hat{f}-f\|_2=O_p(n^{-1/2+\delta_0}),$ where $\delta_0<\min(1/4, 1/d),$ then  
$$ \sqrt{n}(\tilde{\theta}_{w}-\hat{\theta})=o_p(1) \Longrightarrow \sqrt{n}(\tilde{\theta}_{w}-\theta)\sim N\left\{0, \sigma^2(P_X)\right\}$$
in distribution, as $n \rightarrow \infty.$ On the other hand, this convergence rate can be substantially slower than the regular rate, if either $|w-1|$ or $\|\hat{f}-f\|_2$ goes to zero too slowly. In Section \ref{sec:Validity_Meta_Analysis}, we have performed a numerical study to empirically confirm our observations on this convergence rate. 

Note that although in theory $w$ is a user-specified tuning parameter that can be arbitrarily close to 1, in practice, we don't want to use a $w$ too close to 1. Otherwise, the privacy protection will be lost due to the simple fact that the synthetic dataset $\tilde{\mathbf{X}}$ becomes the observed dataset $\mathbf{X}$ as $w\rightarrow 1.$  In Section \ref{sec:Privacy}, we argue that $w=1-O(n^{-2/d})$ is both small enough to provide sufficient privacy protection and big enough to ensure the asymptotic equivalence between $\tilde{\theta}_{w}$ and $\hat{\theta}.$ 

Now, we discuss the possibility of constructing a transport map $\hat{f}$ such that $\|\hat{f}-f\|_2=O_p(n^{-1/2+\delta_0})$ for a sufficiently small constant $\delta_0>0.$ Theorem 5.12 of \citet{gao2024convergence} proves that under regularity conditions for Continuous Normalizing Flows with time component $t \in [0, 1]$ as in (\ref{eq:IV_Problem}), the estimated velocity field $v_{\hat{\theta}}(t, X_t)$ converges to the true velocity field $v_{\theta}(t, x) \in W^{1, \infty}$ in (\ref{eq:Continuous_Formula}), where $W^{1, \infty}$ is the Sobolev Space of Lipschitz Functions whose first weak derivatives are bounded in $L^{\infty},$ satisfying
\begin{equation} \|v_{\hat{\theta}}(t, x)-v_{\theta}(t, x)\|_2 = O\left(n^{-1/(d + 3)}\right). \label{eq:Gao} \end{equation}
 The estimated transport map is determined by the velocity field and can be approximated as
  \begin{align*}
\hat{f}(z) & =  \hat{x}_{J_n-1}+\delta_n v_{\hat{\theta}}(t_{J_n-1}, \hat{x}_{J_n-1})\\  
\hat{x}_{J_n-1} & =  \hat{x}_{J_n-2}+\delta_n v_{\hat{\theta}}(t_{J_n-2}, \hat{x}_{J_n-2})\\
\cdots &\\
\hat{x}_2 & =  x_1+\delta_n v_{\hat{\theta}}(t_1, \hat{x}_1)\\
\hat{x}_1 & =  z+\delta_n v_{\hat{\theta}}(t_0, z) \\
\hat{x}_0 &= z,
\end{align*}
where $0=t_0, t_1, \cdots, t_{J_n-1}, t_{J_n}=1$ are $J_n+1$ equal-spaced points in the unit interval $[0, 1]$ and $\delta_n=J_n^{-1}$ is the step size dependent on sample size $n$. The limit transport map $f$ can be defined similarly based on $v_{\theta}$ with $J_n \rightarrow \infty$.  If $v_{\hat{\theta}}(t, x)$ and $v_{\theta}(t, x)$ are Lipschitz Continuous with a uniform bound, (\ref{eq:Gao}) also suggests that 
$$\|\hat{f}-f\|_2=O\left(n^{-1/(d + 3)}\right).$$
 We provide a more generic result by adopting Theorem 5.12's proof in \citet{gao2024convergence} to generalize the result of (\ref{eq:Gao}) for $v_{\theta} \in W^{s, \infty}$ with arbitrary smoothness $s$. Specifically, by assuming more smoothness on $v_{\theta}$, we can establish a new convergence rate
$$\|\hat{f}-f\|_2=o(n^{-s/(2s + d + 1)+\delta_0}),$$
for any $\delta_0>0.$ See Appendix \ref{app:TransportMap} for the proof sketch. This provides a theoretical justification for achieving a sufficiently fast convergence rate of the estimated transport map $\hat{f}(\cdot)$ to a underlying true optimal transporter $f$, if $f$ is sufficiently smooth. Note that this convergence rate of the transport map estimation is consistent with the convergence rate in Wasserstein Distance between the induced distribution given in (\ref{eq:Non_Transfer_Distance}).

The proof of Theorem \ref{thm:Meta_Analysis_Rate} is provided in Appendix \ref{app:EquivalenceProof}. The rational for the convergence can be explained by a simple example, where $\theta(P_X)=E(X).$  In such a case,
\begin{align*}
\sqrt{n}\left\{\theta(\hat{P})-\theta(P_X)\right\}&=\frac{1}{\sqrt{n}}\sum_{i=1}^{n} X_{i},\\
\sqrt{n}\left\{\theta(\hat{P}_{w})-\theta(P_{w})\right\}&=\frac{1}{\sqrt{n}}\sum_{i=1}^{n} \hat{f}\left(\sqrt{w}\hat{f}^{-1}(X_{i})+\sqrt{1-w}Z_i\right).
 \end{align*}
It follows from the stochastic equi-continuity guaranteed by the assumption on Donsker Class that 
 $$\sqrt{n}\left\{\theta(\hat{P})-\theta(\hat{P}_{w})\right\}=\sqrt{n}\left\{\theta(P_X)-\theta(P_{w})\right\} + o_p(1),$$
for $1-w=o(1).$ Note that if $w=1,$ then 
$$\theta(P_{w})=E\left\{ \hat{f}\left(\hat{f}^{-1}(X)\right) \right\}=E(X)=\theta(P_X).$$
Also, if $\hat{f}=f,$
$$\theta(P_{w})=E\left\{f\left(\sqrt{w}f^{-1}(X)+\sqrt{1-w}Z\right) \right\}=E\left\{ f\left(Z\right) \right\}=E(X)=\theta(P_X).$$
Thus, one may expect that $|\theta(P_X)-\theta(P_{w})|$ is small, when either $|w-1|$  or $\|\hat{f}-f\|_2$ is close to zero. Indeed, we may bound this difference by
$$C\left(\|\hat{f}-f\|_2^2+\sqrt{1-w}\|\hat{f}-f\|_2\right), $$
where $C$ is a constant multiplier. The main conclusion of Theorem \ref{thm:Meta_Analysis_Rate} follows.

We now turn to a formal privacy analysis of the perturbation mechanism applied within a single dataset. This begins with a proof of its compliance with Local Differential Privacy guarantees.

\subsubsection{Privacy Protection of Our Perturbation for a Single Dataset} \label{sec:Privacy}

\label{s:Privacy_Protection}

Local Differential Privacy (Local DP) has become a foundational framework for formalizing privacy guarantees in data analysis. At its core, Local DP ensures that the presence or absence of a single individual in the dataset has a limited effect on the output of a randomized mechanism. While the standard Local $(\epsilon, \delta)$-DP Formulation is widely used \citep{dwork2014algorithmic}, it can be challenging to analyze under composition or continuous settings. To address these limitations, Local Rényi Differential Privacy (Local RDP) was introduced as a generalization of Local DP using Rényi Divergence, offering a more structured and flexible privacy accounting mechanism \citep{mironov2017renyi}. In the following, we present formal definitions of both Local $(\epsilon, \delta)$-DP and RDP, followed by a key result (Theorem \ref{thm:Local_DP}) showing how privacy guarantees are preserved under the perturbation mechanism (\ref{eq:Our_Procedure}). %This theorem will later underpin our analysis of the trade-off between privacy and utility as a function of the perturbation weight $w$. 
First, we introduce the concept of Local DP.

\begin{definition} \label{def:Local_DP}
Local Differential Privacy (Local DP). A randomized mechanism $\mathcal{M}$ satisfies $(\epsilon, \delta)$- Local DP if for each individual $z$ and any possible values $z'$ and for any output $y$ in the output space of $\mathcal{M}$:
\[
\mathbb{P}[\mathcal{M}(z) = y] \leq e^{\epsilon} \mathbb{P}[\mathcal{M}(z') =y ]+\delta,
\]
where the probability is taken over the randomness of the mechanism $\mathcal{M}$ and $\epsilon$ and $\delta$ are two positive constants that can be arbitrarily close to zero. 
\end{definition}

%We also introduce Renyi Differential Privacy, which will come useful later on in our proofs.

%\textbf{Definition 1. Differential Privacy (DP)}. A randomized mechanism $\mathcal{M}$ satisfies $(\epsilon, \delta)$-Differential Privacy if for all measurable subsets $S$ of the output space of $\mathcal{M}$ and for all neighboring datasets $D, D' \in \mathcal{D}$ differing in at most one individual observation:
%\[
%\mathbb{P}[\mathcal{M}(D) \in S] \leq e^{\epsilon} \mathbb{P}[\mathcal{M}(D') \in S] + \delta,
%\]
%where the probability is taken over the randomness of the mechanism $\mathcal{M}$. We also introduce Renyi Differential Privacy, which will come useful later on in our proofs.

\begin{definition} \label{def:Renyi_Divergence}
Renyi Divergence. The Rényi Divergence of Order $\alpha > 1$ between two probability distributions $P$ and $Q$ over a domain $\mathcal{X}$ is defined as:
\begin{equation}
    D_\alpha(P \| Q) = \frac{1}{\alpha - 1} \log E_{x \sim Q} \left[ \left(\frac{P(x)}{Q(x)} \right)^\alpha \right].
\end{equation}
For $\alpha \rightarrow 1$, the Rényi Divergence is equivalent to the Kullback-Leibler (KL) Divergence:
\begin{equation}
    D_{\alpha \rightarrow 1}(P \| Q) = E_{x \sim P} \left[ \log \frac{P(x)}{Q(x)} \right].
\end{equation}
This can be seen by L'Hopital.
\end{definition}

\begin{definition} \label{def:Local_Renyi}
    Local Renyi Differential Privacy.  \textit{A randomized mechanism \(\mathcal{M}: \mathcal{X} \to \mathbb{R}^{d}\) satisfies \((\alpha, \epsilon)\)-Local Rényi Differential Privacy (Local RDP) if for all inputs \(y, z \in \mathcal{X}\) satisfying \(||y - z||_2 \leq b\) for some constant b,}
\begin{equation}
    D_\alpha(\mathcal{M}(y) \| \mathcal{M}(z)) \leq \epsilon,
\end{equation}
where \(D_\alpha(\cdot \| \cdot)\) denotes the Rényi Divergence of Order \(\alpha > 1\).
\end{definition}

\begin{remark}
Local Rényi Differential Privacy (Local RDP) provides a structured and flexible framework for analyzing privacy guarantees at the level of individual data entries, in comparison to standard $(\epsilon, \delta)$-Local Differential Privacy (Local DP) because of the following advantages:
\begin{itemize}
    \item \textbf{Tighter Composition Bounds:} Local RDP provides a control over cumulative privacy loss when composing multiple locally private mechanisms. Unlike $(\epsilon, \delta)$-Local DP, where composition of multiple random mechanisms requires complex summation or amplification arguments, Local RDP obeys a simple additive property:
    $$ {\cal M}_j(\cdot) ~\mbox{ is }~ (\alpha, \epsilon_j)\mbox{-Local RDP}, j=1,2 \Rightarrow 
    {\cal M}_1\left\{{\cal M}_2(\cdot)\right\} ~\mbox{ is }~ (\alpha, \epsilon_1 + \epsilon_2)\text{-Local RDP}.$$
%    \begin{equation}
%        (\alpha, \epsilon_1)\text{-Local RDP} + (\alpha, \epsilon_2)\text{-Local RDP} = (\alpha, \epsilon_1 + \epsilon_2)\text{-Local RDP}.
 %   \end{equation}
    %This significantly simplifies privacy accounting under multiple queries.

    \item \textbf{Conversion to $(\epsilon, \delta)$-Local DP:}  
    Proposition 3 of \citep{mironov2017renyi} shows that any $(\alpha, \epsilon)$ Local RDP guarantee can be converted into an $(\epsilon', \delta)$-Local DP guarantee, where
    \begin{equation}
        \epsilon' = \epsilon + \frac{\log(1/\delta)}{\alpha - 1}.
    \end{equation}
    This allows for finer-grained tracking of privacy loss while still being able to express results in the standard $(\epsilon, \delta)$-LDP form.

    \item \textbf{Stronger Privacy Guarantees:}  
    Local RDP offers stronger tail bounds on privacy loss compared to $(\epsilon, \delta)$-Local DP. While $(\epsilon, \delta)$-Local DP allows for a small probability $\delta$ of an unbounded privacy breach, Local RDP ensures that extreme deviations are tightly controlled, thereby preventing catastrophic privacy failure, however small the probability is.

    \item \textbf{Compatibility with Continuous Domains:}  
    Local RDP is naturally defined over probability distributions rather than discrete databases. This makes it particularly well-suited for continuous data settings, where inputs may differ smoothly or infinitesimally, which is common in high-dimensional generative modeling.

    \item \textbf{Intuitive Interpretation:}  
    The core intuition behind Local RDP is that it measures how distinguishable the outputs of a mechanism are on neighboring individual inputs, using the Rényi Divergence rather than the standard likelihood ratio. This provides a more nuanced and robust quantification of privacy leakage.
\end{itemize}
\end{remark}

Now, with Definitions \ref{def:Local_DP}, \ref{def:Renyi_Divergence}, and \ref{def:Local_Renyi}, we move on to Lemma \ref{lem:Local_RDP} and Theorem \ref{thm:Local_DP} below, which ultimately provide a nice result of how $\hat{f}\left\{\sqrt{w} \hat{f}^{-1}(x) + \sqrt{1-w} Z\right\}$ satisfies $(\epsilon^*, \delta)$-Differential Privacy with high probability as sample size $n \rightarrow \infty$, where $Z \sim N(0, \mathbf{I}_d)$ and $\hat{f}$ is the estimated invertible transport map from MAF.

\begin{lemma} \label{lem:Local_RDP}
Set $\mathcal{M}(\cdot)$ to be the perturbation generated by applying Gaussian noise 
\begin{equation}\mathcal{M}(g(x)) = g(x) + N(0, \sigma^2 \mathbf{I}_d), \label{eq:GaussianInjection}\end{equation}
where $g(x) \in \mathbb{R}^d$ is a function such that its sensitivity 
\begin{equation} \label{eq:Sensitivity}
\Delta_2(g) = \sup_{||x - y||_2 \leq 1} ||g(x) - g(y)||_2
\end{equation}
is bounded, i.e., 
$\Delta_2(g) \leq B,$ for a constant $B > 0.$  Then, the mechanism $\mathcal{M}(\cdot)$ satisfies $(\alpha, \epsilon_R)$ Local RDP of order $\alpha$ with 
$$\epsilon_R = \frac{\alpha \times \Delta_2(g)^2}{2 \sigma^2}.$$ 
\end{lemma}
Note that the noise injection on the latent space in our method for the true transport map $f$ is
$$X \rightarrow \mathcal{M}\left\{f^{-1}(X)\right\} = \sqrt{w}f^{-1}(X)  + N(0, (1 - w) \mathbf{I}_d).$$
It can be viewed as a special case of the Gaussian Mechanism (\ref{eq:GaussianInjection}) in Lemma \ref{lem:Local_RDP} for both $f$ and when replacing $f$ with estimated $\hat{f}$. Coupled with the boundedness of $\Delta_2(f^{-1})$, Lemma \ref{lem:Local_RDP} provides a theoretical backbone for Theorem \ref{thm:Local_DP} below.

\begin{theorem} \label{thm:Local_DP}
First, consider a trained MAF transport map, $\hat{f} = \hat{f}_k \circ \cdots \circ \hat{f}_1,$  where each $\hat{f}_j, j \in \{1, \dots, k\}$ has mapping defined in (\ref{eq:Forward}). We assume the following:
\begin{enumerate}
    \item each function $\mu_i(\cdot)$ and $\sigma_i(\cdot)$ in (\ref{eq:Forward}) are trained neural networks whose layers are spectrally normalized with the same number of layers $L$ for each $\hat{f}_j, j \in \{ 1, \dots, k \}$. That is, each weight matrix has spectral norm at most $1$.
    \item the data domain $\mathcal{X} \subset \mathbb{R}^d$ is compact.
    \item $\hat{f}$ converges to a a deterministic transport map $f$ uniformly in probability as the sample size increases and the sensitivity of $f^{-1}$ is bounded, i.e.,  $\Delta_2(f^{-1})<C$ for a constant $C.$ 
\end{enumerate}
%Under the same assumptions as in Theorem \ref{thm:Local_RDP_Proof}, 
%Consider a Masked Autoregressive Flow (MAF) Transformation defined in Equation \ref{eq:Forward}. 
Now, let
\begin{equation} \label{eq:Mechanism_Privacy}
\tilde{\mathcal{M}}(x)  = \sqrt{w} \hat{f}^{-1}(x) + \sqrt{1-w} Z,
\end{equation}

where $w \in (0,1)$ is a parameter controlling the perturbation size and $Z \sim N(0, \mathbf{I}_d).$ Then, the mechanism $\hat{f}\left\{\tilde{\mathcal{M}}(\cdot) \right\}$ satisfies $(\epsilon^{*}, \delta)$-Local DP for any $0 < \delta < 1$ and a prescribed $\epsilon^{*}> 0$ and some constant $C' > 0,$ with
    \[
%    \epsilon^* = \frac{w \Delta_2^2(f^{-1}}{2(1-w)} + \frac{\Delta_2(f^{-1}) \sqrt{2w \log(1/\delta)}}{\sqrt{1-w}}.
 \epsilon^* = \frac{w C^2}{2(1-w)} + \frac{C \sqrt{2w \log(1/\delta)}}{\sqrt{1-w}}
    \]
in probability, i.e., 
$$ \mathbb{P}\left(\mathbb{P}[\hat{f} \left\{ \tilde{\mathcal{M}}(x)\right\} = y] \leq e^{\epsilon^*} \mathbb{P}[\hat{f} \left\{\tilde{\mathcal{M}}(x') \right\} =y ]+\delta\right) \rightarrow 1$$
as $n \rightarrow \infty$ for any $x, x'$ and $y.$

% \textcolor{orange}{[WONDERING WHY Knowing $f$ is Lipschitz IS NEEDED AS ASSUMPTION (Address)]}
% If $f$ \textcolor{orange}{[WONDERING WHY THIS IS NEEDED AS ASSUMPTION]}, $f^{-1}$ are Lipschitz Continuous and the data domain $\mathcal{X}$ is compact, so that the sensitivity $\Delta_2(f^{-1})$ is finite, $f(\mathcal{M}(X))$ satisfies $(\epsilon^*, \delta)$-Differential Privacy with $\mathcal{M}(X)$ from Equation \ref{eq:Our_Procedure} with \[
% \epsilon^* = \frac{w \Delta_2^2(f^{-1})}{2(1 - w)} + \frac{\Delta_2(f^{-1}) \sqrt{2w \log(1/\delta)}}{\sqrt{1 - w}}.
% \] for any $0 < \delta < 1.$ 
\end{theorem}
The proof of Lemma \ref{lem:Local_RDP} and Theorem \ref{thm:Local_DP} are provided in Appendix \ref{app:LocalDP_Appendix}.  Note that Lemmas \ref{lem:Lipschitz_Spectral} and \ref{lem:Bounded_Lat_Space} imply that $f^{-1}$ is Lipschitz and $\Delta_2(f^{-1})$ is bounded, which justifies bounding $\Delta_2(f^{-1})$ by a positive constant $C.$ The interpretation is that as $w \rightarrow 0$, the mechanism becomes more private, i.e., $\epsilon \rightarrow 0$, and as $w \rightarrow 1$, the mechanism becomes more accurate at the cost of weaker privacy. 

Theorem \ref{thm:Local_DP} suggests that when the perturbation parameter $w$ is sufficiently close to zero, privacy can be well protected. However, to preserve the utility of the synthetic data, namely its ability to reproduce statistical analysis results from the original data, we prefer to use a value of $w$ that is not too small. This raises the question: what is the largest value of $w$ that still provides a meaningful level of privacy protection? To explore this, let $X \in \mathbf{X}=\{X_1, \cdots, X_n\},$ and let $\tilde{X}$ be its perturbed version using a given $w.$  To ensure adequate privacy protection,  the distance between $X$ and $\tilde{X}$ should exceed the minimum distance between $X$  and any other data point in the database $\mathbf{X}:$
$$d_{\min}(X)= \min_{X_i\in \mathbf{X}-\{X\}}\|X_i-X\|_2.$$ 
Otherwise, $\tilde{X}$ would be closet to its original counterpart $X$, making it identifiable and compromising privacy. Under mild regularity conditions, i.e., assuming $X_1, \cdots, X_n$ are i.i.d. observations from a smooth distribution on a compact set, it is well known that $d_{\min}(X)$ is in the order of $O(n^{-1/d}),$ where $d$ is the data dimension.  On the other hand, the distance between $X$ and $\tilde{X}$ is approximately proportional to $\sqrt{1-w}$ for $w$ close 1.  Equating this with $d_{min}(X)$, we find that to match the perturbation magnitude with the minimal pairwise distance, $w$ should be in the order of 
$$1-O(n^{-2/d}).$$ 
 This suggests that choosing $w$ on the order of $1-O(n^{-2/d})$ may provide a basic level of privacy protection, ensuring that synthetic points are not trivially matched to their original counterparts.

Despite this theoretical rate, the perturbation parameter $w$ must still be selected empirically in practice. Following the same rational, \(w\) can be chosen such that
\begin{equation} \label{eq:Probability4w}
\mathbb{P}\left(\|X_j - X_i\|_2 < \|\tilde{X}_i - X_i\|_2\right) \approx \frac{1}{n(n-1)}\sum_{1\le i<j\le n} I\left(\|X_j-X_i\|_2<\|\tilde{X}_i-X_i\|_2\right)
\end{equation}
exceeds a pre-specified threshold. This criterion ensures that, for a randomly selected observation $X_i$ from the original dataset $\mathbf{X},$ there is a reasonable chance that another real observation $X_j$ is closer to $X_i$ than its synthetic counterpart $\tilde{X}_i.$  Selecting $w$ in this way helps reduce the risk that synthetic samples closely replicate their original inputs, thereby mitigating potential memorization and enhancing privacy protection.

$w$ selected based on this criterion ensures that there is a reasonable probability that for a random selected observation in $\mathbf{X},$ there is another real observation closer to it than its perturbed value in the synthetic dataset, reducing the risk that synthetic samples memorize their original inputs. 

Alternatively, we can split observed data into two sets: a training set $D_1$ and a test set $D_0$. We then train a MAF model using $D_1$  and then generate a synthetic dataset $\tilde{\cal D}_1$ by perturbing observations in $D_1$ using the trained MAF model and a given perturbation size $w$. For each $X \in D_1 \cup D_0$, we compute a membership score $$d_{\tilde{D}_1}(X)=\min_{\tilde{X} \in \tilde{D}_1} \|X - \tilde{X}\|_2,$$ 
where a smaller value suggests that \( X \) may have been memorized by the generative model as an observation from the original training set $D_1$. We assign binary membership labels:  $Y=1$ if \( X \in D_1 \) (training set) and \( Y=0 \) if \( X \in D_0 \) (test set),  and then compute the AUC under the ROC curve \citep{hayes2017logan, chen2020gan} using pairs \( \left\{d_{\tilde{D}_1}(X), Y\right\},  X \in D_1 \cup D_0 \). An AUC value close to $0.5$ indicates that the synthetic data does not reveal whether a given data point was part of the training set, thus suggesting low risk of privacy leakage. In practice, we may choose the largest perturbation parameter $w$ such that the resulting AUC remains below a pre-specified threshold (e.g., 0.55). This criterion is closely related to Membership Inference Attacks (MIAs), which assesses whether a particular observation was used to train the generative model.  More discussion on MIA can be found in Section \ref{s:Simulation_Experiment_1}

\subsubsection{ Meta Analysis based on Synthetic Data}

\label{sec:Meta_Analysis_Syn}

Suppose that there are $K$ datasets $\mathbf{X}_1, \cdots, \mathbf{X}_K$ with their observations being drawn from distributions $P_1, \cdots, P_K$, respectively.  We generate perturbed versions of the original data in each of the \( K \) studies using Algorithm \ref{ag:Perturb}.  This produces \( K \) synthetic sets $\tilde{\mathbf{X}}_1, \cdots, \tilde{\mathbf{X}}_K,$ which can be used to enable exploratory and inferential analyses without compromising the confidentiality of the original individual level records.  Specifically, let 
$$\mathbf{X}_k=\{X_{k1} \cdots, X_{kn_k}\} ~~\mbox{and}~~ \tilde{\mathbf{X}}_k=\{\tilde{X}_{k1w}, \cdots, \tilde{X}_{kn_kw}\},$$ 
where $X_{ki}, i=1, \cdots, n_k$ are $n_k$ i.i.d. copies of a $d$-dimensional random vector $X_k\sim P_k,$  $\tilde{X}_{kiw}$ is perturbed $X_{ki}$ with an estimated transport map $\hat{f}$ and a tuning parameter $w\in [0, 1]$. Also, let the empirical distribution of $\mathbf{X}_k$ and $\tilde{\mathbf{X}}_k$ be denoted by $\hat{P}_k$ and $\hat{P}_{w,k}$ respectively.  The justification of privacy protection is already established in Section \ref{s:Privacy_Protection} under appropriate regularity conditions.  We are focusing on validity of meta analysis based on synthetic data in this section. After generating these synthetic datasets, we may estimate the parameter of interest based on them, i.e., obtaining 
$$\left\{\theta(\hat{P}_{w,k}), \sigma^2(\hat{P}_{w,k}) \mid k=1, \cdots, K\right\}$$
and proceed with a meta analytic model to combine the results.  Here, we used the notations $\theta(\hat{P}_{w,k})$ and $\sigma^2(\hat{P}_{w,k})$ to emphasize their dependence on synthetic data $\tilde{\mathbf{X}}_k$, where $\hat{P}_{w,k}$ is the probability measure induced by the empirical distribution of $n_k$ observations in $\tilde{\mathbf{X}}_k.$ When there is no ambiguity, we also use $\tilde{\theta}_{w,k}$ and $\tilde{\sigma}^2_{w, k}$ to replace $\theta(\hat{P}_{w,k})$ and $\sigma^2(\hat{P}_{w,k}),$ respectively. 

In the random effects model for meta analysis, 
$$\theta(\hat{P}_{w,k}) \sim N\left(\theta_0, \sigma^2(\hat{P}_{w,k})+\tau_0^2 \right), k=1, \cdots, K.$$
Moreover, $\theta_0$ can be estimated as in Section \ref{sec:Meta_Analysis}, i.e., by
\begin{equation} \label{eq:Synthetic_Estimator}
\tilde{\theta}_R=\frac{\sum_{k=1}^K (\sigma^{2}(\hat{P}_{w,k})+\tilde{\tau}^2)^{-1}\theta(\hat{P}_{w,k}) }{\sum_{k=1}^K (\sigma^{2}(\hat{P}_{w,k})+\tilde{\tau}^2)^{-1}},    
\end{equation}
where
$$ \tilde{\tau}^2=\max\left\{0,  \frac{\sum_{k=1}^K (\theta(\hat{P}_{w,k})-\tilde{\theta}_F)^2/\sigma^2(\hat{P}_{w,k})-(K-1)}{\sum_{k=1}^K 1/\sigma^2(\hat{P}_{w,k})-(\sum_{k=1}^K 1/\sigma^{4}(\hat{P}_{w,k}))/(\sum_{k=1}^K 1/\sigma^{2}(\hat{P}_{w,k}))} \right\} $$
and
$$\tilde{\theta}_F=\frac{\sum_{k=1}^K \theta(\hat{P}_{w,k})/ \sigma^{2}(\hat{P}_{w,k})}{\sum_{k=1}^K 1/\sigma^{2}(\hat{P}_{w,k})}. $$
As $K \rightarrow \infty,$ we have
$$\sqrt{K}\left(\tilde{\theta}_R -\theta_0  \right) \rightarrow N(0, \sigma_R^2)$$
in distribution, where 
$$\sigma_R^2= \left( \sum_{k=1}^K (\sigma_k^{2}(P_k)+\tau_0^2)^{-1} \right)^{-1},$$
which can be estimated as 
$$\tilde{\sigma}_R^2= \left( \sum_{k=1}^K (\sigma_k^{2}(\hat{P}_{w,k})+\tilde{\tau}_0^2)^{-1} \right)^{-1}.$$
Consequently, the 95\% CI for $\theta_0$ can be constructed as 
\[
\left(\tilde{\theta}_R - z_{\alpha/2} \times \tilde{\sigma}_R, \tilde{\theta}_R + z_{\alpha/2} \times \tilde{\sigma}_R \right)
\]
where \( z_{\alpha/2} \) is the critical value from the standard Gaussian distribution.
On the other hand, a parallel meta analysis based on observed data $\mathbf{X}_1, \cdots, \mathbf{X}_K$ can also be conducted to obtain 
$$(\hat{\theta}_R, \hat{\sigma}_R^2).$$
A natural question is if there is any efficiency loss due to the fact that the synthetic rather than observed data is used to obtain an estimator for $\theta_k$ in individual studies and $\theta_0$ in the subsequent meta analysis. %We will investigate relevant theoretical properties of estimators based on synthetic data in the next subsection.

%\subsubsection{Theoretical Properties of Perturbed Estimator \(\hat{\theta}_k\) and Meta Analysis}
%\label{sec:Validity_Perturbed_Est_Meta_Analysis}

Now, we analyze the asymptotic properties of the meta analysis estimator $ \tilde{\theta}_R$ derived from generated synthetic data based on the results in Section \ref{sec:Validity_Perturbed_Est_Single_Study}. The basic idea is that while each study may incur perturbation bias, aggregation across \(K\) studies mitigates this effect. Indeed, under appropriate regularity conditions, the perturbed estimator remains asymptotically equivalent to the non-private estimator, as shown below.

\begin{theorem} \label{thm:Meta_Analysis_CLT}
 Now, let $n=(n_1+\cdots+n_K)/K$ be the average sample size of $K$ studies and $n_k/n, k=1, \cdots, K$ is uniformly bounded below away from zero and above. Suppose that data from different studies are independent and the regularity conditions in Theorem \ref{thm:Meta_Analysis_Rate} hold for all studies. In addition, we assume the following:
 \begin{enumerate}
\item $K^{1/2}(\delta_{1n}^2+\delta_{1n}\sqrt{\delta_{2n}})=o(1).$ 
\item $ K/n=o(1).$
\item $\tau_0>0.$
\end{enumerate}
Then,
\[
\sqrt{K}(\tilde{\theta}_R - \hat{\theta}_R) = o_p(1)
\]
and
\[
\sqrt{K}(\tilde{\theta}_R - \theta_0) \overset{d}{\to} N\left(0, \frac{1}{K^{-1}\sum_{k=1}^K (\sigma_k^{2}+\tau_0^2)^{-1} }\right).
\] 
\end{theorem}
The justification is straightforward. Based on Theorem \ref{thm:Meta_Analysis_Rate},
$$\theta(\hat{P}_{w,k}) - \theta(\hat{P}_k) = O_p(\delta_{1n}^2+\delta_{1n}\sqrt{\delta_{2n}}) + o_p(n^{-1/2}).$$
Since $\hat{\theta}_R$ and $\tilde{\theta}_R$ are a weighted average of $\{\theta(\hat{P}_k), k=1,\cdots, K\}$ and $\{\theta(\hat{P}_{w,k}), k=1, \cdots, K\}$ respectively, with equivalent weighting schemes, i.e., the weight for the $k$th study $\propto (\sigma^2(P_k)+\hat{\tau}^2_0)^{-1},$
$$\sqrt{K}(\tilde{\theta}_R -\hat{\theta}_R)=O_p(1) $$ 
under the assumption $\sqrt{K}(\delta_{1n}^2+\delta_{1n}\sqrt{\delta_{2n}})=o(1).$ The weights for constructing $\hat{\theta}_R$ and $\tilde{\theta}_R$ are equivalent, which is due to the equivalence of $\hat{\theta}_F$ and $\tilde{\theta}_F.$  Since $K$ is typically substantially smaller than $n_k,$ the approximation of $\tilde{\theta}_R$ to $\hat{\theta}_R$ is oftentimes better than the approximation of $\theta(\hat{P}_{w,k})$ to $\theta(\hat{P}_k)$ at the individual study level. Therefore, the reproducibility of meta analysis based on synthetic data is expected to be high. 

In summary, Theorem \ref{thm:Meta_Analysis_CLT} implies that, asymptotically, the perturbed meta analysis estimator \( \tilde{{\theta}}_R \) shares the same limiting distribution as the non-private estimator \( \hat{\theta}_R \). Thus, privacy-preserving estimators retain consistency and asymptotic normality in meta analysis under appropriate regularity conditions. Although the perturbation error may decay slowly within each study, it can be effectively mitigated through aggregation. When the number of studies satisfies \( K = O(n^{1/2 + \epsilon}) \), the aggregate error vanishes asymptotically, rendering \( \tilde{{\theta}}_R \) an efficient estimator. These results illustrate a favorable privacy-utility trade-off: valid inference based on synthetic and private data is attainable when combining information across studies. In this way, meta analysis based on generated synthetic data provides a powerful framework for conducting statistically valid inference under privacy constraints.

\section{Numerical Study}

In this section, we have performed several simulation studies to examine the finite-sample performance of proposal. The simulation studies have two objectives: (1) to examine the effect of using generated synthetic data to replace the originally observed data in subsequent statistical inference; (2) to examine the privacy protection provided by sharing the synthetic data.

\subsection{Simulation Experiments}

% \subsubsection{Simulation Experiment 1}
% \label{s:Simulation_Experiment_1}

\subsubsection{Convergence Rate of $\tilde{\theta}$ in a fixed Study based on the Synthetic Data}
\label{s:Simulation_Experiment_1}

In this section, a simulation study is conducted to evaluate the performance of several methods of generating synthetic data for downstream statistical analysis. Specifically, the study compares three techniques:
\begin{enumerate}
    \item \textbf{Flow Sampling}: Synthetic Data are sampled directly from a trained MAF based on observed data. The sample size of the synthetic data can be determined by the user.
    \item \textbf{Latent Noise Injection}: Observed Data are mapped to a latent space, perturbed by noise injection, and then reconstructed. The mapping to the latent space and subsequent reconstruction are based on trained MAF described in Section \ref{sec:NewProposal}.
    \item \textbf{Direct Noise Injection}: Observed Data are perturbed by directly injecting an independent Gaussian noise. 
\end{enumerate}
The goal is to analyze how well a simple correlation coefficient estimator based on synthetic data estimates the true correlation coefficient.  Specifically, we are interested in the relationship between estimation precision and sample size of observed data for each aforementioned method generating synthetic data.  

In one iteration of the simulation study, the ``observed'' data of size $n$ are generated from a multivariate Gaussian distribution with a compound symmetric variance-covariance matrix of 
$$\Sigma = (1 - \rho) \mathbf{I}_5 + \rho 11^T,$$
which gives a correlation coefficient of  $\rho = 0.9$ between any two different components of $X\in \mathbb{R}^5$. A MAF is then trained on the generated dataset such that $X = \hat{f}(Z)$ is approximately $N(0, \Sigma),$ the distribution used to generate the observed data, where $Z\sim N(0, \mathbf{I}_5).$  
The model consists of one hidden layer of size $50$ with $5$ autoregressive flow transformations. We train $>100$ iteration steps using the Adam Optimizer to maximize the corresponding log-likelihood function.  We simply set the batch size to be $n$, the total sample size of generated data, in maximizing the objective function. We also normalize the weight matrix by its maximum singular value to improve training stability and for enforcing Lipschitz Continuity \citep{miyato2018spectral}. See Lemma \ref{lem:Lipschitz_Spectral} in Appendix. With trained MAF, $\hat{f}(\cdot),$ we then generate 
\begin{itemize}
\item a synthetic dataset based on the MAF model (Method 1: Flow Sampling), 
$${\cal D}_1=\left\{\hat{f}(Z_i) \mid Z_i\sim N(0, \mathbf{I}_5), i=1, \cdots, 50,000 \right\},$$ 
where the sample size of ${\cal D}_1$ can be arbitrarily large, and we select 50,000. 
\item a synthetic dataset of size $n$ by perturbing ``observed data'' on the latent space (Method 2: Latent Noise Injection), 
$${\cal D}_2=\left\{\hat{f}(\sqrt{w}\hat{f}^{-1}(x_i)+\sqrt{1-w}Z_i) \mid Z_i\sim N(0, \mathbf{I}_5), i=1, \cdots, n \right\},$$  
where $w$ is a tuning parameter between 0 and 1. 
\item a synthetic dataset of size $n$ by injecting a random noise to ``observed data'' (Method 3: Direct Noise Injection), 
$${\cal D}_3=\left\{\sqrt{w}x_i+\sqrt{1-w}\mbox{diag}\{\hat{\sigma}_1, \cdots, \hat{\sigma}_d\}Z_i \mid Z_i\sim N(0, \mathbf{I}_5), i=1, \cdots, n \right\},$$
where $\hat{\sigma}_k$ is the empirical standard deviation of the $k$th component of $X$,  $Z_i \sim N(0, \mathbf{I}_d)$ and $w \in (0, 1)$ is a tuning parameter. 
\end{itemize}
To assess the performance of correlation coefficient estimator based on synthetic data, we compute the MLE of \( \rho \) for each synthetic dataset under the compound symmetry assumption for the variance-covariance matrix:
\[
 \frac{2}{p(p - 1)} \sum_{1\le i<j\le p} \hat{\rho}_{ij},
\]
where $\hat{\rho}_{ij}$ is the sample Pearson's correlation coefficient between the $i$th and $j$th components of $X$. Now, let the estimator based on synthetic data ${\cal D}_j$ be $\hat{\rho}_j, j=1, 2, 3.$  In addition, for Latent Noise Injection and Direct Noise Injection methods, different $w$'s are used to control the perturbation size. Specifically, we use $w=0, 0.25, 0.5, 0.75$ and $1.$  When $w=0,$ synthetic dataset generated via Latent Noise Injection degenerates into a data generated via Flow Sampling (Method 1) except with a sample size of $n$ instead of $50000.$  When $w=1,$ synthetic datasets generated via Latent Noise Injection and Direct Noise Injection are identical to the original ``observed'' data, since the injected noises become zero.

After repeating this process $B=100$ times, the mean absolute difference (MAD) and empirical bias (EB) are computed as  
$$ MAD_j=\frac{1}{B}\sum_{b=1}^B |\hat{\rho}_j^{(b)}-\rho|~~\mbox{and}~~  Bias_j=\frac{1}{B}\sum_{b=1}^B \hat{\rho}_j^{(b)}-\rho, $$
respectively, where $\hat{\rho}_j^{(b)}$ is the MLE for $\rho$ in the $b$th iteration of the simulation based on synthetic dataset ${\cal D}_j, j=1, 2, 3,$ and $\rho=0.9$ is the true correlation coefficient. This simulation is repeated for
different sample sizes 
$n \in \{2500k \mid k = 1, 2, \dots, 20\}.$  We then examine if a power law of the form $MAD_j\propto n^{-\alpha}$ can characterize the impact of sample size of observed data on the estimation precision for estimators from synthetic data measured by MAD and EB. If $\alpha\approx 0.5,$ then the estimator based on synthetic data has the same convergence rate as the MLE based on ``observed'' data, i.e.,  converges to the true parameter at the regular root $n$ rate \citep{casella2001theory}.  $\alpha< 0.5$ indicates that there is nontrivial information loss due to the use of synthetic data rather than ``observed'' data for estimating parameters of interest. 

\begin{figure}[H]
    \noindent
    \begin{subfigure}[t]{0.5\textwidth}
        \centering
        \includegraphics[width=\linewidth]{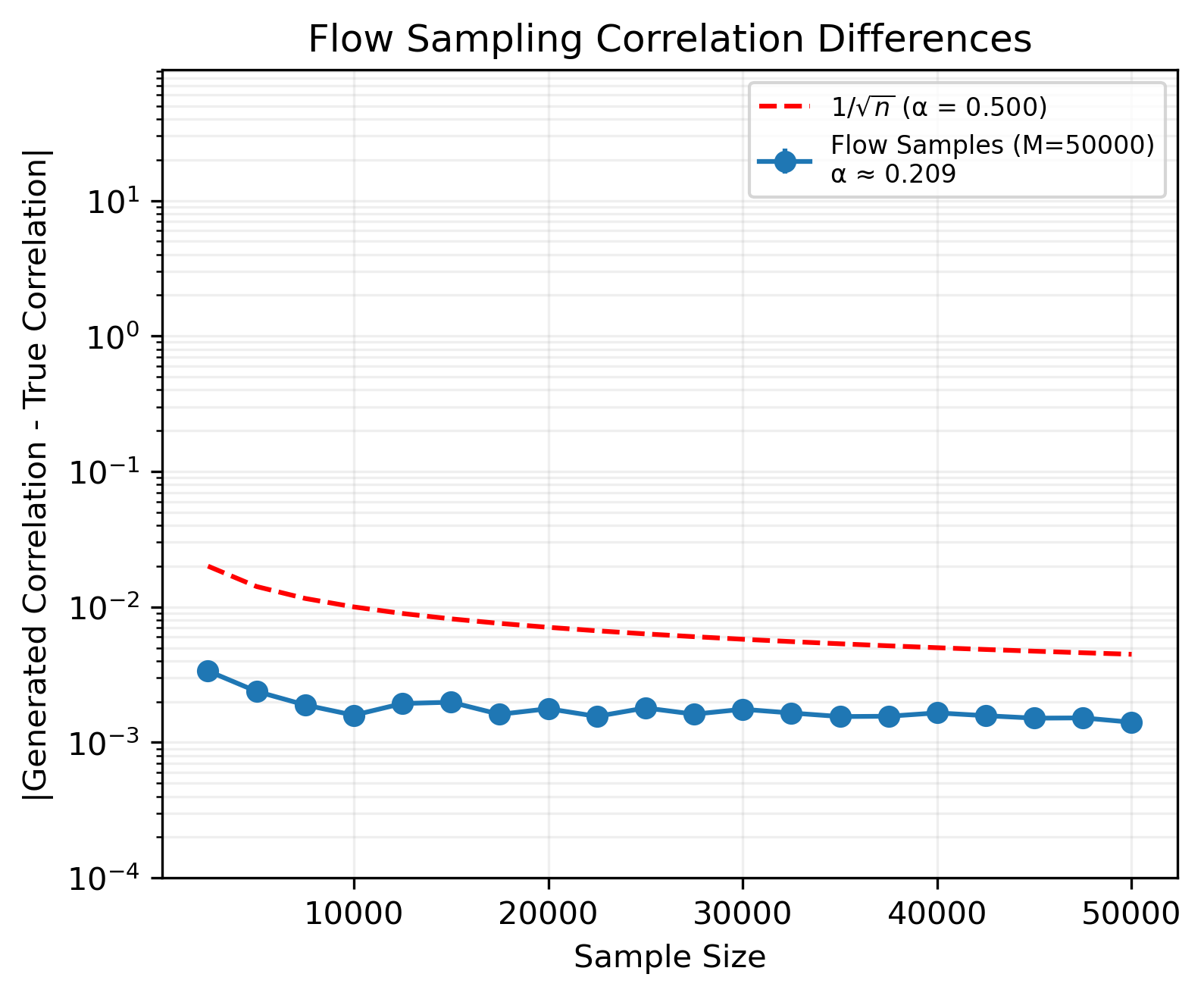}
        \caption{Flow Sampling Absolute Correlation Differences}
        \label{fig:Flow_Sampling_Abs}
    \end{subfigure}
    \hspace{0.02\textwidth}
    \begin{subfigure}[t]{0.5\textwidth}
        \centering
        \includegraphics[width=\linewidth]{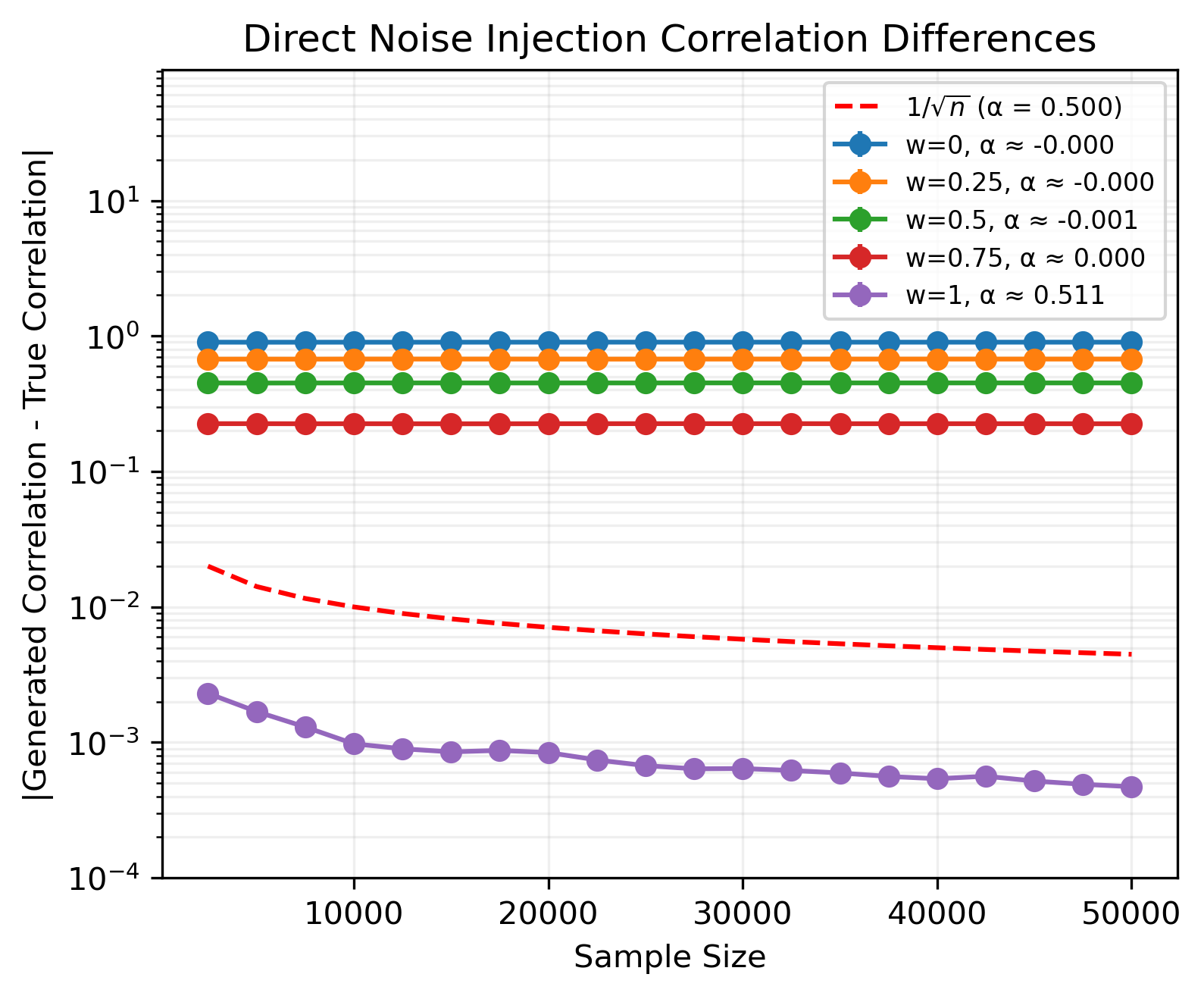}
        \caption{Direct Noise Injection Absolute Correlation Differences}
        \label{fig:Dir_Noise_Abs}
    \end{subfigure}

    \vspace{1em}

    \raggedright
    \begin{subfigure}[t]{0.5\textwidth}
        \includegraphics[width=\linewidth]{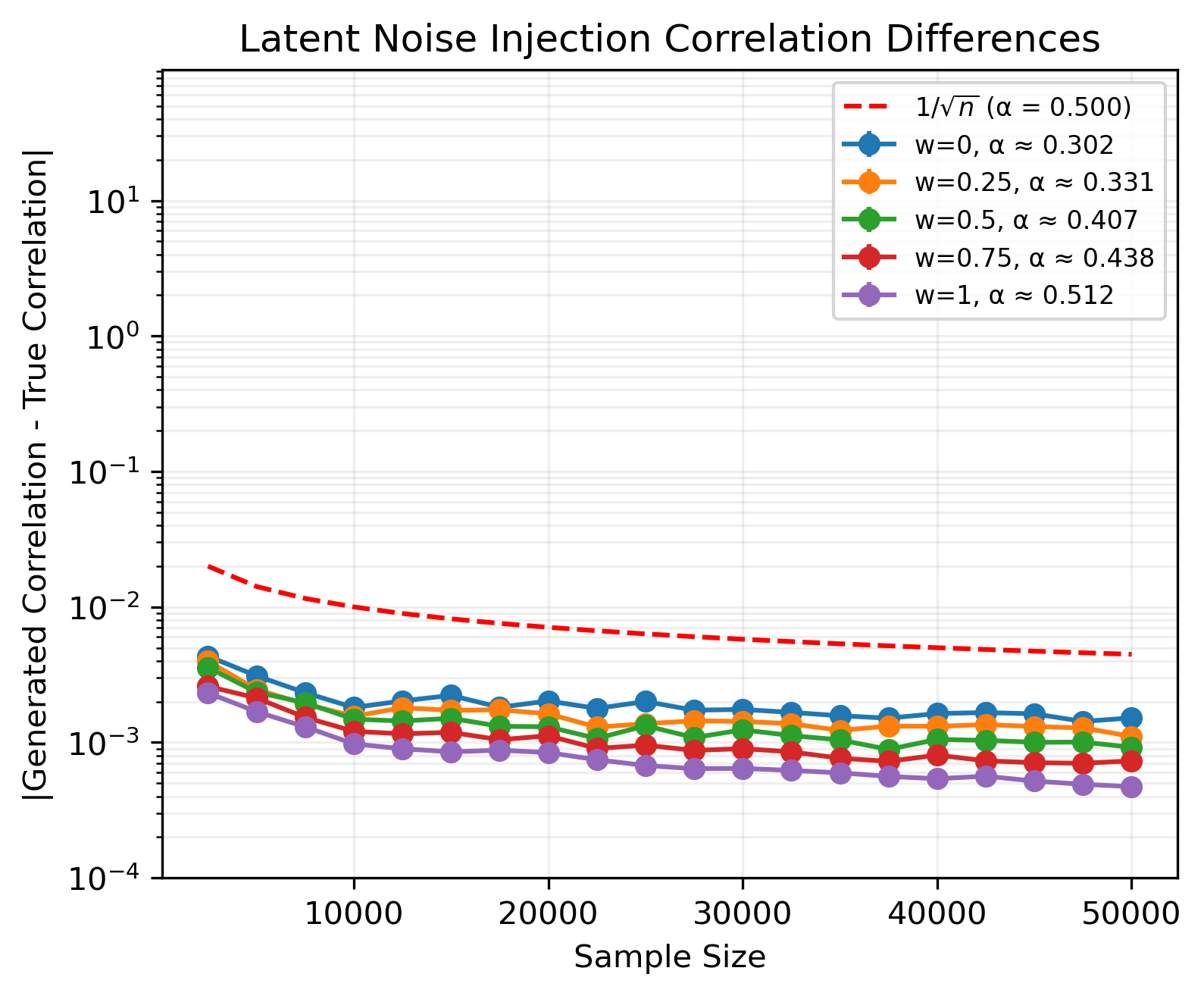}
        \caption{Latent Noise Injection Absolute Correlation Differences}
        \label{fig:Lat_Noise_Abs}
    \end{subfigure}

    \caption{A comparison of mean absolute differences in estimating the correlation coefficient based on $3$ synthetic data generation schemes ($100$ simulations); the
    red curve represents the benchmark $n^{-1/2}$ rate.}
    \label{fig:Corr_Diff}
\end{figure}

\begin{figure}[H]
    \noindent
    \begin{subfigure}[t]{0.45\textwidth}
        \centering
        \includegraphics[width=\linewidth]{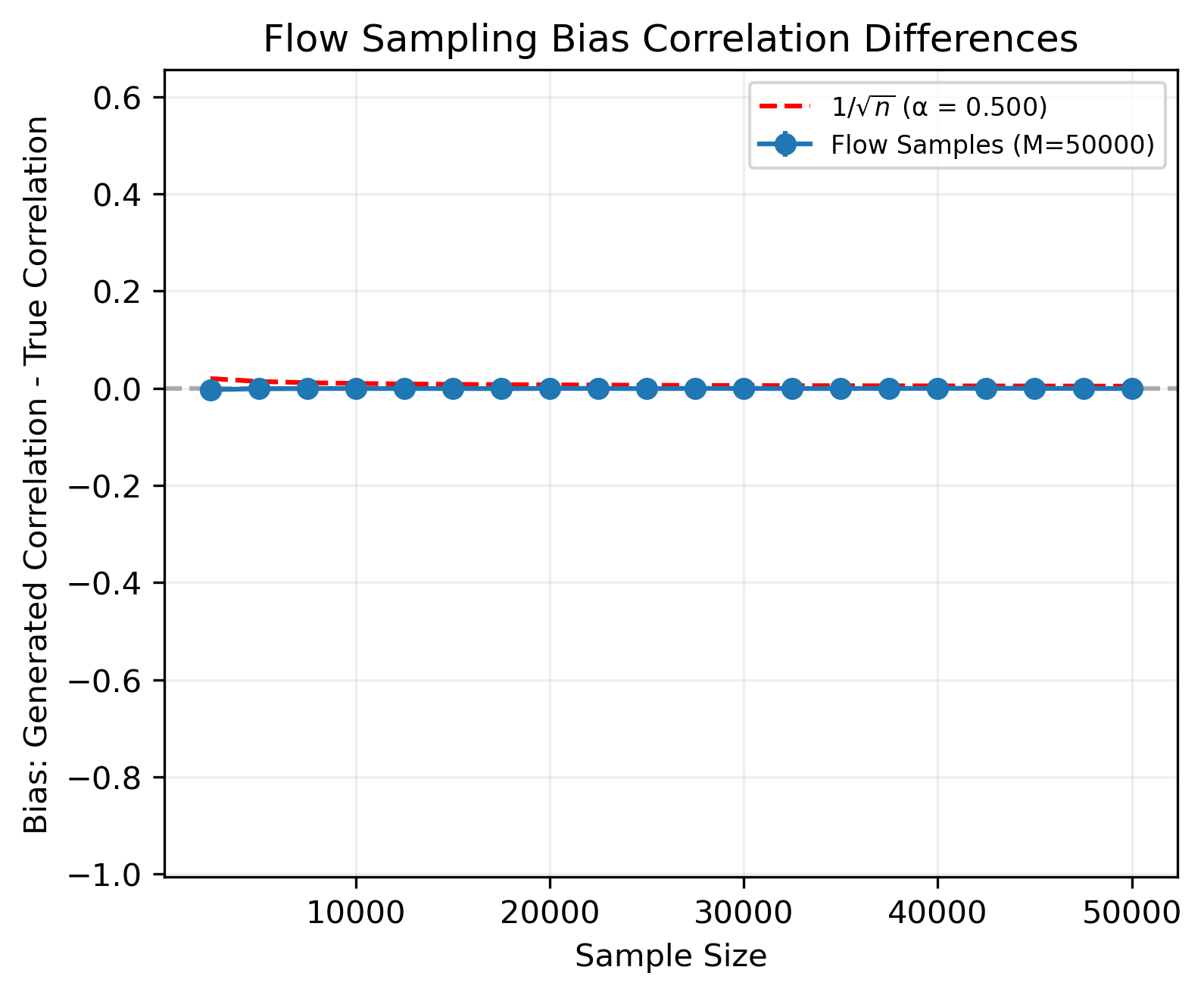}
        \caption{Flow Sampling Bias Correlation Differences}
        \label{fig:Flow_Sampling_Bias}
    \end{subfigure}
    \hspace{0.05\textwidth}
    \begin{subfigure}[t]{0.45\textwidth}
        \centering
        \includegraphics[width=\linewidth]{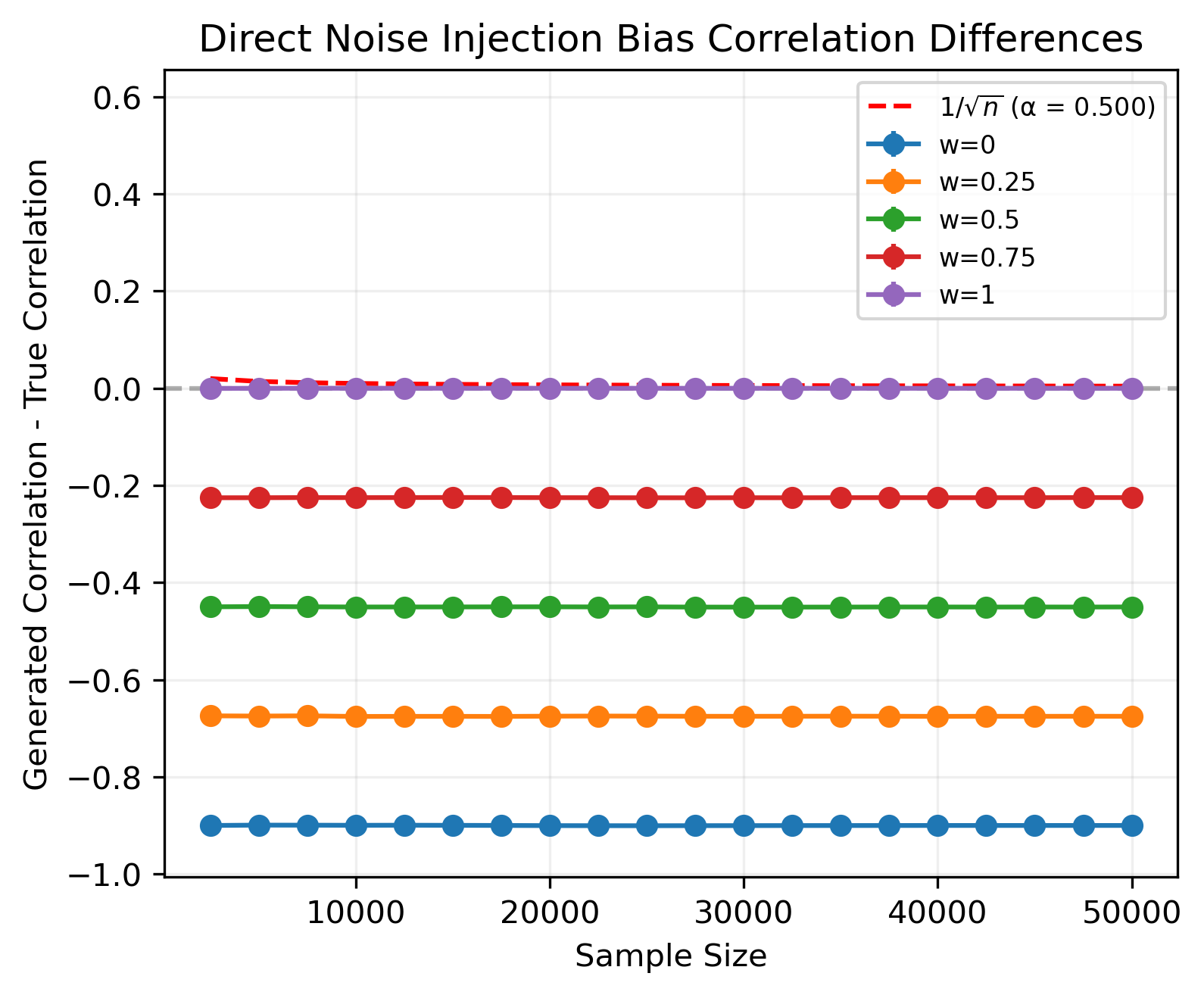}
        \caption{Direct Noise Injection Bias Correlation Differences}
        \label{fig:Dir_Noise_Bias}
    \end{subfigure}

    \vspace{1em}

    \raggedright
    \begin{subfigure}[t]{0.45\textwidth}
        \includegraphics[width=\linewidth]{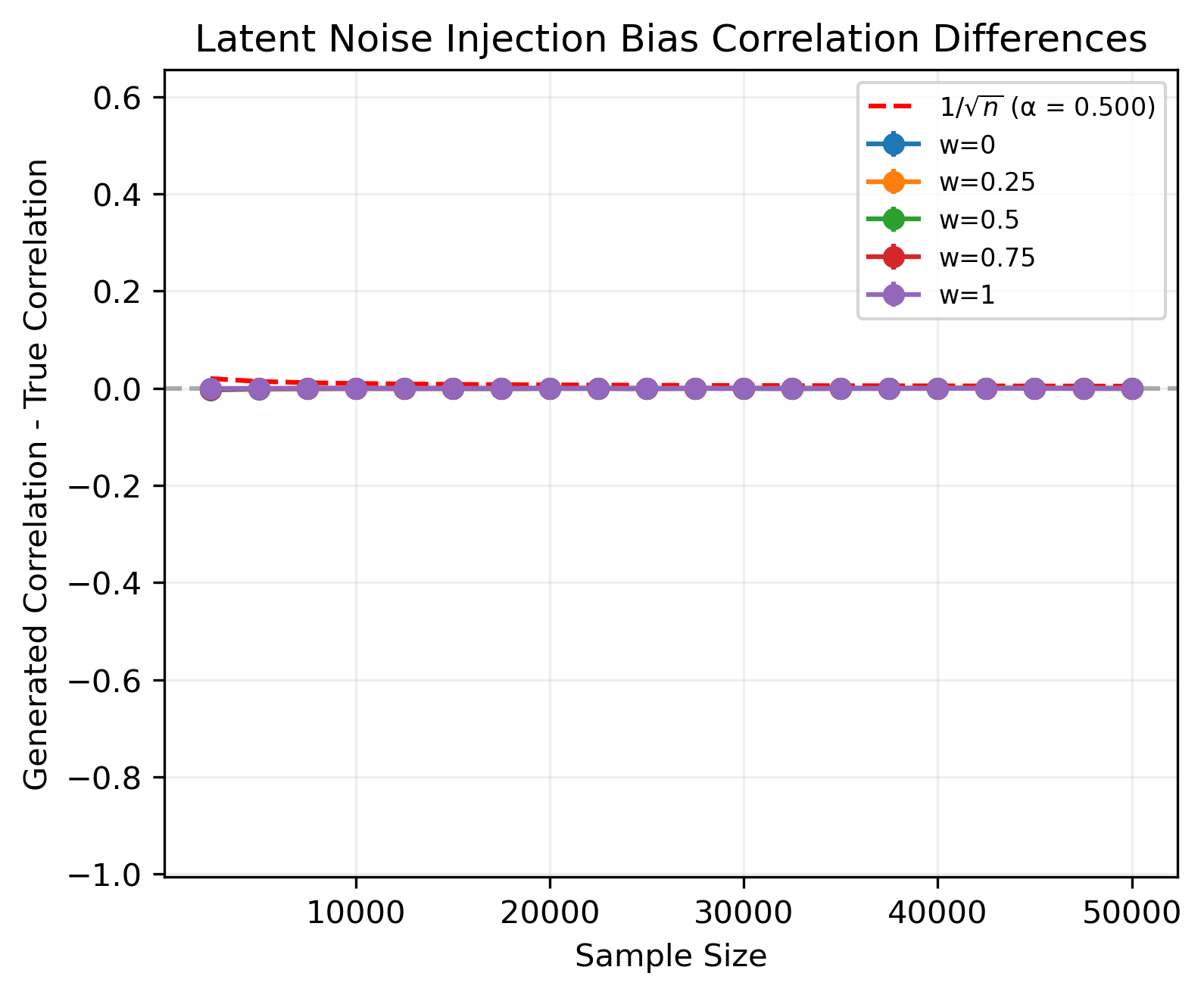}
        \caption{Latent Noise Injection Bias Correlation Differences}
        \label{fig:Lat_Noise_Bias}
    \end{subfigure}

    \caption{A comparison of biases in estimating the correlation coefficient based on $3$ synthetic data generation schemes ($100$ simulations).}
    \label{fig:Corr_Diff_Bias}
\end{figure}

The simulation results are summarized in Figures \ref{fig:Corr_Diff}, \ref{fig:Corr_Diff_Bias}, and Tables \ref{tab:Mean_Distances_Absolute_Truth}, \ref{tab:Mean_Distances_Bias_Truth}.  Figure \ref{fig:Corr_Diff} presents the MAD between estimated and true correlation coefficient for each method as a function of sample size of observed data. The theoretical convergence rate, $O_p(n^{-1/2})$, is plotted as a reference. Panel (a) of Figure 1 indicates that the estimator based on synthetic data generated via Flow Sampling exhibits a slower than the regular root $n$ rate in estimating the true correlation coefficient. It may be due to the fact that the transport map trained via MAF fails to generate synthetic data from a distribution that is sufficiently close to the underlying distribution for the ``observed'' data even if the sample size $n$ is large. Indeed, we know that in general, the Wasserstein Distance between the true distribution and the distribution induced by a Normalizing Flow trained based on a training dataset of size $n$ is slower than the regular root $n$ rate, suggesting that the estimator based on synthetic data also converges at a slower than root $n$ rate. Panel (b) of Figure \ref{fig:Corr_Diff} demonstrates that the estimator based on synthetic data generated via Latent Noise Injection converges to the true correlation coefficient at approximately a root $n$ rate when $w$ is close to 1. The convergence rate is slower than the root $n$ rate when $w$ is close to 0, as in Panel (a). This observation suggests that perturbing the observed data by injecting a random noise at a controlled level in the latent space can replace the original ``observed'' data for downstream analysis without sacrificing the estimation precision. Finally, directly injecting independent noise to observed data distorts the distribution of observed data and may severely bias the related parameter estimation, especially when $w$ is away from 1. In the current case, injecting a Gaussian noise component wise would dilute the correlation coefficient, leading to a downward bias in estimating $\rho.$  This bias doesn't diminish with increased sample size as shown in Panel (c) of Figure \ref{fig:Corr_Diff}, since it is caused by the perturbation mechanism itself. The detailed results on MAD from the simulation study are also reported in Tables \ref{tab:Mean_Distances_Absolute_Truth} and \ref{tab:Mean_Distances_Bias_Truth}. Note, we only plot for $n \in \{10000, 20000, 30000, 40000, 50000\}$ to avoid verbosity.

Figure \ref{fig:Corr_Diff_Bias} presents the EB for correlation coefficient estimator as a function of sample size of observed data. Panel (a) of Figure \ref{fig:Corr_Diff_Bias} indicates that the estimator based on synthetic data generated via Flow Sampling exhibits a near zero bias even when the sample size is as small as 2500.  It is expected that the transport map trained via MAF can induce a distribution consistently approximating the true distribution of $X.$  Panel (b) of Figure \ref{fig:Corr_Diff_Bias} demonstrates that the estimator based on synthetic data generated via Latent Noise Injection is approximately unbiased for all $w$ and all sample sizes investigated.  This observation further supports using synthetic data generated via the proposed Latent Noise Injection to replace the original data for statistic analysis.  Finally, the Direct Noise Injection method introduces a downward bias in estimating the correlation coefficient, and this bias persists with increasing sample size in Panel (c) of Figure \ref{fig:Corr_Diff_Bias}. The detailed results on EB from the simulation study are also reported in Table \ref{tab:Mean_Distances_Bias_Truth}. 

\begin{table}[H]
    \centering
    \caption{The mean absolute distances, $\mbox{average}|\hat{\rho}_{MLE}-\rho|,$ estimates over $100$ simulations for different methods and sample sizes ($\rho=0.9$) with $w \in \{0, 0.25, 0.5, 0.75, 1\}$.}
    \label{tab:Mean_Distances_Absolute_Truth}
    \begin{tabular}{ccccccc}
        \toprule
        \textbf{n} & \textbf{Method} & \textbf{w = 0} & \textbf{w = 0.25} & \textbf{w = 0.5} & \textbf{w = 0.75} & \textbf{w = 1} \\
        \midrule
        \multirow{3}{*}{10000} 
        & Flow Sampling & 0.0016 & - & - & - & - \\
        & Latent Noise Injection & 0.0018 & 0.0016 & 0.0015 & 0.0012 & 0.0010 \\
        & Direct Noise Injection & 0.8996 & 0.6755 & 0.4502 & 0.2251 & 0.0010 \\
        \midrule
        \multirow{3}{*}{20000} 
        & Flow Sampling & 0.0018 & - & - & - & - \\
        & Latent Noise Injection & 0.0020 & 0.0016 & 0.0013 & 0.0011 & 0.0008 \\
        & Direct Noise Injection & 0.9003 & 0.6750 & 0.4498 & 0.2251 & 0.0008 \\
        \midrule
        \multirow{3}{*}{30000} 
        & Flow Sampling & 0.0018 & - & - & - & - \\
        & Latent Noise Injection & 0.0018 & 0.0014 & 0.0012 & 0.0009 & 0.0006 \\
        & Direct Noise Injection & 0.9003 & 0.6753 & 0.4505 & 0.2252 & 0.0006 \\
        \midrule
        \multirow{3}{*}{40000} 
        & Flow Sampling & 0.0017 & - & - & - & - \\
        & Latent Noise Injection & 0.0016 & 0.0013 & 0.0011 & 0.0008 & 0.0005 \\
        & Direct Noise Injection & 0.9000 & 0.6754 & 0.4502 & 0.2251 & 0.0005 \\
        \midrule
        \multirow{3}{*}{50000} 
        & Flow Sampling & 0.0014 & - & - & - & - \\
        & Latent Noise Injection & 0.0015 & 0.0011 & 0.0009 & 0.0007 & 0.0005 \\
        & Direct Noise Injection & 0.9000 & 0.6752 & 0.4501 & 0.2250 & 0.0005 \\
        \bottomrule
    \end{tabular}
\end{table}

\begin{table}[H]
    \centering
    \caption{The empirical bias,  $\mbox{average}(\hat{\rho}_{MLE}) - \rho,$ over $100$ simulations for different methods and sample sizes ($\rho=0.9$) with $w \in \{0, 0.25, 0.5, 0.75, 1\}$.}
    \label{tab:Mean_Distances_Bias_Truth}
    \begin{tabular}{ccccccc}
        \toprule
        \textbf{n} & \textbf{Method} & \textbf{w = 0} & \textbf{w = 0.25} & \textbf{w = 0.5} & \textbf{w = 0.75} & \textbf{w = 1} \\
        \midrule
        \multirow{3}{*}{10000} 
        & Flow Sampling & -0.0004 & - & - & - & - \\
        & Latent Noise Injection & -0.0007 & -0.0004 & -0.0002 & -0.0001 & 0.0000 \\
        & Direct Noise Injection & -0.8996 & -0.6755 & -0.4502 & -0.2251 & 0.0001 \\
        \midrule
        \multirow{3}{*}{20000} 
        & Flow Sampling & -0.0002 & - & - & - & - \\
        & Latent Noise Injection & -0.0005 & -0.0002 & -0.0001 & 0.0000 & 0.0000 \\
        & Direct Noise Injection & -0.9003 & -0.6750 & -0.4498 & -0.2251 & 0.0001 \\
        \midrule
        \multirow{3}{*}{30000} 
        & Flow Sampling & -0.0003 & - & - & - & - \\
        & Latent Noise Injection & -0.0004 & -0.0002 & -0.0001 & -0.0001 & -0.0000 \\
        & Direct Noise Injection & -0.9003 & -0.6753 & -0.4505 & -0.2252 & -0.0000 \\
        \midrule
        \multirow{3}{*}{40000} 
        & Flow Sampling & -0.0001 & - & - & - & - \\
        & Latent Noise Injection & -0.0003 & -0.0000 & -0.0002 & -0.0000 & -0.0000 \\
        & Direct Noise Injection & -0.9000 & -0.6754 & -0.4502 & -0.2251 & -0.0000 \\
        \midrule
        \multirow{3}{*}{50000} 
        & Flow Sampling & -0.0004 & - & - & - & - \\
        & Latent Noise Injection & -0.0006 & -0.0003 & -0.0002 & -0.0001 & 0.0000 \\
        & Direct Noise Injection & -0.9000 & -0.6752 & -0.4501 & -0.2250 & 0.0000 \\
        \bottomrule
    \end{tabular}
\end{table}

Next, we assess empirical privacy leakage through membership inference attacks (MIA) and monitor how attack success varies with the perturbation parameter \(w\).  A particularly informative variant  of MIA is the distance-to-synthetic attack, applicable to models such as GANs, VAEs, and Diffusion~\citep{chen2020gan}. Unlike binary classification approaches that attempt to distinguish synthetic from real data, this method instead focuses on identifying which real data points were likely seen during training and perturbed to generate synthetic observations. Specifically, we generate synthetic data with different $w$'s in our proposed Latent Noise Injection method. To quantify the privacy protection via MIA, we attempt to differentiate originally observed data, whose privacy needs to be protected, from freshly generated new observations based on the synthetic dataset $\tilde{\mathbf{X}}.$ If there is no or very little difference between them, i.e., based on synthetic data one can't tell if a given observation $X$ is from observed data or a fresh independent observation, then there is a strong privacy protection. 

\begin{figure}[H]
    \centering
    \begin{subfigure}[t]{0.48\textwidth}
        \centering
        \includegraphics[width=\linewidth]{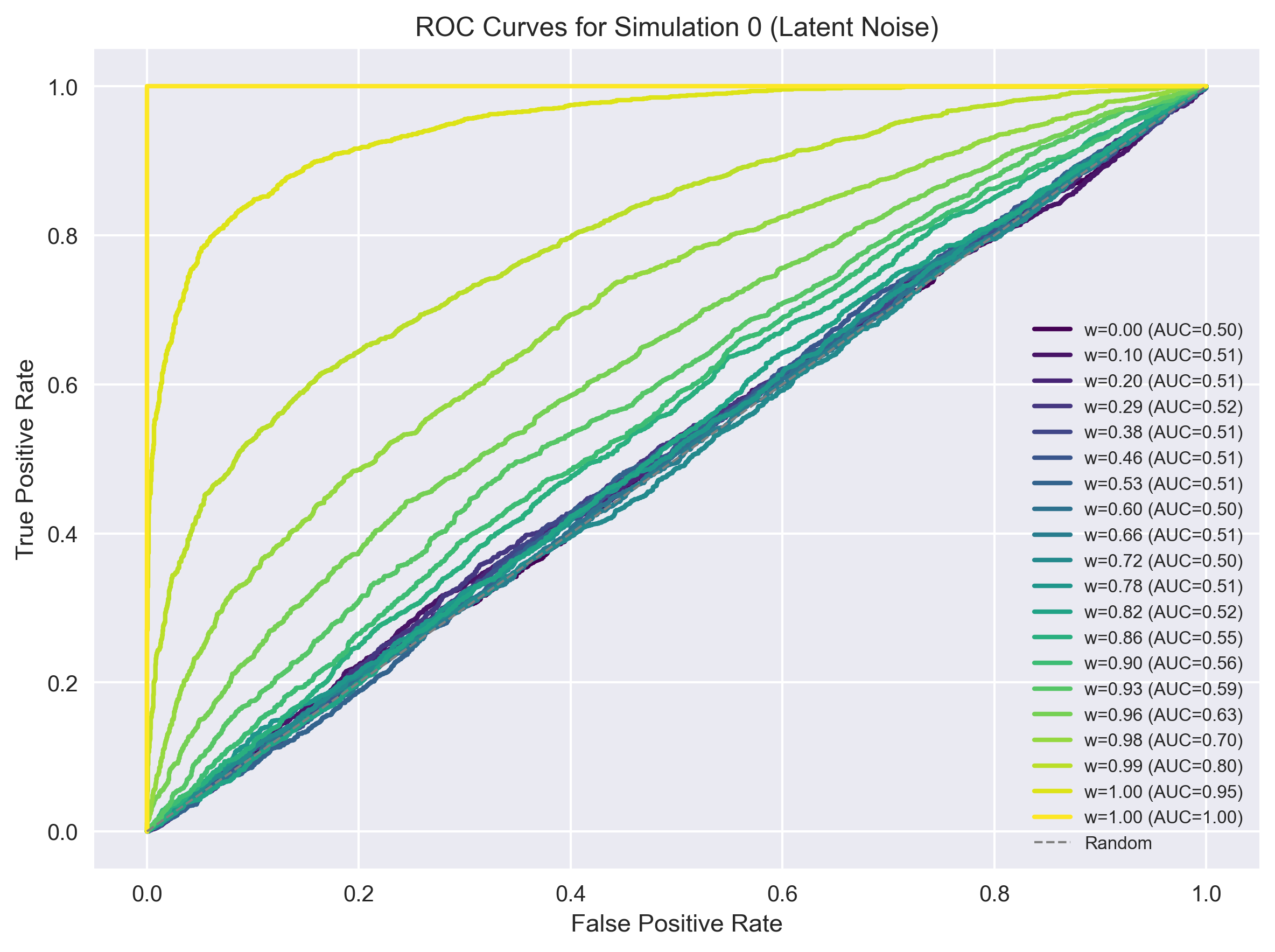}
        \caption{ROC curves of membership scores in MIA based on one set of synthetic data generated via the Latent Noise Injection method.}
        \label{fig:Roc_Latent}
    \end{subfigure}
    \hfill
    \begin{subfigure}[t]{0.48\textwidth}
        \centering
        \includegraphics[width=\linewidth]{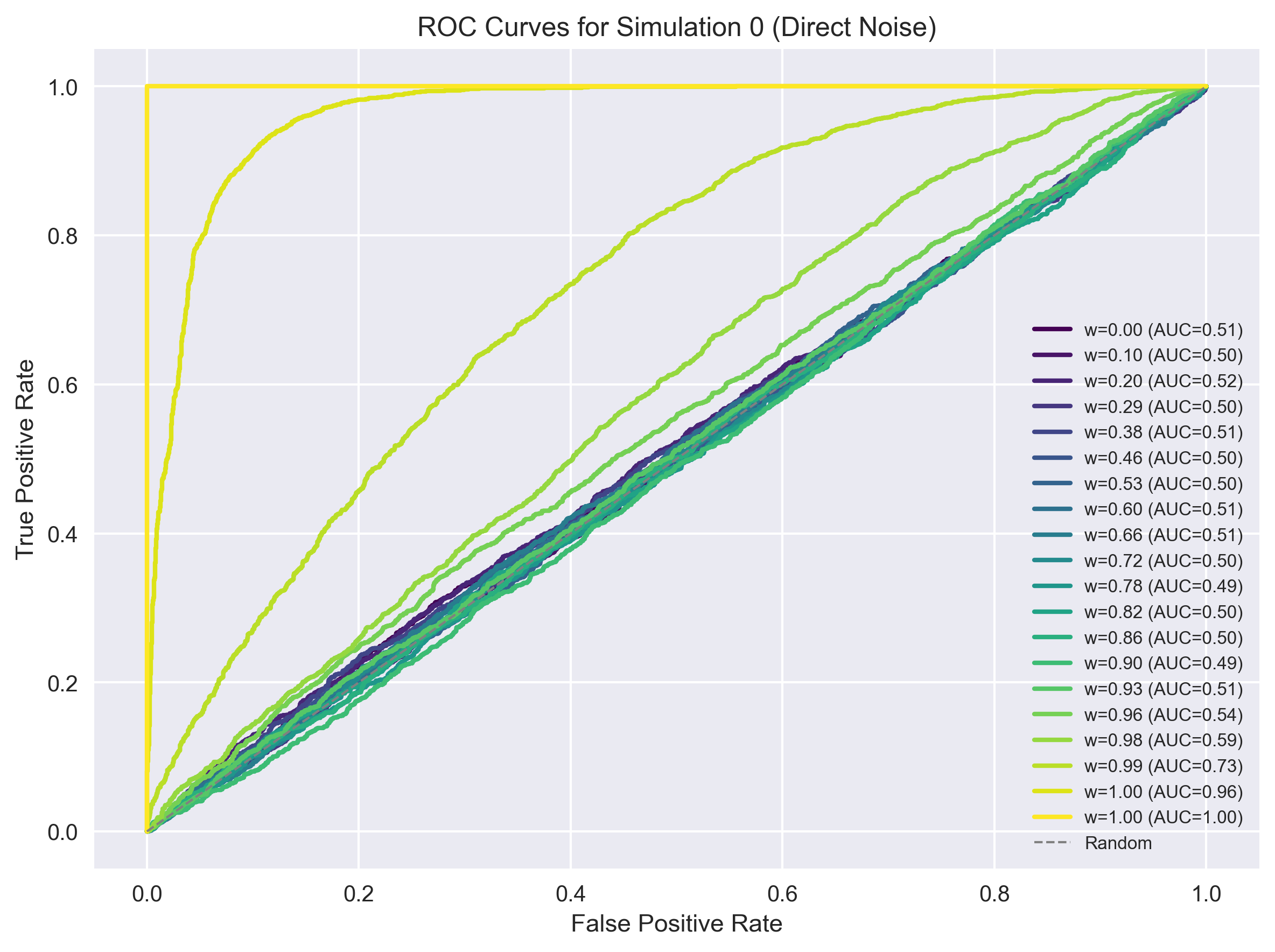}
        \caption{ROC curves of membership scores in MIA based on one set of synthetic data generated via Direct Noise Injection method.}
        \label{fig:Roc_Naive}
    \end{subfigure}

    \vspace{0.5cm}

    \raggedright

    \begin{subfigure}[t]{0.48\textwidth}
        \includegraphics[width=\linewidth]{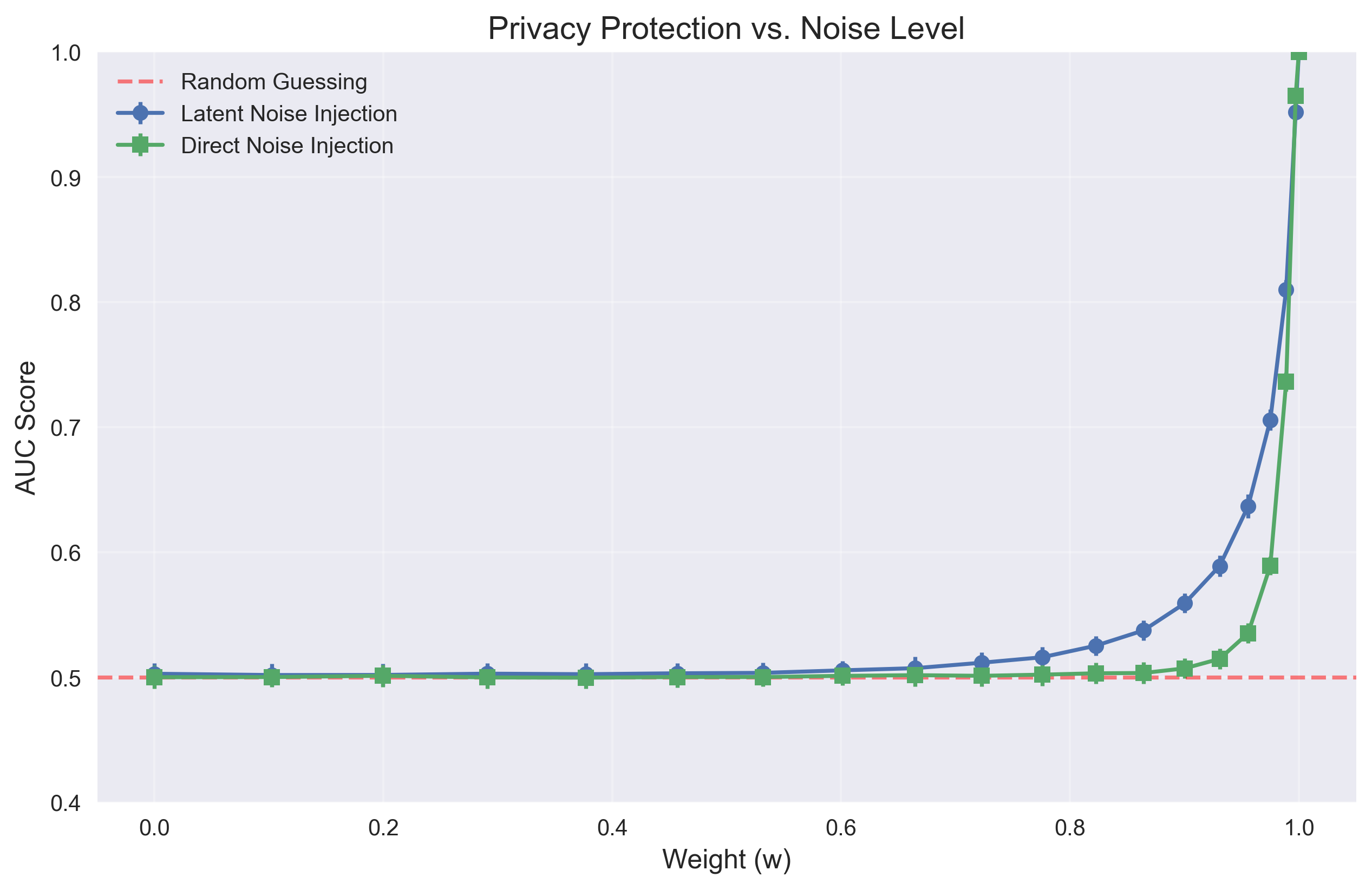}
        \caption{AUC under the ROC curve of membership scores from MIA for synthetic data generated via Latent and Direct Noise Injection Methods with different $w$.}
        \label{fig:Privacy_Auc}
    
    \end{subfigure}

    \caption{A privacy protection assessment for synthetic data generated using Noise Injection Techniques.}
    \label{fig:Privacy_Analysis}
\end{figure}

In this simulation study, we examine the performance of a specific classifier based on the distance between observation of interest $X_0$ and synthetic data $\tilde{\mathbf{X}}$ released for public use:
\begin{equation} \label{eq:MIA_Attack}
d_{\tilde{\mathbf{X}}}(X_0)= \min_{\tilde{X}\in \tilde{\mathbf{X}}}\|X_0-\tilde{X}\|_2.
\end{equation}
The performance of this classifier can be measured by the AUC under the ROC curve.  If the AUC is close to 0.5, then this classifier can't effectively separate fresh observations from old observations in the original dataset, i.e., a good privacy protection.   Based on this observation, for each pair of observed and synthetic datasets in our simulation, we also generate a fresh dataset from the same distribution of observed data with the same sample size and then calculate the AUC under the ROC curve for $d_{\tilde{\mathbf{X}}}(X).$ In particular, the AUC can be calculated as 
\begin{equation} \label{eq:AUC}
\frac{1}{n^2} \sum_{i=1}^n \sum_{j=1}^n \left[ I\left\{ d_{\tilde{\mathbf{X}}}(X_i) < d_{\tilde{\mathbf{X}}}(X^*_j) \right\} + \frac{1}{2} I\left\{ d_{\tilde{\mathbf{X}}}(X_i) = d_{\tilde{\mathbf{X}}}(X^*_j) \right\} \right],
\end{equation}
where the observed data consists of $\{X_1, \cdots, X_n\}$ and freshly generated independent data consists of $\{X^*_1, \cdots, X^*_n\}.$  We use a sample size of $n = 2500$ across $B = 100$ simulation runs. We summarize the AUC for different methods and different $w$ for Latent Noise Injection and Direct Noise Injection methods in Figure \ref{fig:Privacy_Analysis}. Panels $(a)$ and $(b)$ display the AUC-ROC curves for one simulation for illustration, and Panel $(c)$ displays the mean AUC across all $B = 100$ runs as a function of $w\in [0, 1]$ for both Latent and Direct Noise Injection methods. In particular, when $w=0.75$, the mean AUC for Latent Noise Injection is $0.52$, implying a fairly good privacy protection. On the other hand, when $w=0.975$, the mean AUC for Latent Noise Injection method is $0.71$, suggesting loss in privacy protection. 

A justification for why Direct Noise Injection allows for a larger noise weight without immediately compromising privacy protection, as seen in Panel (c) of Figure~\ref{fig:Privacy_Analysis}, is that it perturbs the output data more ``chaotically'' and directly in the observation space. This unstructured distortion makes it more difficult to identify specific individuals, thereby impeding the model’s ability to extract structured, sensitive patterns. However, while this approach may better obfuscate identifiable information, it comes at the cost of degrading the underlying data structure essential for statistical inference.

In contrast, our Latent Noise Injection method strikes a more favorable balance. Specifically, at a noise weight of $w = 0.75$, it achieves two key advantages. First, the resulting synthetic data retains a high degree of utility: estimates of key statistical quantities (e.g., precision or model parameters) are comparable to those obtained using the original data, as seen in Figures \ref{fig:Corr_Diff} and \ref{fig:Corr_Diff_Bias}. Second, privacy protection remains strong, with AUC scores close to random guessing. By comparison, although Direct Noise Injection offers similar privacy at higher noise levels, it fails to preserve inferential fidelity, significantly compromising precision, as seen in Figures \ref{fig:Corr_Diff} and \ref{fig:Corr_Diff_Bias}. Thus, our approach outperforms existing methods by achieving both privacy and utility simultaneously.

Overall, this study provides insights into the efficacy of Normalizing Flows and perturbation-based reconstruction methods in preserving correlation structures in generated data. The key findings suggest Flow Sampling maintains correlation fidelity but does not fully reach the theoretical decay rate. Second, the Latent Noise Injection method improves with increasing reconstruction weight $w$, making it a viable method for structure-preserving data generation. Finally, the Direct Noise Injection method fails to maintain correlation, especially for higher values of $w$.

We note that future work may explore adaptive perturbation strategies or alternative generative models to further enhance correlation preservation.

\subsubsection{Validity of Meta Analysis Results}

\label{sec:Validity_Meta_Analysis}

The goal of this simulation study is to evaluate the ability of MAF to generate synthetic datasets that preserve key statistical properties of real data across multiple studies to permit meta analysis based on synthetic data. Specifically, we assess how well synthetic data can reproduce regression coefficient estimates obtained from ``originally observed'' data using synthetic data generated via the Latent Noise Injection method in light of the findings in Chapter \ref{sec:Real_Data}. This is particularly relevant for generating and sharing synthetic data for subsequent meta analysis with a good privacy protection.

Specifically, for each simulation iteration, we simulate data from $K = 10$ independent studies. In particular, data for study $k$ consists of $n_k$ i.i.d. copies of $(X_1, X_2, X_3, X_4,  Y)$ following a simple linear model:
\[
Y = \beta_{0k} + \beta_{1k} X_1 + \beta_{2k} X_2 + \beta_{3k} X_3 + \beta_{4k} X_4 + \varepsilon, \quad \varepsilon \sim N(0, \sigma^2_\varepsilon)
\]
where $(X_1, X_2, X_3, X_4)' \sim N(0, \mathbf{I}_4),$ and $\sigma^2_\varepsilon = 0.5$. Moreover, the regression coefficients $(\beta_{0k}, \beta_{1k}, \beta_{2k}, \beta_{3k}, \beta_{4k})'$ for each study are drawn from a multivariate Gaussian distribution:
\[ 
\begin{bmatrix} \beta_{0k} \\ \beta_{1k} \\ \beta_{2k} \\ \beta_{3k} \\ \beta_{4k} \end{bmatrix} \sim N\left\{\begin{bmatrix} \beta_0 \\ \beta_1 \\ \beta_2 \\ \beta_3 \\ \beta_4 \end{bmatrix}, \quad \Sigma_\beta\right\}=N\left\{\begin{bmatrix} 0 \\ 1 \\ 1 \\ 1 \\ 1 \end{bmatrix}, \quad   \begin{bmatrix}  10^{-4} & 5  \cdot 10^{-5} & 5 \cdot 10^{-5} & 5 \cdot 10^{-5} & 5 \cdot 10^{-5} \\ 5 \cdot 10^{-5} & 5 \cdot 10^{-3} & 5 \ \cdot 10^{-5} & 5 \cdot 10^{-5} & 5 \cdot 10^{-5} \\ 5 \cdot 10^{-5} & 5 \cdot 10^{-5} & 10^{-4} & 5 \cdot 10^{-5} & 5 \cdot 10^{-5}\\ 5 \cdot 10^{-5} & 5 \cdot 10^{-5} & 5 \cdot 10^{-5} & 10^{-4} & 5 \cdot 10^{-5} \\ 5 \cdot 10^{-5} & 5 \cdot 10^{-5} & 10^{-4} \cdot 10^{-5} & 5 \cdot 10^{-5} & 10^{-4}\end{bmatrix}\right\}.\]

\begin{comment}
N\left\{\begin{bmatrix} 0 \\ 1 \\ 1 \\ 1 \\ 1 \end{bmatrix}, \quad   0.15^2 \times \begin{bmatrix}  1 & 0.5 & 0.5 & 0.5 & 0.5 \\ 0.5 & 1 & 0.5 & 0.5 & 0.5 \\ 0.5 & 0.5 & 1 & 0.5 & 0.5\\ 0.5 & 0.5 & 0.5 & 1 & 0.5 \\ 0.5 & 0.5 & 0.5 & 0.5 & 1\end{bmatrix}\right\}.   
\end{comment}

\begin{figure}[H]

    % Row 1
    \begin{subfigure}{0.32\textwidth}
        \includegraphics[width=\linewidth]{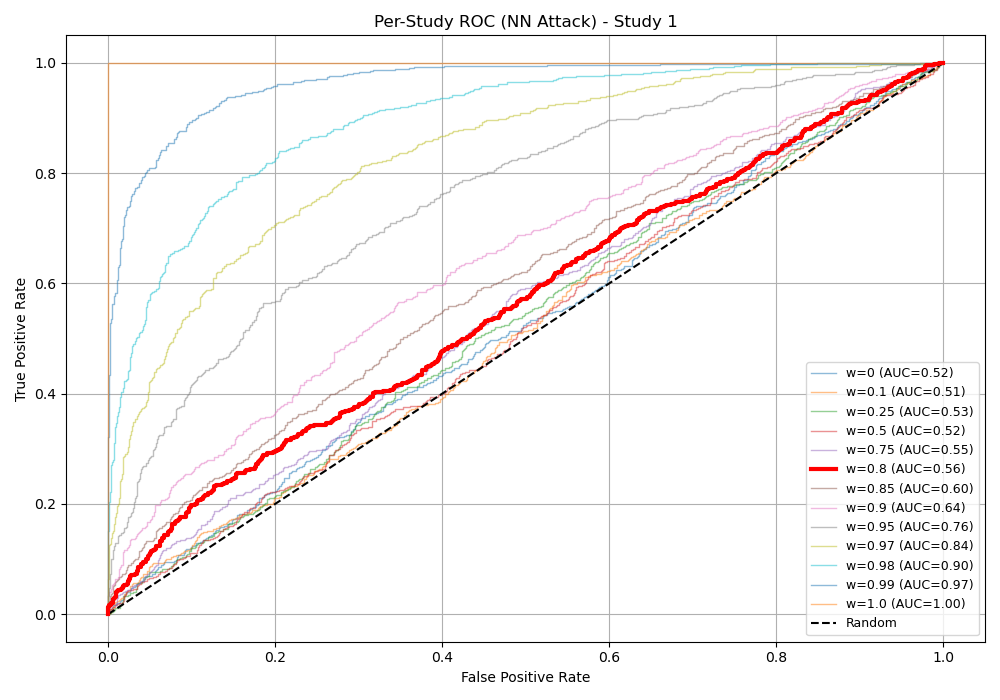}
        \caption{Study 1}
    \end{subfigure}
    \begin{subfigure}{0.32\textwidth}
        \includegraphics[width=\linewidth]{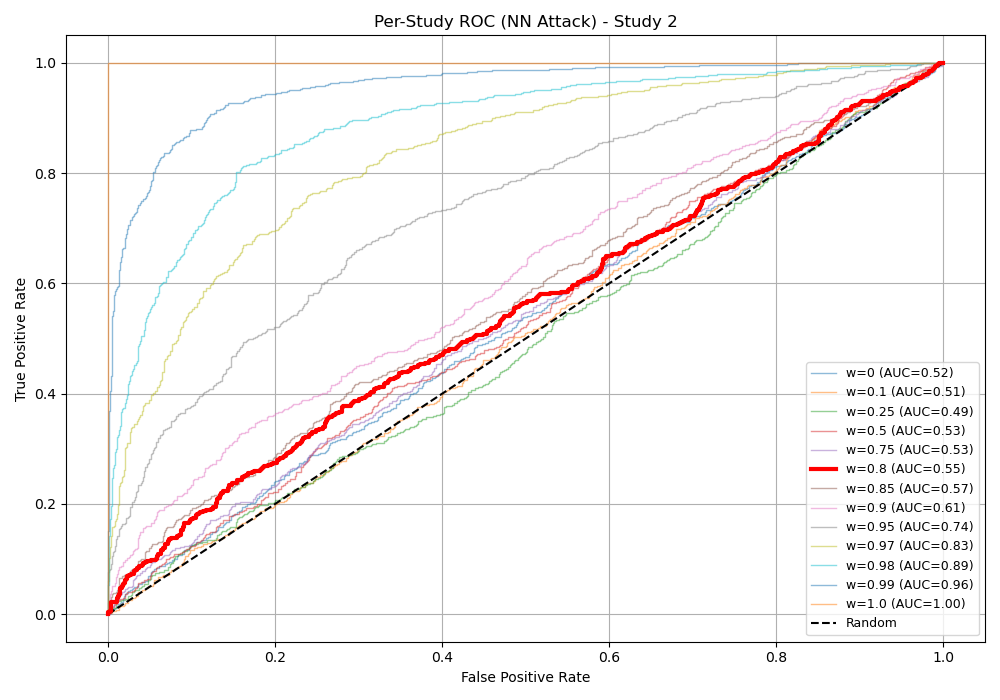}
        \caption{Study 2}
    \end{subfigure}
    \begin{subfigure}{0.32\textwidth}
        \includegraphics[width=\linewidth]{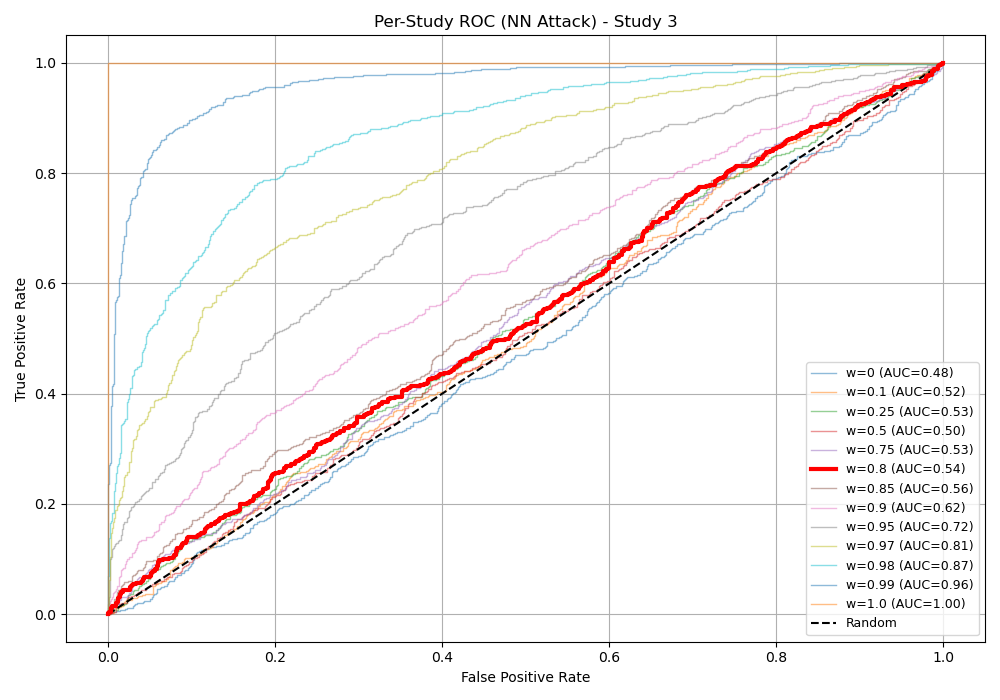}
        \caption{Study 3}
    \end{subfigure}

    % Row 2
    \begin{subfigure}{0.32\textwidth}
        \includegraphics[width=\linewidth]{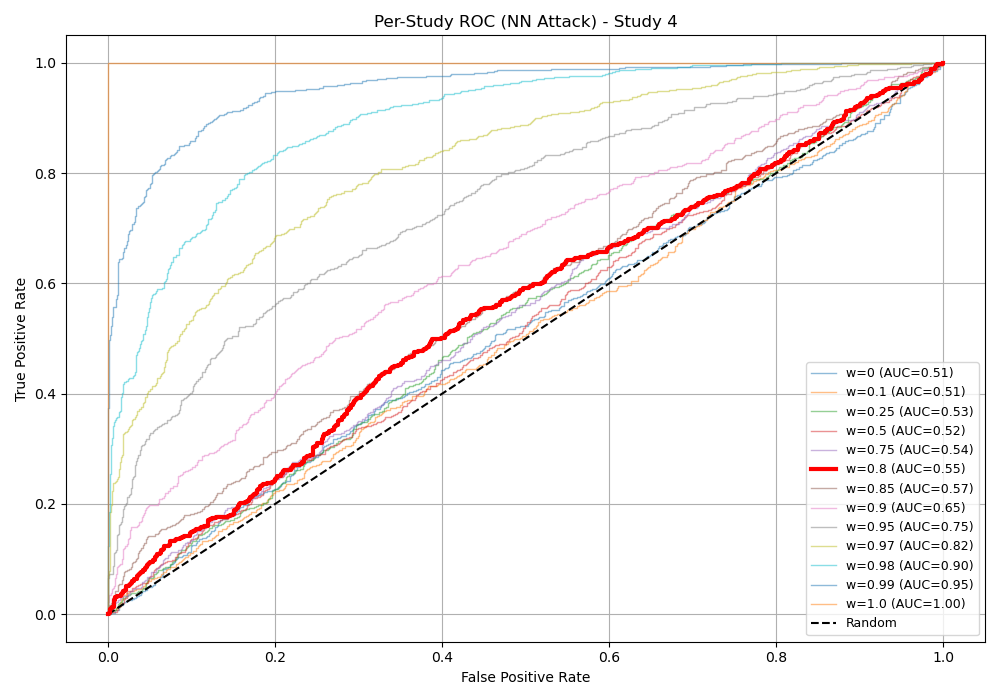}
        \caption{Study 4}
    \end{subfigure}
    \begin{subfigure}{0.32\textwidth}
        \includegraphics[width=\linewidth]{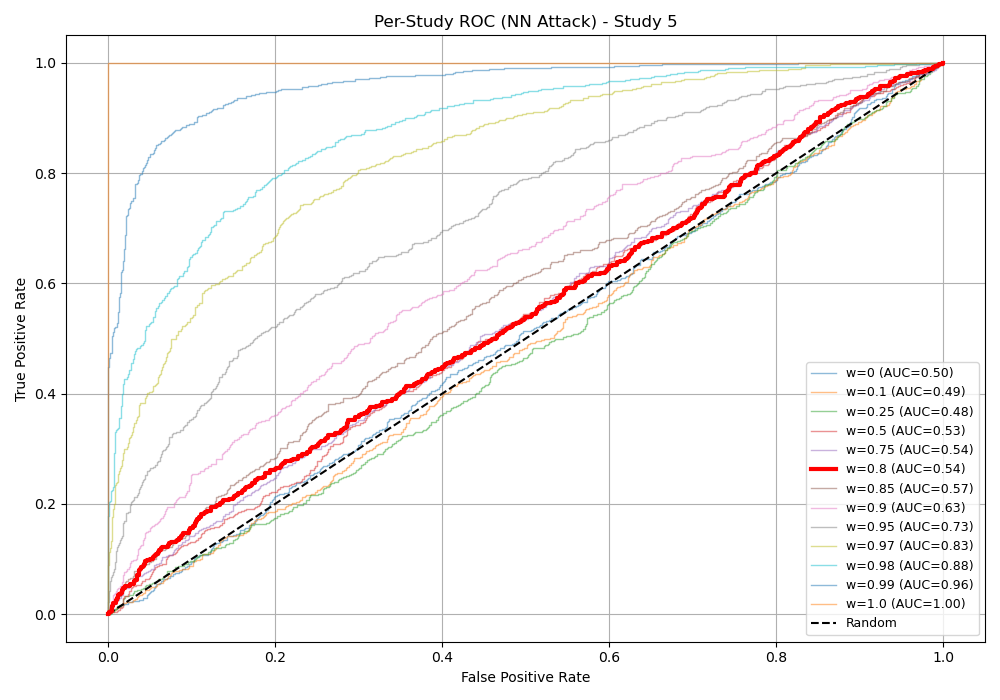}
        \caption{Study 5}
    \end{subfigure}
    \begin{subfigure}{0.32\textwidth}
        \includegraphics[width=\linewidth]{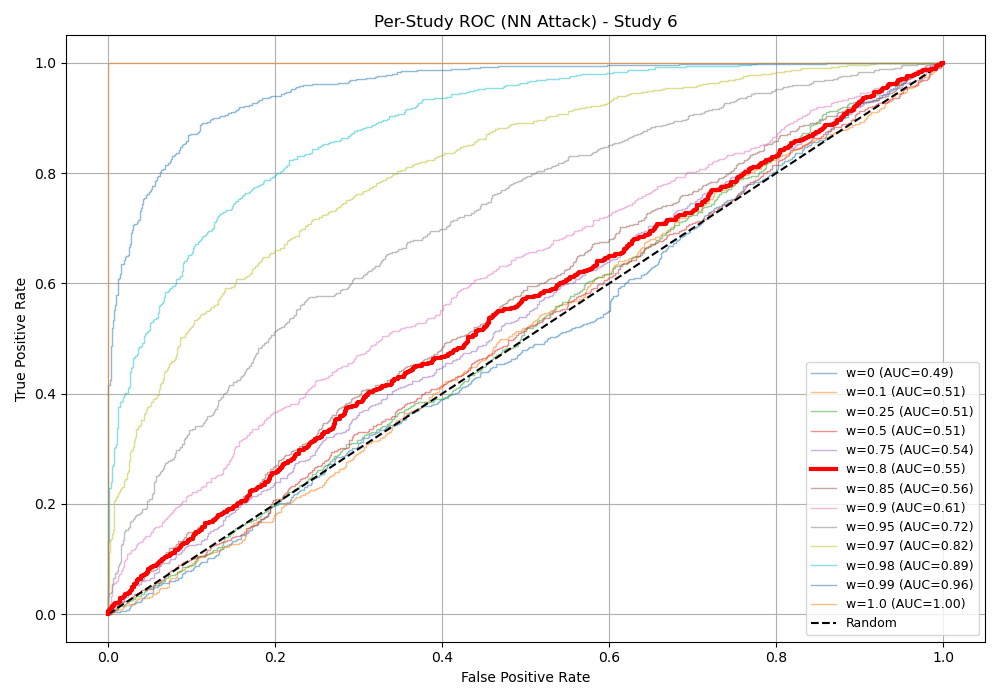}
        \caption{Study 6}
    \end{subfigure}

    % Row 3
    \begin{subfigure}{0.32\textwidth}
        \includegraphics[width=\linewidth]{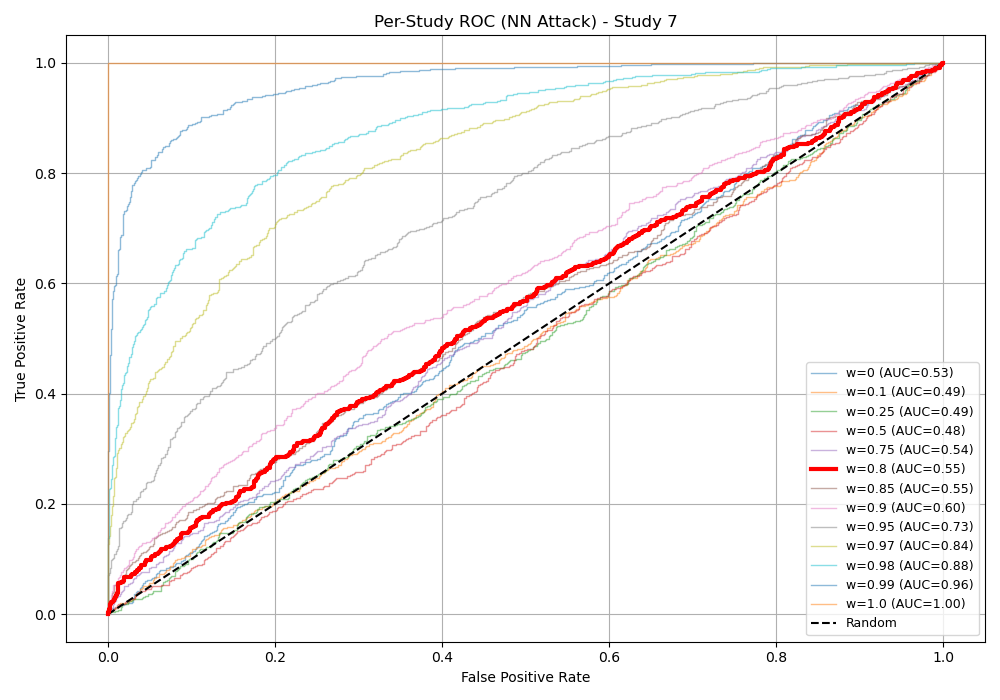}
        \caption{Study 7}
    \end{subfigure}
    \begin{subfigure}{0.32\textwidth}
        \includegraphics[width=\linewidth]{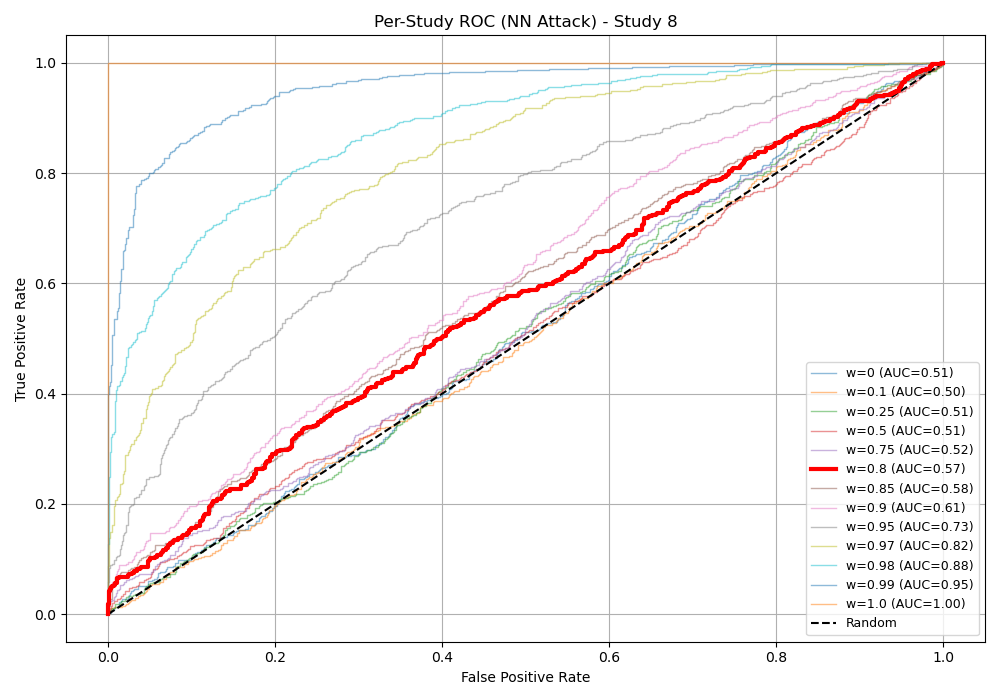}
        \caption{Study 8}
    \end{subfigure}
    \begin{subfigure}{0.32\textwidth}
        \includegraphics[width=\linewidth]{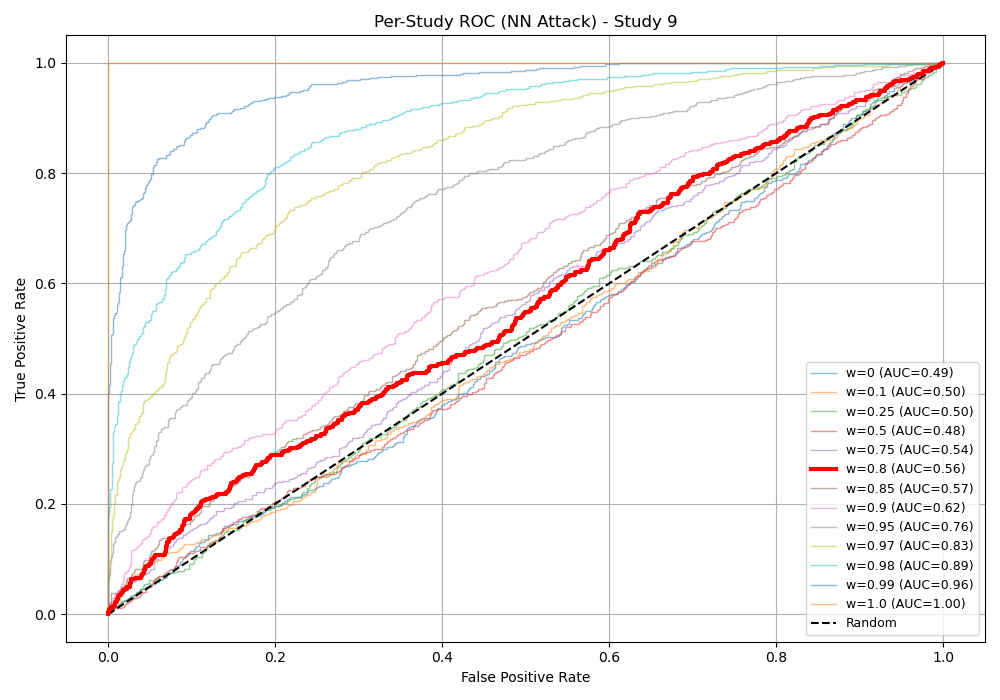}
        \caption{Study 9}
    \end{subfigure}

    \raggedright

    \begin{subfigure}{0.32\textwidth}
        \includegraphics[width=\linewidth]{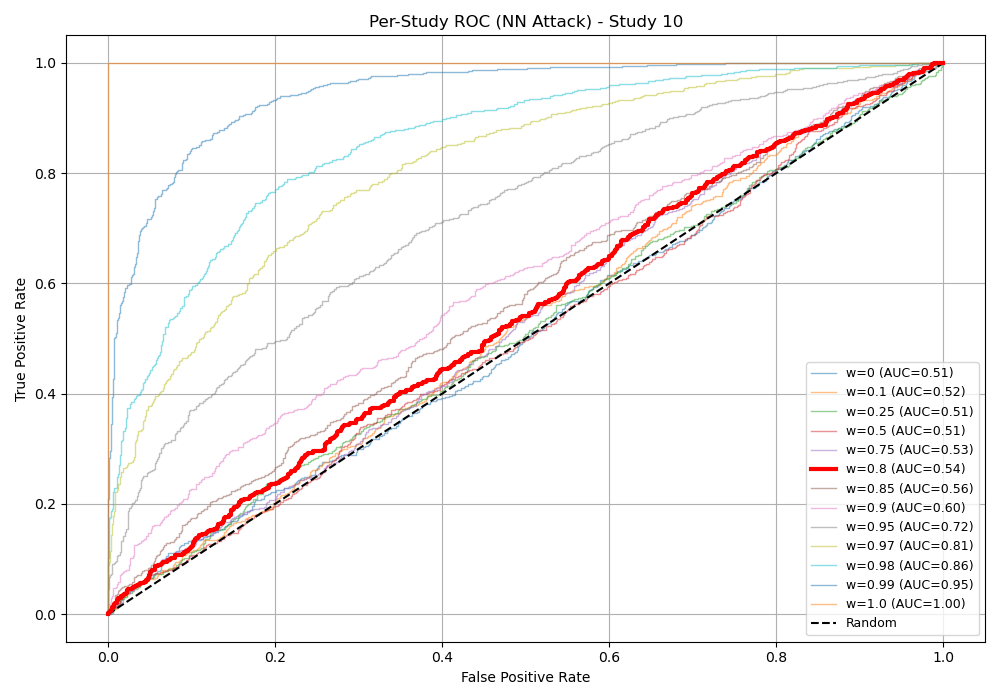}
        \caption{Study 10}
    \end{subfigure}

    \caption{ROC Curves of membership scores in MIA for each study for one simulation iteration. ROC curves are color-coded according to perturbation weight $w$.}

    \label{fig:Per_Study_ROC_Sim}
\end{figure}

The sample size $n_k$ for each study is drawn uniformly from the range $[750, 1000]$. For each study, MAF is trained on observed data. After training, synthetic data is generated by the proposed Latent Noise Injection method using the resulting MAF. 13 synthetic datasets are generated corresponding with $w=\{0, 0.1, \cdots, 1.0\}$, as seen in the legends of Figure \ref{fig:Per_Study_ROC_Sim}, controlling the perturbation size. In generating these $13$ synthetic datasets, the same set of $Z_i\in N(0, \mathbf{I}_5)$ in (\ref{eq:Our_Procedure}) is used to ensure consistency in the evaluating the effect of the perturbation size on statistical analysis based on synthetic data. The parameter of interest is the common regression slope of $X_2$, i.e., $\beta_2=1,$ which can be estimated via meta analysis. Note that we specify the variance of $\beta_{1k}$ to be $5 \cdot 10^{-3}$, compared to $10^{-4}$ for other components $\beta_{jk}, j\neq 1$. In each iteration, we
\begin{enumerate}
    \item generate $K$ observed datasets for $K$ studies;
    \item compute $\hat{\beta}_{2k}$ and its variance $\hat{\sigma}_{2k}^2$ using ordinary least square method for each study based on observed data;
    \item use the DerSimonian-Laird Method from Section \ref{sec:Meta_Analysis} to compute an aggregated estimator of $\beta_2,$ $\hat{\beta}_2,$ and its standard error based on $(\hat{\beta}_{2k}, \hat{\sigma}_{2k}^2), k=1, \cdots, K;$
    \item generate $K$ synthetic datasets for $K$ studies using the Latent Noise Injection method for each $w\in \{0, 0.1, \cdots,  1.0\}$ as seen in the legends of Figure \ref{fig:Per_Study_ROC_Sim};
    \item compute $\tilde{\beta}_{2k}(w)$ and its variance $\tilde{\sigma}_{2k}^2(w)$ using the ordinary least square method for each study based on synthetic data generated with a given $w;$
    \item use the DerSimonian-Laird method to compute a aggregated estimator of $\beta_2,$ $\tilde{\beta}_2(w),$ and its standard error based on $(\tilde{\beta}_{2k}, \tilde{\sigma}_{2k}^2), k=1, \cdots, K.$
\end{enumerate}

We train the MAF based on observed data with a hidden size of $32$ and only one hidden layer. The model is composed of $2$ flow steps and is trained with a $70 \%-30 \%$ train-validation split to select the number of iterations (up to 1000).  Specifically, we continue to train MAF for an additional 100 steps with a learning rate of $10^{-4}$ after the loss in the validation set stops to improve. The process is repeated $B=500$ times. Then, we compare the meta analysis estimator based on observed data with its counterpart based on synthetic data using different $w.$

\begin{figure}[H]
    \centering    \includegraphics[width=0.7\textwidth]{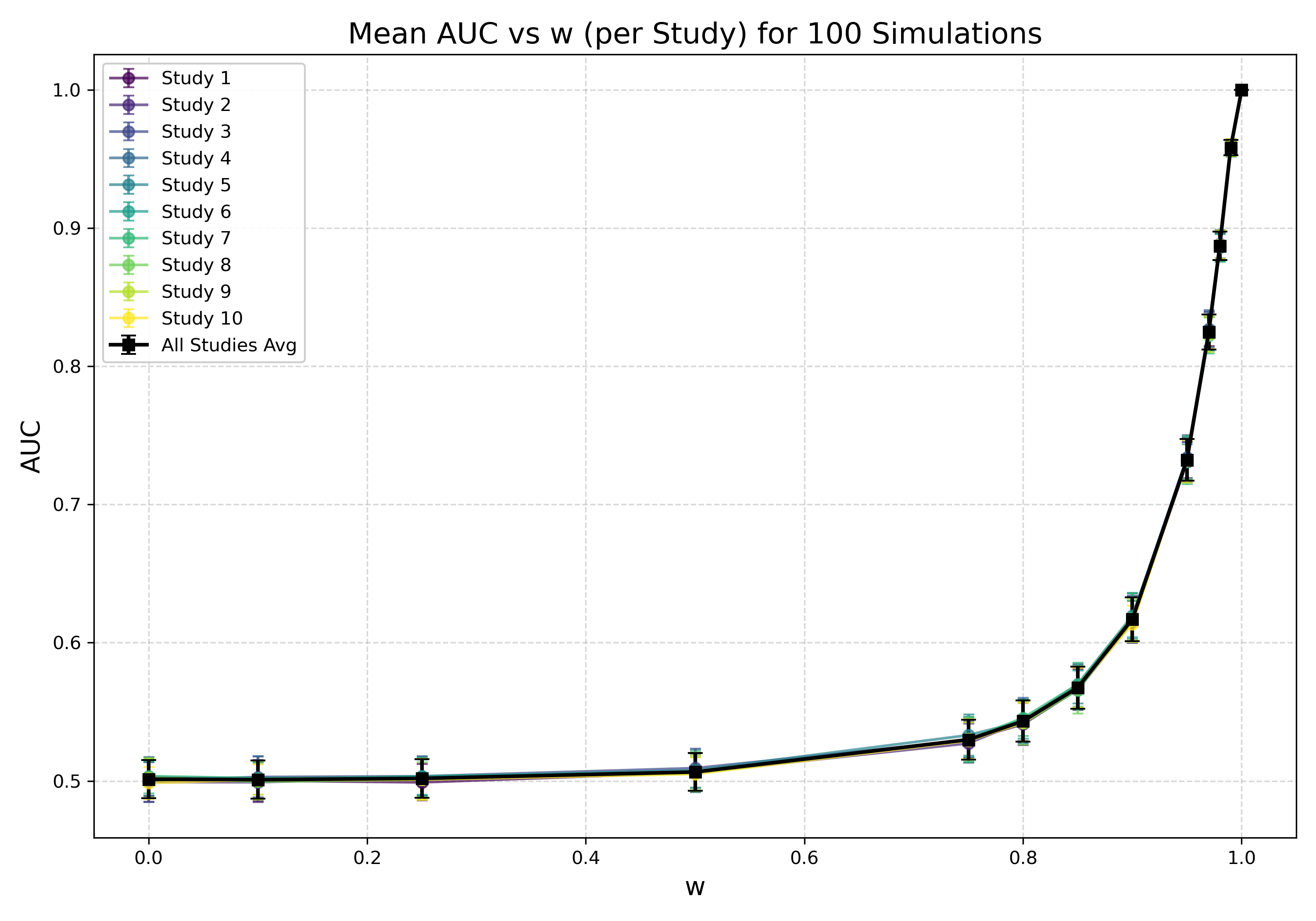}

    \caption{Mean AUC under ROC curve of membership scores in MIA for synthetic data generated with different $w$ across 100 simulations. For each of the ten studies (colored lines), the curve represents the relationship between average AUC and $w$, and the vertical short bars show the mean AUC $\pm $ one standard error.}

    \label{fig:Auc_vs_w}
\end{figure}

The choice of the tuning parameter \( w \) controlling the perturbation size plays a critical role in balancing the trade-off between privacy protection and data fidelity. A lower value of \( w \) typically provides stronger privacy protection, but can come at the cost of reduced inferential performance based on synthetic data. Therefore, selecting an appropriate \( w \) is essential to ensure the synthetic data remains a useful replacement of original data, while limiting potential information leakage. In our simulation study, we assessed the AUC of MIA in a range of \( w \) values for each study with (\ref{eq:MIA_Attack}) and (\ref{eq:AUC}). Using a single simulation realization as an illustrative example, we have reported ROC curves for the generated 10 studies with different $w$ in Figure \ref{fig:Per_Study_ROC_Sim}. We also report the average AUC across $500$ simulation runs for each study and as a function of $w \in [0, 1]$ in Figure \ref{fig:Auc_vs_w}. The ROC curves show that as \( w \) increases, the ability to distinguish between real and synthetic data also increases. In general, \( w = 0.8 \), which is highlighted in bold red in Figure \ref{fig:Per_Study_ROC_Sim}, was found to offer a practical balance, as the average AUC is approximately $55 \%$ across the $10$ studies.

% \begin{figure}[h!]
%     \centering
%     \begin{subfigure}[b]{0.44\textwidth}
%         \includegraphics[width=\textwidth, height=6cm]{Images/Synthetic_vs_Real_Beta1.png}
%         \caption{DerSimonian-Laird Combined from $K = 10$ Studies for Synthetic and Real $\beta_1$}
%         \label{fig:Real_Synthetic}
%     \end{subfigure}
%     \hfill
%     \begin{subfigure}[b]{0.44\textwidth}
%         \includegraphics[width=\textwidth, height=6cm]{Images/Overlayed_Histograms_Beta1_Transparent.png}
%         \caption{Histogram of Overlayed $\beta_1$ }
%         \label{fig:Overlayed_Histograms}
%     \end{subfigure}
% \end{figure}

% \begin{figure}[h!]
%     \centering
%     \includegraphics[width=0.5 \textwidth]{Images/ConfidenceIntervals_Beta1_OneSim.png}
%     \caption{Forest Plot (DerSimonian-Laird) for One Simulation Run}
%     \label{fig:Forest_Plot_One_Sim}
% \end{figure}

\begin{figure}[H]
    \centering
    \includegraphics[width=0.7\textwidth]{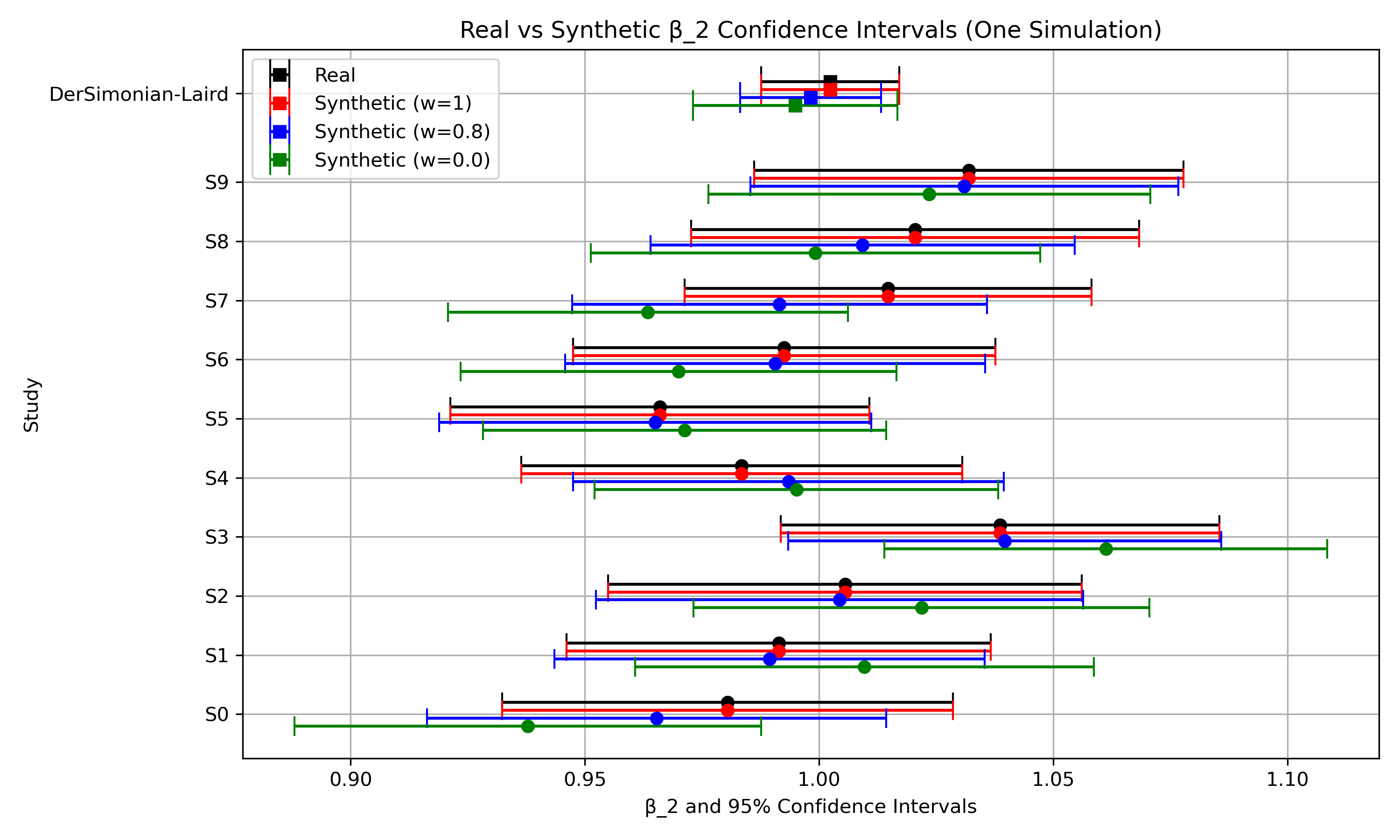}

    \caption{A forest plot for one simulation run. The point estimator and associated 95\% CIs from individual studies as well as the meta analysis are reported; estimates based on synthetic data with different $w$ are colored differently: green ($w=0$), blue ($w=0.8$), and red ($w=1.0$).}
    \label{fig:Forest_Plot_One_Sim}
\end{figure}

\begin{figure}[H]
    \centering
    \includegraphics[width=0.7\textwidth]{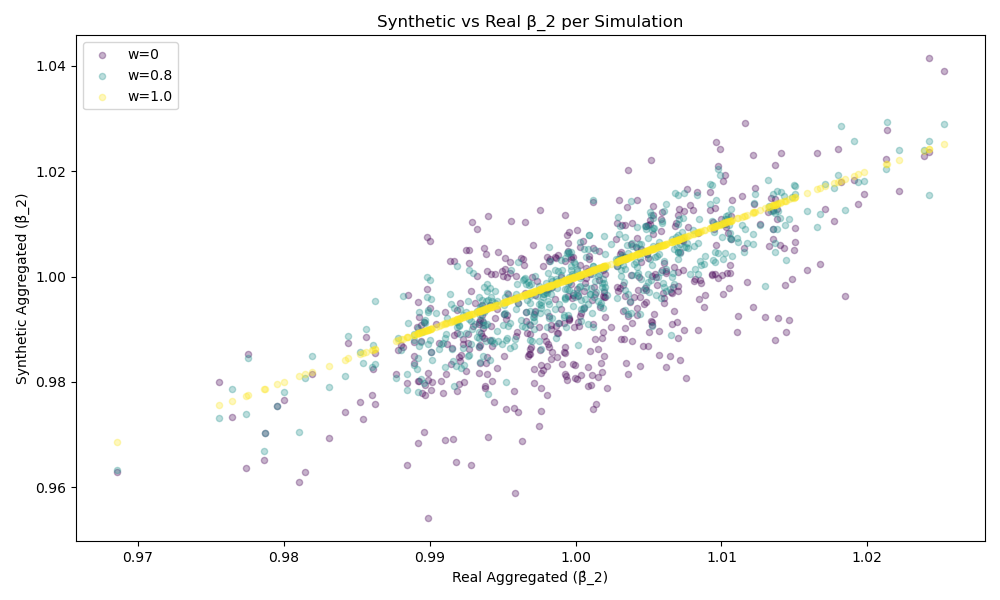}
    \caption{The scatter plot of point estimates of $\beta_2$ from meta analysis based on real vs synthetic data ($500$ simulations); estimates based on synthetic data generated with different $w$ are color-coded differently: purple ($w=0$), teal ($w=0.8$), and
yellow ($w=1.0$).}
    \label{fig:Real_Synthetic}
\end{figure}

\begin{figure}[H]
    \centering
    \includegraphics[width=0.7\textwidth]{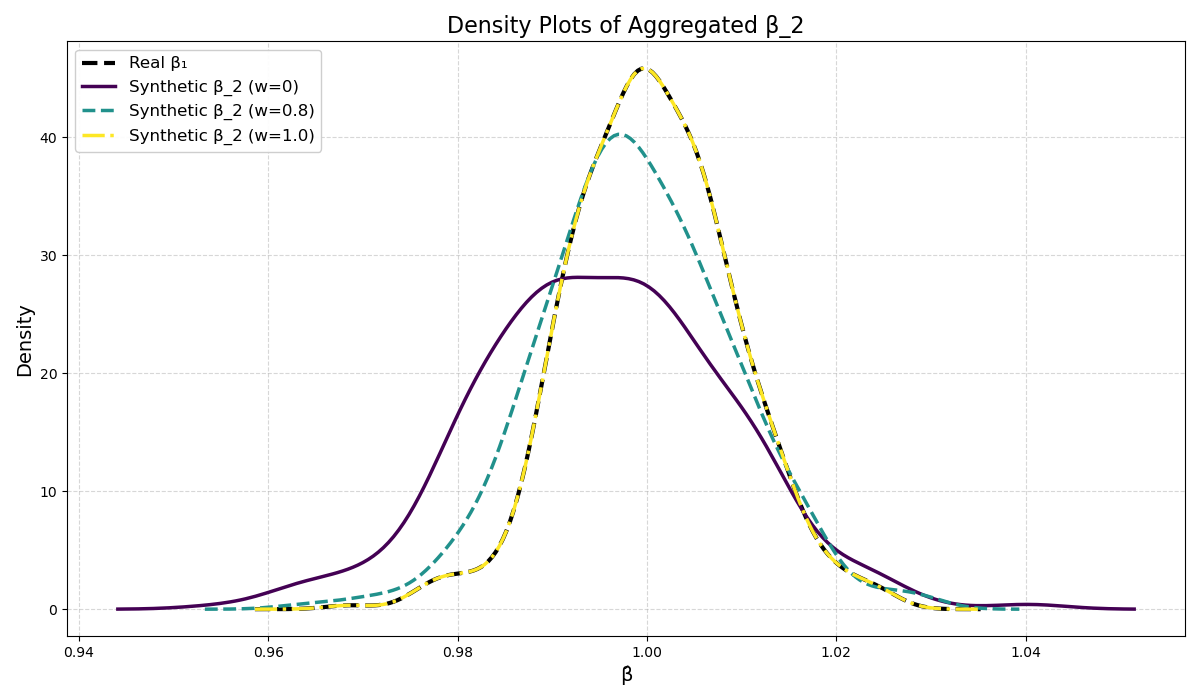}
    \caption{The density estimates of the estimator of $\beta_{2}$ from meta analysis ($500$ simulations); estimates based on synthetic data generated with different $w$ are color-coded differently: purple ($w=0$), teal ($w=0.8$), and yellow ($w=1.0$). The estimates based on original data are colored as black.}
    \label{fig:Overlayed_Densities}
\end{figure}

With \( w = 0.8 \) fixed as our chosen tuning parameter, we next investigate how well synthetic data generated under this choice supports downstream meta analytic inference. The results are summarized in Figures  \ref{fig:Forest_Plot_One_Sim}, \ref{fig:Real_Synthetic}, \ref{fig:Overlayed_Densities}. Figure  \ref{fig:Forest_Plot_One_Sim} shows study level estimates of \(\beta_{2k}\) and their corresponding 95\% CIs as well as the meta analysis estimator of $\beta_2$ and its 95\% CI based on the random effects model from a single simulation run for illustrative purposes. The synthetic data is generated with $w=0, 0.8$ and 1.0. Note, synthetic data generated with $w=0$ does not retain any one to one correspondence at the individual level and offers full privacy protection. On the other hand, synthetic data generated with $w=1$ is identical to the real data, and the results from meta analysis should be identical to that based on real data. This case is presented for a sanity check.  While there are noticeable differences in point estimators and 95\% CIs between observed and synthetic data for some individual studies, e.g., Study $1$, the aggregated results from meta analysis based on synthetic data are very similar to those based on original data. Figure \ref{fig:Real_Synthetic} presents a scatterplot comparing aggregated \(\beta_2\) estimates from observed data with those from synthetic data across $500$ simulations. Note that the strong linear relationship, evident from the tight clustering around the diagonal line, indicates high fidelity between estimates from real and synthetic data in general. When the perturbation size $w$ is close to $1$ ($w=0.8$), the concordance between estimates from original and synthetic data becomes high (i.e., light blue dots in Figure \ref{fig:Real_Synthetic}). The concordance is substantially diluted when the synthetic data is generated with $w=0.$ The MAD between $\hat{\beta}_2$ and $\tilde{\beta}_2(w)$ is $0.0042$ for $w=0.8$ and almost doubled to $0.0089$ for $w=0$. Obviously, when $w=1$,  $\hat{\beta}_2=\tilde{\beta}_2(w)$ as indicated in Figure \ref{fig:Real_Synthetic}, since there is essentially no perturbation. In addition, the MAD between the true parameter $\beta_2 = 1$ and $\tilde{\beta}_2(w)$ based on synthetic data, or $\hat{\beta}_2$ based on original data, is $0.0067$ for $w = 1$ (i.e., no perturbation), $0.0080$ for $w=0.8$, and $0.0111$ for $w=0$, suggesting that the estimation precision for $w=0$ is the poorest and substantially improved after anchoring synthetic data with original data by setting $w=0.8.$ We further plot the densities of $\hat{\beta}_2$ from original data and $\tilde{\beta}_2(w)$ from synthetic data in Figure \ref{fig:Overlayed_Densities}, demonstrating that the variance of the estimator based on synthetic data with a perturbation size of $w=0.8$ is indeed similar to that based on observed data and substantially smaller than the variance of the estimator based on synthetic data with a perturbation size of $w=0.$

In general, these results collectively validate the synthetic data generation pipeline, demonstrating that it reliably reproduces both point estimates and CIs in the meta analysis setting beyond a single study, when $w$ is appropriately selected to balance the privacy-reproducibility trade-off.

\section{Real Data Example}
\label{sec:Real_Data}

We conduct experiments to evaluate the effectiveness of the proposed Latent Noise Injection method in generating synthetic data while preserving statistical and predictive properties of real data. The dataset used consists of common clinical variables including time from baseline to the first cardiovascular disease (CVD) event subjective to right censoring, sex, baseline age, baseline total cholesterol level, baseline high-density lipoprotein cholesterol level (HDL), and baseline systolic blood pressure (SBP) from six cohort studies established for studying risk factors associated with cardiovascular health: Atherosclerosis Risk in Communities study (ARIC) \citep{liao1999arterial, schmidt2005identifying}, Coronary Artery Risk Development in Young Adults (CARDIA) study \citep{carnethon2004risk, friedman1988cardia}, Cardiovascular Health (CHS) study \citep{fried1991cardiovascular}, Multi-Ethnic Study of Atherosclerosis (MESA) study \citep{bild2002multi}, Framingham Offspring study \citep{feinleib1975framingham}, and Jackson Heart study \citep{sempos1999overview}. ARIC, MESA, and the Jackson Heart studies probe novel risk factors for sub-clinical atherosclerosis and their downstream impact on heart failure across diverse U.S.\ populations. CARDIA and CHS follow participants longitudinally to chart how cardiovascular and stroke risks evolve with age. The Framingham Offspring study focuses on the heritable nature of cardiovascular disease across generations.

\subsection{Original Data Analysis}
After excluding participants with missing data, there are a total of $8101$ female and $6762$ male participants in ARIC, $2743$ female and $2297$ male participants in CARDIA, $3254$ female and $2413$ male participants in CHS, $3552$ female and $3176$ male participants in MESA, $1766$ female and $1608$ male participants for Framingham Offspring study, and $1913$ female and $1196$ male participants for Jackson Heart study.  We divide each cohort study into male and female subgroups, yielding a total of $12$ ``sub-studies''. We select age, total cholesterol level, HDL, and SBP as potential CVD risk factors with a special interest in SBP. We have summarized their distribution in 12 sub-studies in Table \ref{tab:Summary_Stats}. The continuous measurements are summarized by mean and standard deviation and the binary variable is summarized by proportion.

\begin{table}[H]
    \small
    \centering
    \caption{The baseline summary statistics for the 12 sub-studies: mean (standard deviation) for continuous measures (time to CVD, Age, total cholesterol level, HDL and SBP) and percentage of participants who experienced a CVD Event.}
    \label{tab:Summary_Stats}
    \begin{tabular}{lccccccc}
        \toprule
        \textbf{Study} & \textbf{$n$} & \textbf{ CVD Event} & \textbf{TIME TO CVD} & \textbf{AGE} & \textbf{CHL} & \textbf{HDL} & \textbf{SBP}\\
         & & (\%) & (year) & (year) & (mg/dL) & (mg/dL) & (mmHg)\\
        \midrule
        ARIC (F)             & 8101  & 16  & 20 (6) & 54 (6) & 218 (43) & 57 (17) & 120 (19)\\
        ARIC (M)             & 6762  & 26  & 18 (7) & 55 (6) & 211 (40) & 44 (14) & 122 (18)\\
        CARDIA (F)           & 2743  & 4   & 33 (4) & 25 (4) & 177 (33) & 56 (13) & 107 (10)\\
        CARDIA (M)           & 2297  & 7   & 32 (6) & 25 (4) & 176 (34) & 50 (13) & 115 (10)\\
        CHS (F)              & 3254  & 42  & 13 (7) & 72 (5) & 221 (39) & 59 (16) & 137 (22)\\
        CHS (M)              & 2413  & 50  & 10 (7) & 73 (6) & 198 (36) & 48 (13) & 136 (21)\\
        MESA (F)             & 3552  & 9   & 14 (4) & 62 (10) & 200 (36) & 56 (15) & 127 (23)\\
        MESA (M)             & 3176  & 13  & 14 (4) & 62 (10) & 188 (35) & 45 (12) & 126 (19)\\
        Framingham (F)       & 1766  & 13  & 27 (7) & 44 (10) & 201 (40) & 54 (13) & 118 (17)\\
        Framingham (M)       & 1608  & 22  & 25 (9) & 44 (10) & 204 (37) & 43 (11) & 126 (16)\\
        Jackson Heart (F)    & 1913  & 4   & 10 (2) & 50 (12) & 197 (38) & 54 (14) & 124 (17)\\
        Jackson Heart (M)    & 1196  & 5   & 9 (2)  & 49 (12) & 197 (40) & 45 (12) & 126 (16)\\
        \bottomrule
    \end{tabular}
\end{table}

The objective of the statistical analysis is to examine the associations between SBP level at the baseline and the hazard of subsequent CVD events. To this end, we fit two Cox regression models on data from each sub-study and obtain the estimated regression coefficient of SBP and the corresponding CI: one without any adjustment and one with adjustment of age, total cholesterol level, and HDL. Then, we aggregate results from all 12 sub-studies into one point estimator with a CI for the hazard ratio (HR) associated with a change of 10 mmHg in SBP using a random effects model-based meta analysis (\ref{eq:RandomEffect}). In particular, the Dersimonian-Laird Method to aggregate information from all 12 sub-studies is used. The results are reported in Figure \ref{fig:SBP_Meta}.  
Without any adjustment of potential confounders, the estimated HR of 10 mmHg increase in SBP is 1.277 (95\% CI: 1.221 to 1.333). After adjusting for age, total cholesterol level, and HDL, the estimated hazard ratio is 1.171 (95\% CI: 1.134 to 1.207), suggesting that a higher SBP is statistically significantly associated with increased CVD risk. Specifically, every 10 mmHg higher in SBP corresponds to a $17.1\%$ increase in CVD hazard after adjusting for other common CVD risk factors.

\subsection{MAF Training}
Suppose that individual level data from the aforementioned studies cannot be shared due to privacy concerns. We plan to generate synthetic data using the proposed Latent Noise Injection method and conduct meta analysis based on generated data. To this end, we need to first train MAF as a generative model for each sub-study. Before MAF training, we transform the event time using a scaled min-max logit transformation since event time always falls into a range determined by the maximum study follow-up time. For binary censoring indicator, we continualize it by adding an independent noise from \(U(0, 1)\) to their original binary values. Moreover, all other variables were standardized by applying $Z$-score normalization to improve training stability. After training MAF, all the variables were transformed back to their original scales by inverting the standardization steps when generating the final synthetic data. Specifically, for censoring indicator, this means values less than \(0\) are set to \(0\), and values greater than or equal to \(0\) are set to \(1\). All training is conducted with PyTorch as the deep learning framework. We apply an initial learning rate of $10^{-4}$ with a minimum threshold of $10^{-7}$. In addition, we tune our MAF to select the ``best" model configuration by optimizing the validation loss using 80\% vs 20\% cross-validation across multiple candidate configurations specified apriori for each sub-study. Specifically, we perform a grid search over the following hyperparameters: hidden layer size, number of hidden layers, and number of Normalizing Flow layers. The parameter grids varied by study according to respective sample size, and the final selected configuration is reported in Table \ref{tab:MAFtuning}.

\subsection{Selection of Perturbation Size}
\begin{figure}[H]
    \centering

    \begin{subfigure}[b]{0.31\textwidth}
        \includegraphics[width=\linewidth]{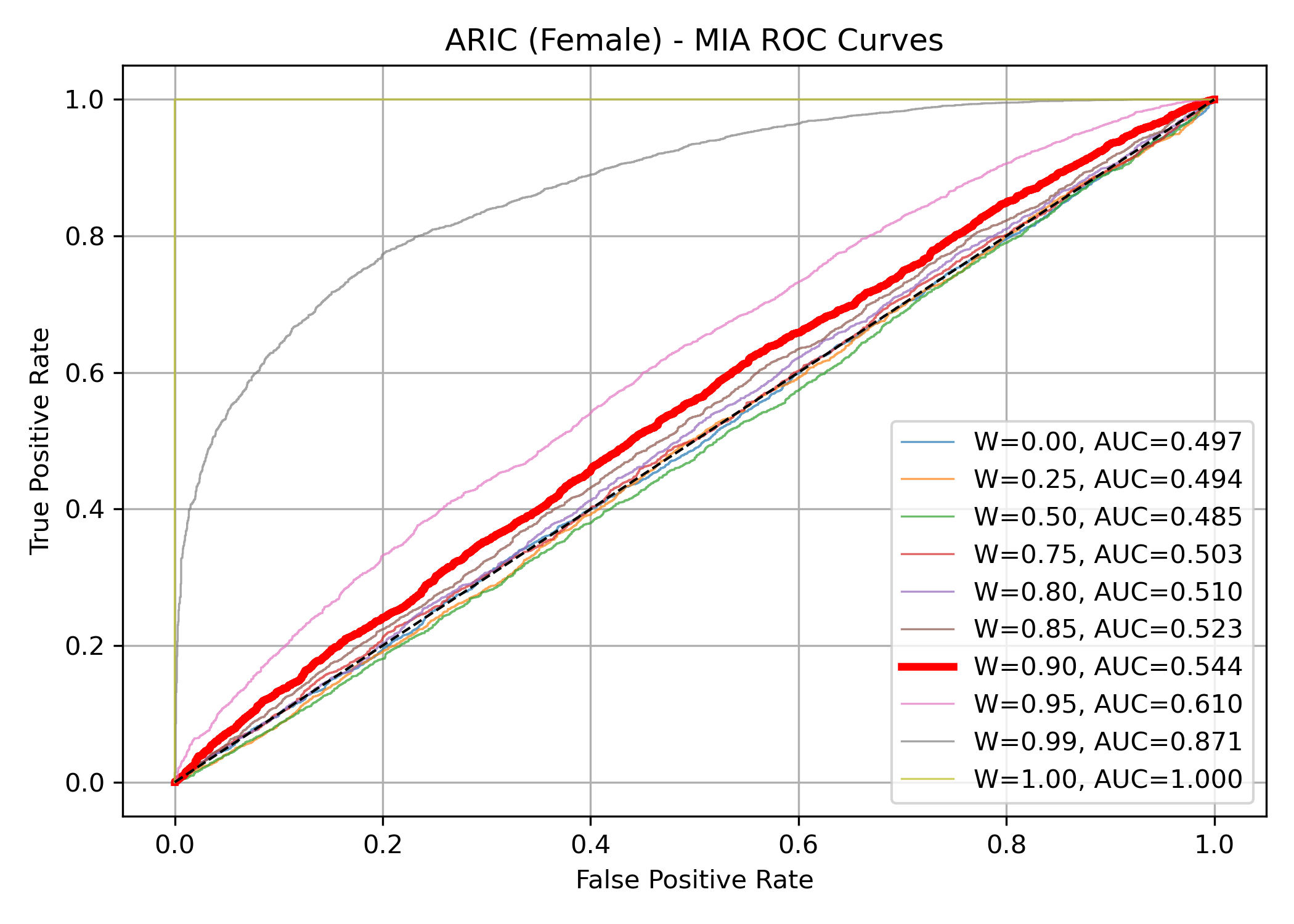}
        \caption{ARIC (Female)}
    \end{subfigure}
    \hfill
    \begin{subfigure}[b]{0.31\textwidth}
        \includegraphics[width=\linewidth]{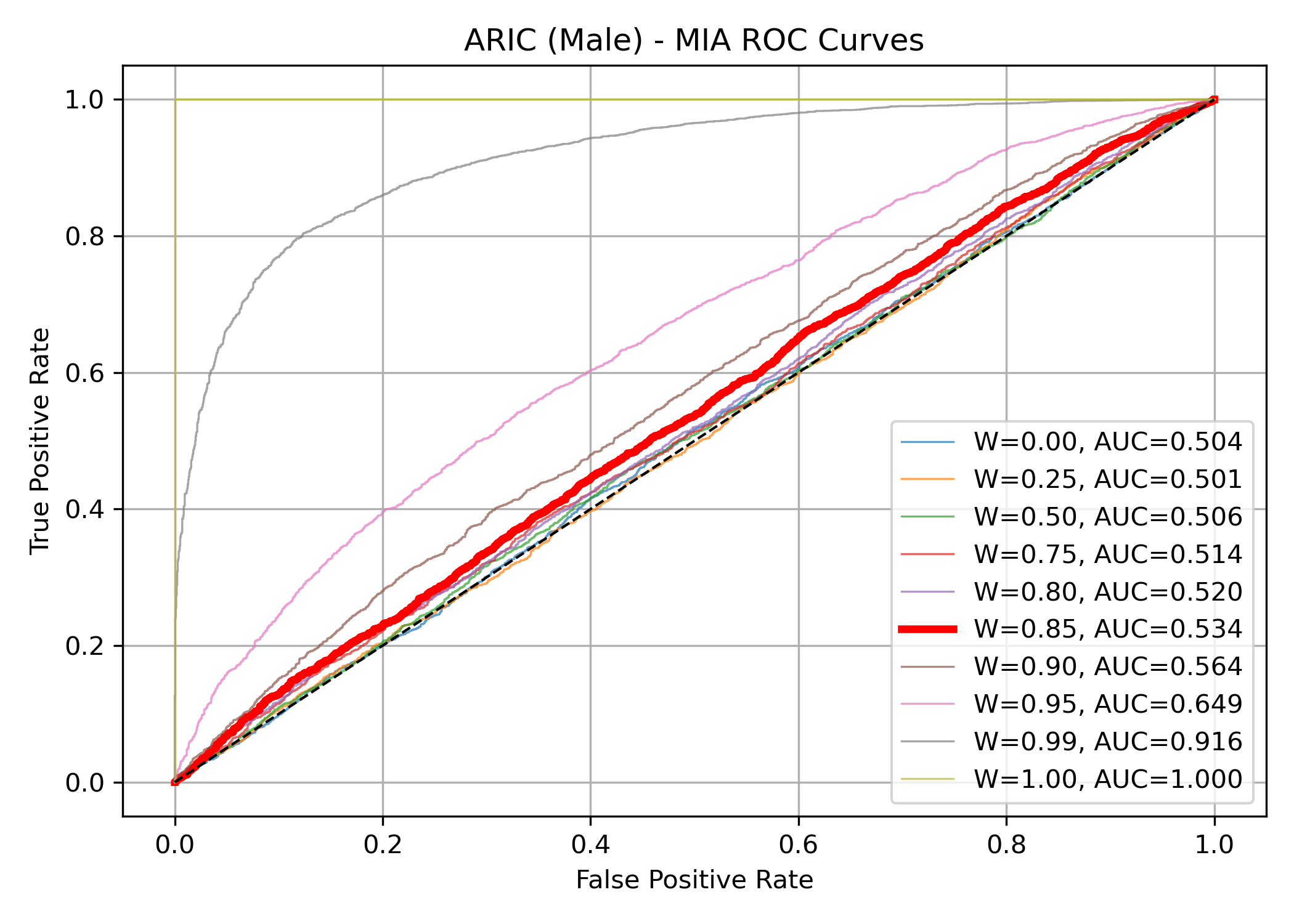}
        \caption{ARIC (Male)}
    \end{subfigure}
    \hfill
    \begin{subfigure}[b]{0.31\textwidth}
        \includegraphics[width=\linewidth]{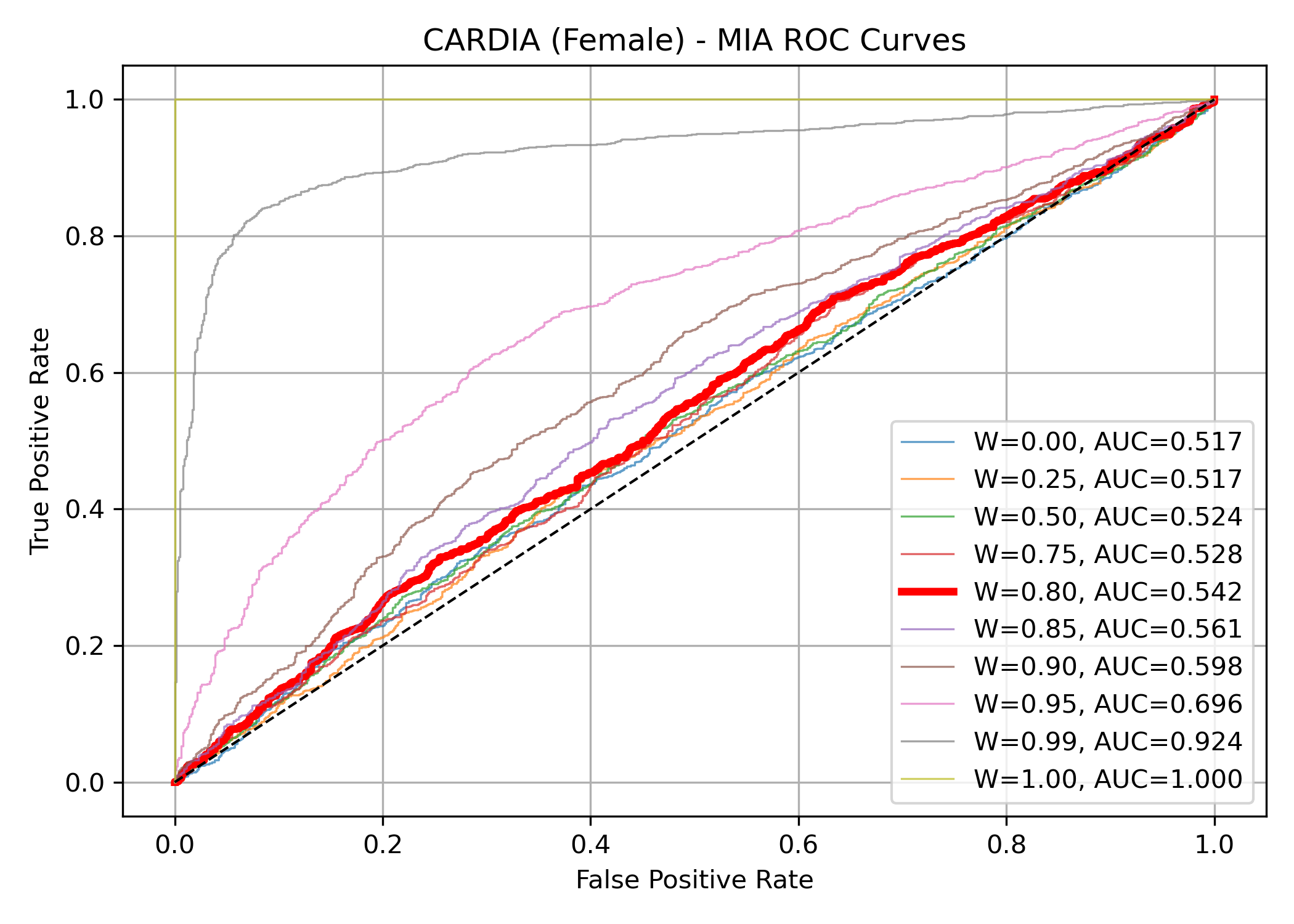}
        \caption{CARDIA (Female)}
    \end{subfigure}

    \vspace{0.5cm}

    \begin{subfigure}[b]{0.31\textwidth}
        \includegraphics[width=\linewidth]{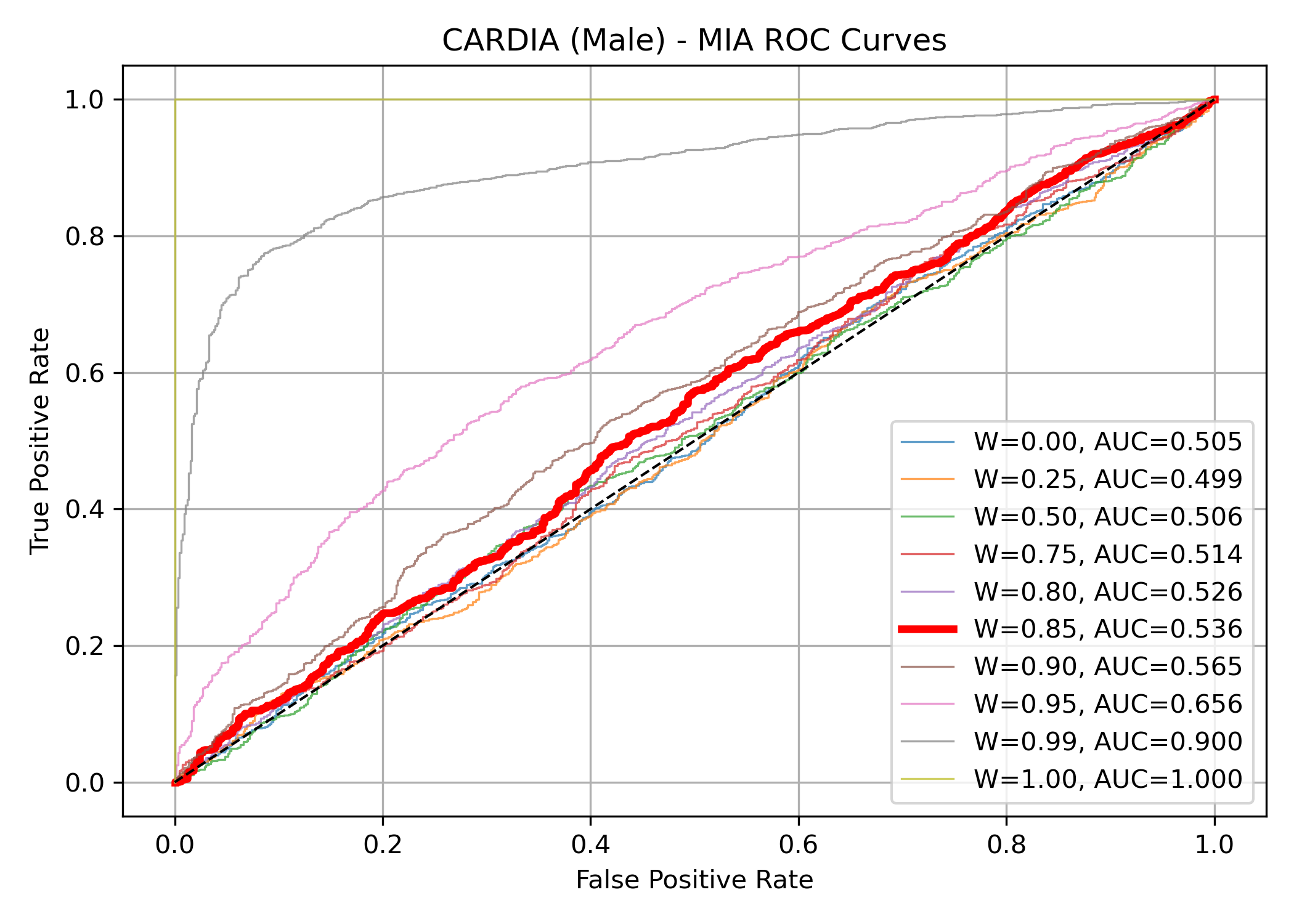}
        \caption{CARDIA (Male)}
    \end{subfigure}
    \hfill
    \begin{subfigure}[b]{0.31\textwidth}
        \includegraphics[width=\linewidth]{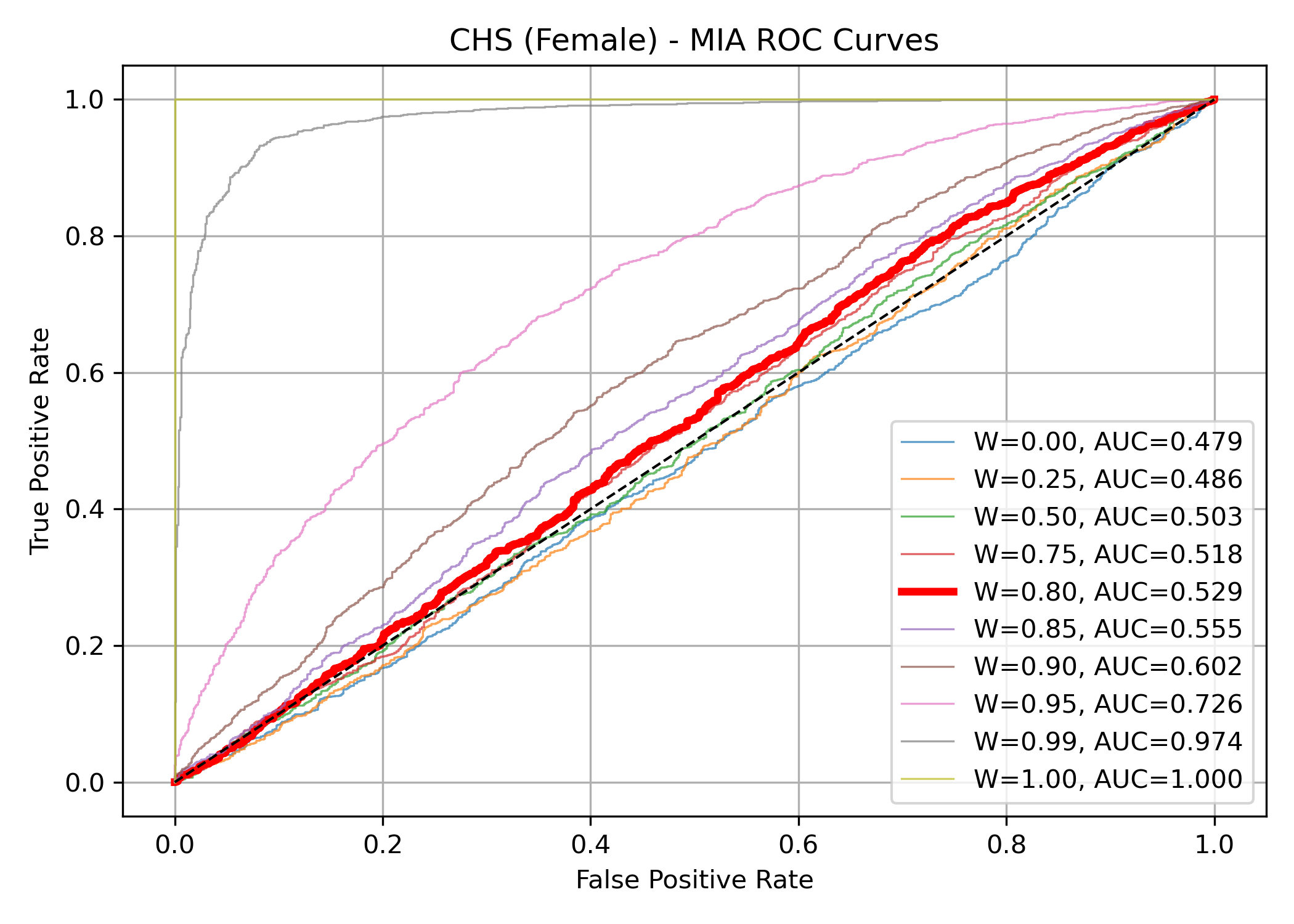}
        \caption{CHS (Female)}
    \end{subfigure}
    \hfill
    \begin{subfigure}[b]{0.31\textwidth}
        \includegraphics[width=\linewidth]{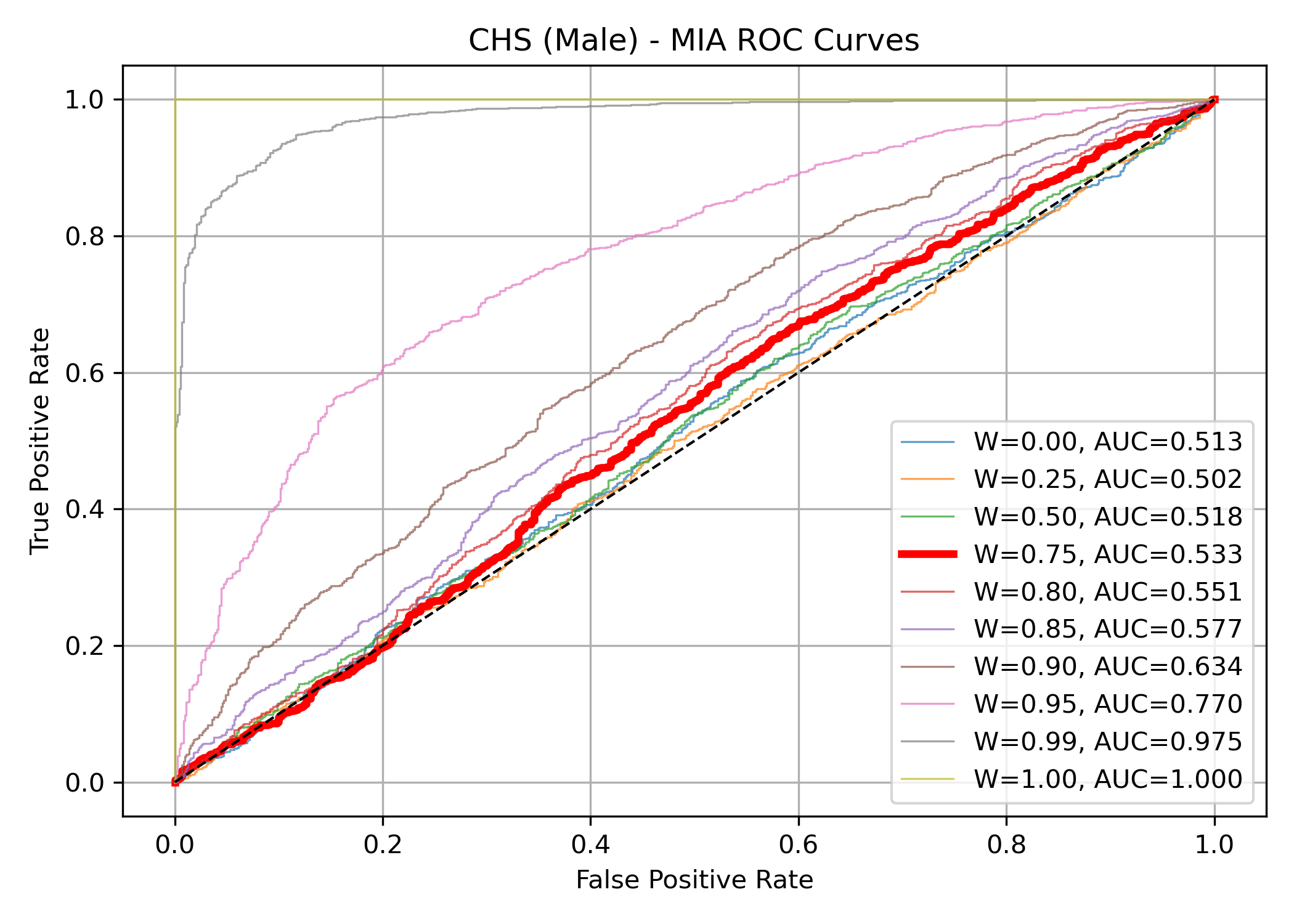}
        \caption{CHS (Male)}
    \end{subfigure}

    \vspace{0.5cm}

    \begin{subfigure}[b]{0.31\textwidth}
        \includegraphics[width=\linewidth]{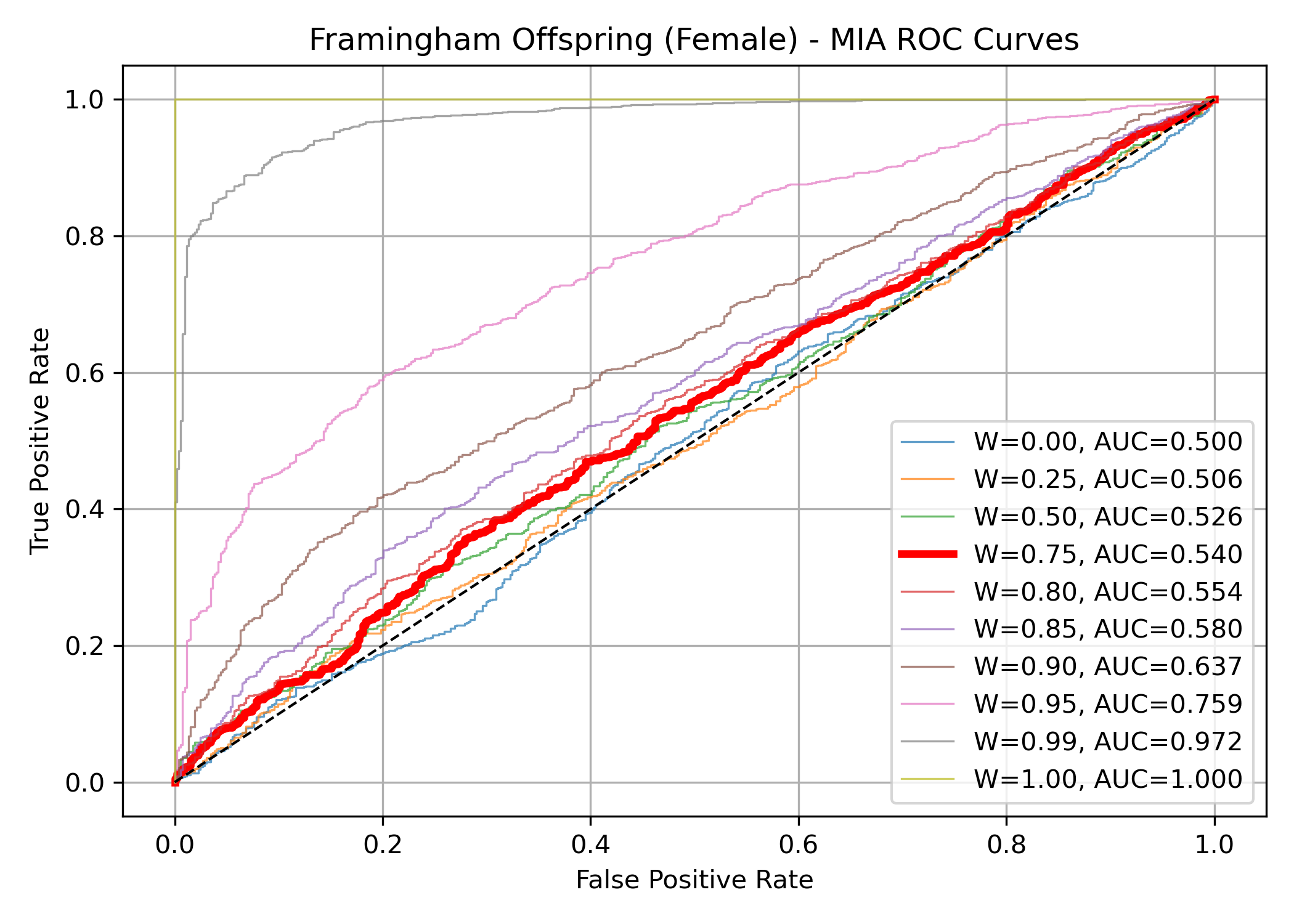}
        \caption{Framingham (Female)}
    \end{subfigure}
    \hfill
    \begin{subfigure}[b]{0.31\textwidth}
        \includegraphics[width=\linewidth]{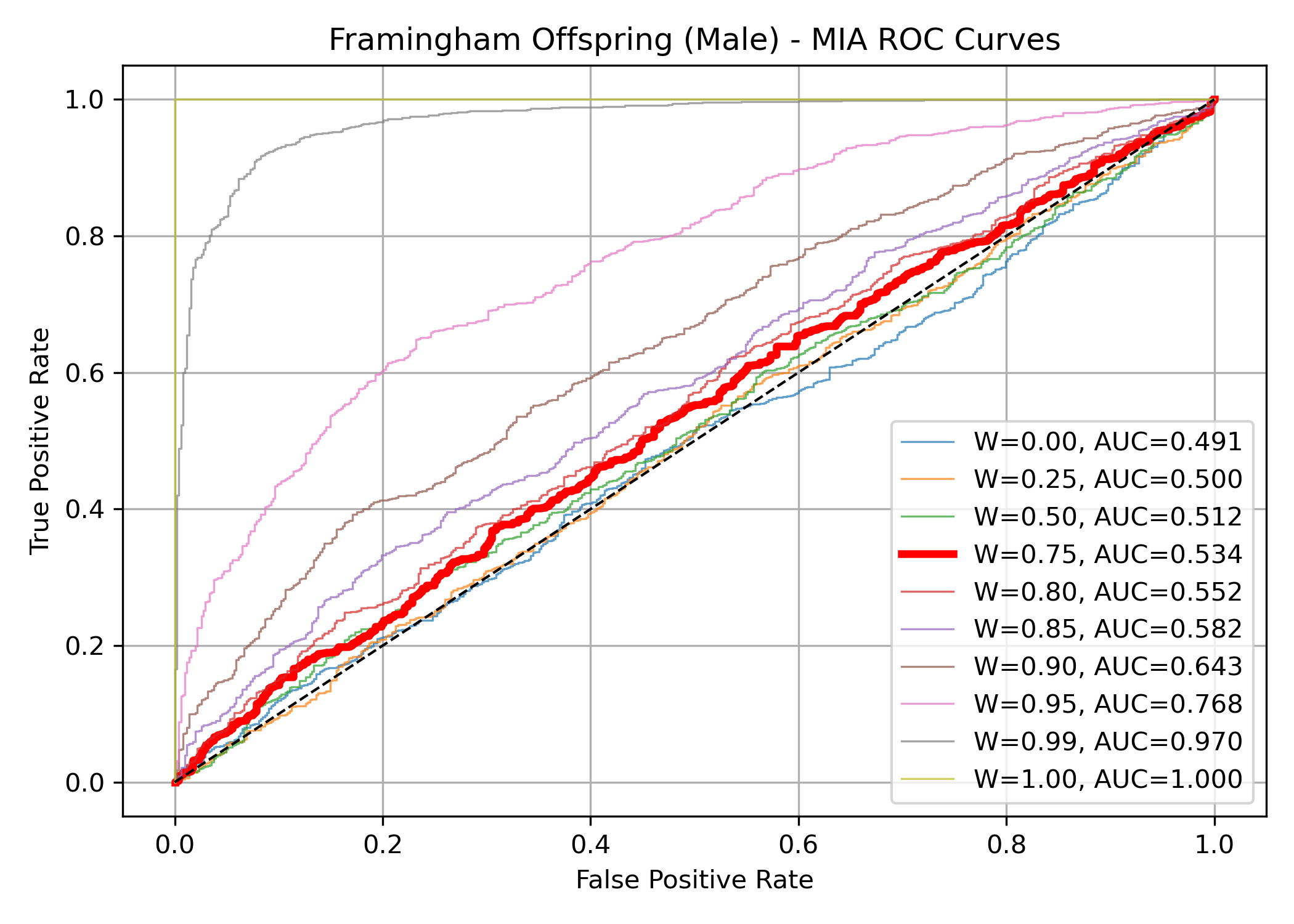}
        \caption{Framingham (Male)}
    \end{subfigure}
    \hfill
    \begin{subfigure}[b]{0.31\textwidth}
        \includegraphics[width=\linewidth]{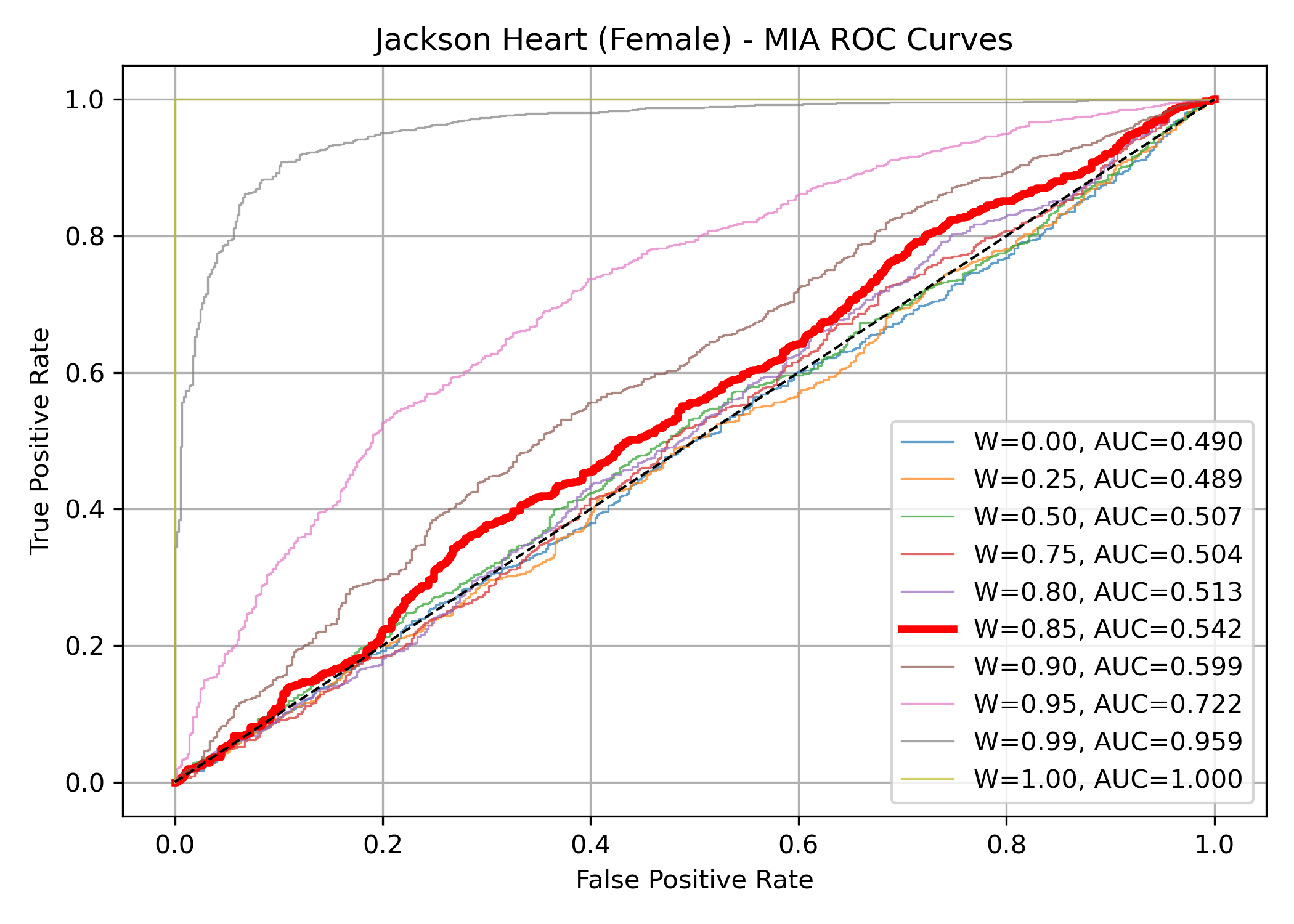}
        \caption{Jackson Heart (Female)}
    \end{subfigure}

    \vspace{0.5cm}

    \begin{subfigure}[b]{0.31\textwidth}
        \includegraphics[width=\linewidth]{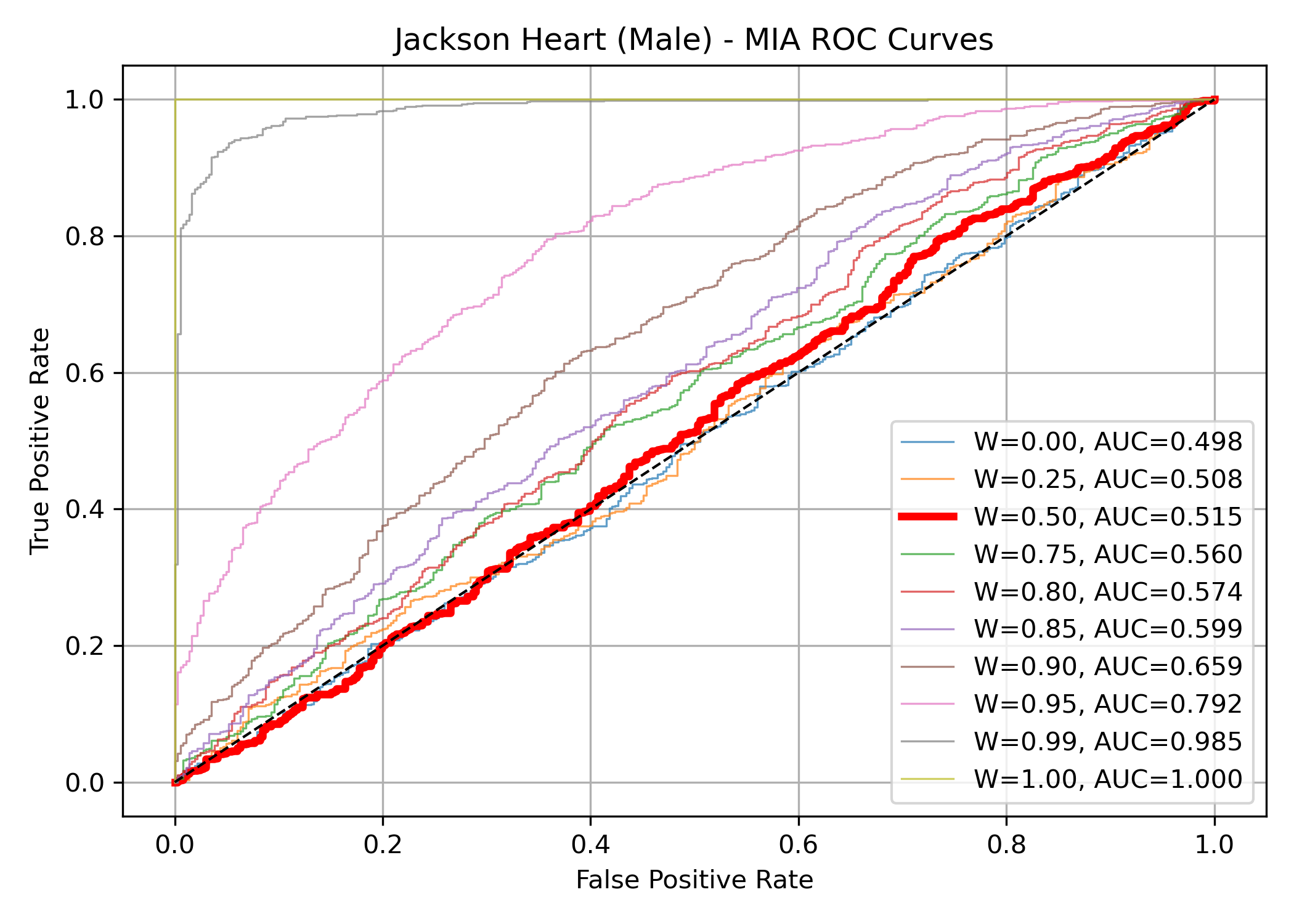}
        \caption{Jackson Heart (Male)}
    \end{subfigure}
    \hfill
    \begin{subfigure}[b]{0.31\textwidth}
        \includegraphics[width=\linewidth]{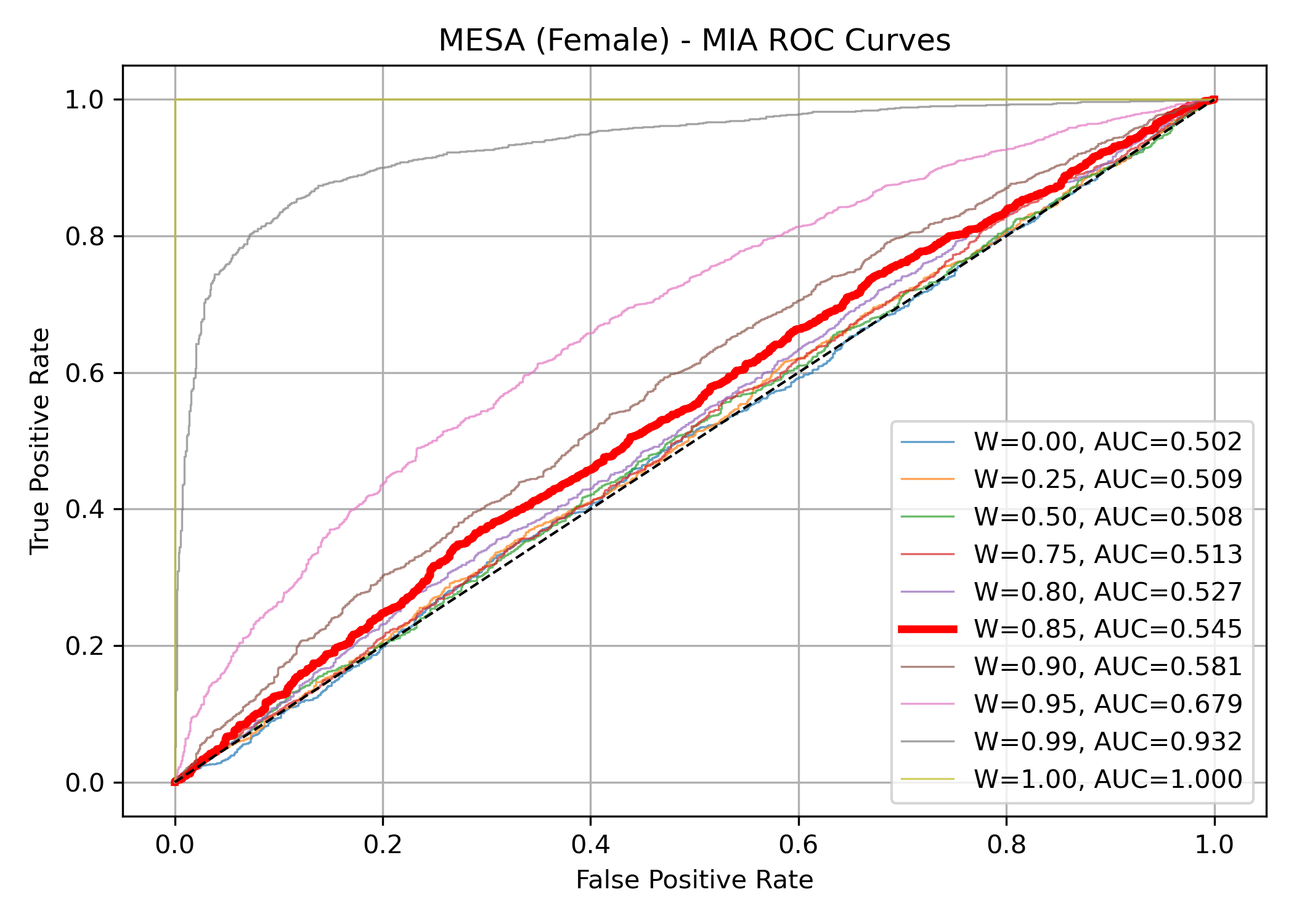}
        \caption{MESA (Female)}
    \end{subfigure}
    \hfill
    \begin{subfigure}[b]{0.31\textwidth}
        \includegraphics[width=\linewidth]{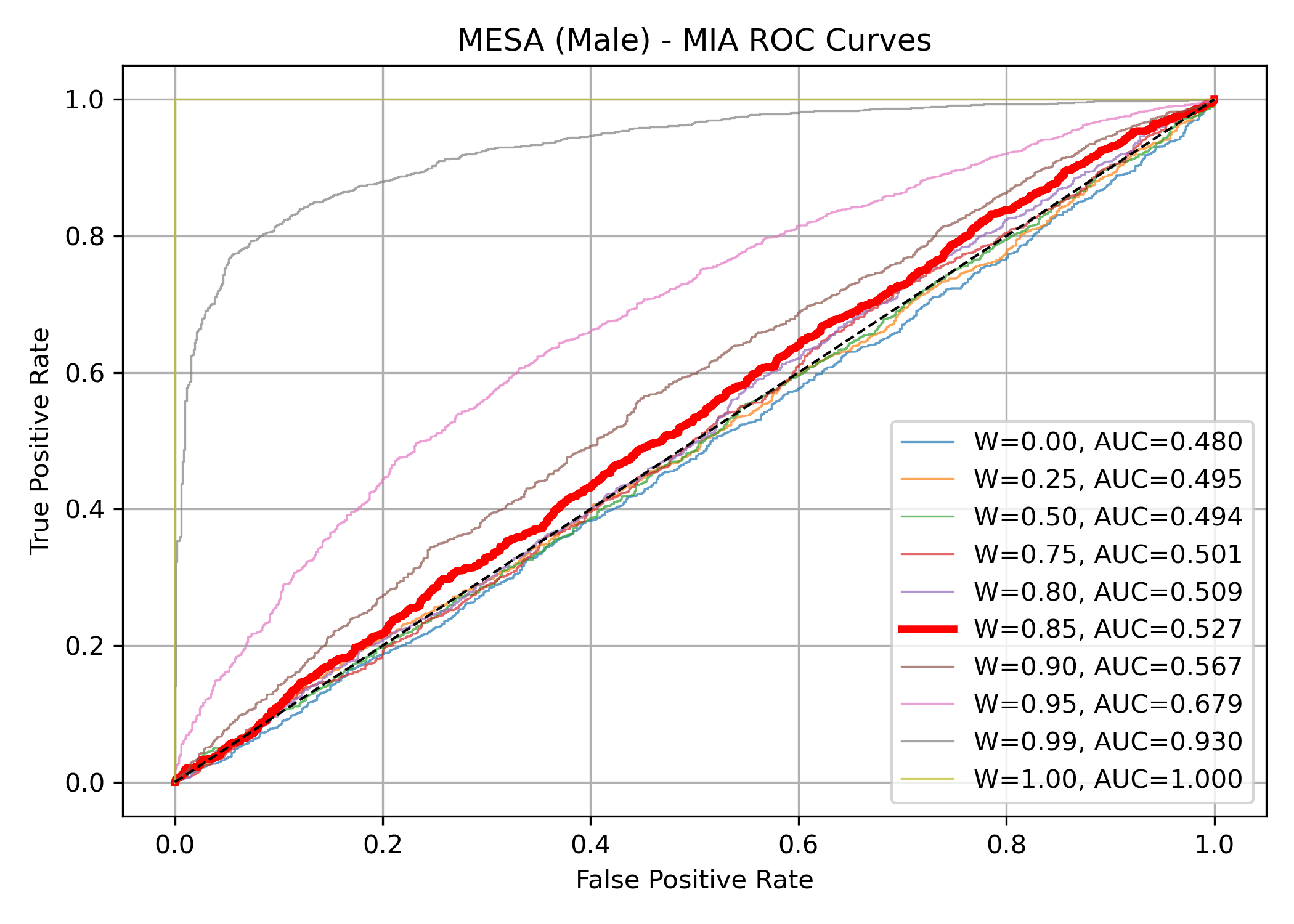}
        \caption{MESA (Male)}
    \end{subfigure}

    \caption{ROC curves of membership scores in MIA for different $w$ in each sub-study.  The red ROC curve corresponds to the selected perturbation size $\hat{w}$.}
    \label{fig:Study_Mia_Rocs}
\end{figure}

Once the MAF is trained for each sub-study, we need to determine the optimal perturbation size $w$ in our Latent Noise Injection method.  To select a $w$ to ensure sufficient privacy protection, we assess empirical privacy leakage through MIA described in Section \ref{s:Simulation_Experiment_1} with different $w$ values for each sub-study.  Unfortunately, we don't have access to a fresh new dataset. Thus, we randomly split observed data into two sets: $D_1,$ a training set consisting of 80\% of the original data, and $D_0,$ a test set consisting of 20\% of the original data. We then train a MAF based on the training set $D_1$  and use the Latent Noise Injection method with trained MAF and given perturbation size $w$ to generate a synthetic dataset $\tilde{\cal D}_1$ by perturbing $D_1$. As discussed in Section \ref{s:Simulation_Experiment_1}, for each $X \in D_1 \cup D_0$, a membership score $d_{\tilde{D}_1}(X)$ is calculated. We then compute the AUC under the ROC curve of the membership score $d_{\tilde{D}_1}(X)$ in classifying  $X \in D_1$ vs $X\in D_0$. AUC values near $0.5$ indicate that synthetic data does not leak training samples and provides empirical support for privacy preservation. We select the largest value of the perturbation parameter \( w \in (0,1) \) such that the MIA AUC remains below $0.55$. The selected perturbation size, $\hat{w},$ is different for each sub-study and reported in Table \ref{tab:Best_w_Per_Study}. This criterion balances the tradeoff between fidelity and privacy: as \( w \to 1 \), the synthetic output becomes more faithful to the original data (but may risk memorization of individual observations), whereas as \( w \to 0 \), the output resembles purely sampled noise from the latent prior, offering stronger privacy but potentially lower statistical utility. Figure \ref{fig:Study_Mia_Rocs} displays sub-study specific ROC curves for varying $w$ value. The AUC values remain near $0.5$ when \( w \) is small, indicating no detectable memorization. However, as \( w \) increases beyond a threshold (i.e., between 0.7 and 0.9), AUC values rise sharply to $1.0$, which signals potential privacy leakage due to inadequate perturbation. This empirical trend is consistent across studies and reinforces the need to calibrate \( w \) based on an acceptable privacy-utility tradeoff. 

\begin{figure}[H]
    \centering

    \begin{subfigure}[b]{0.31\textwidth}
        \includegraphics[width=\linewidth]{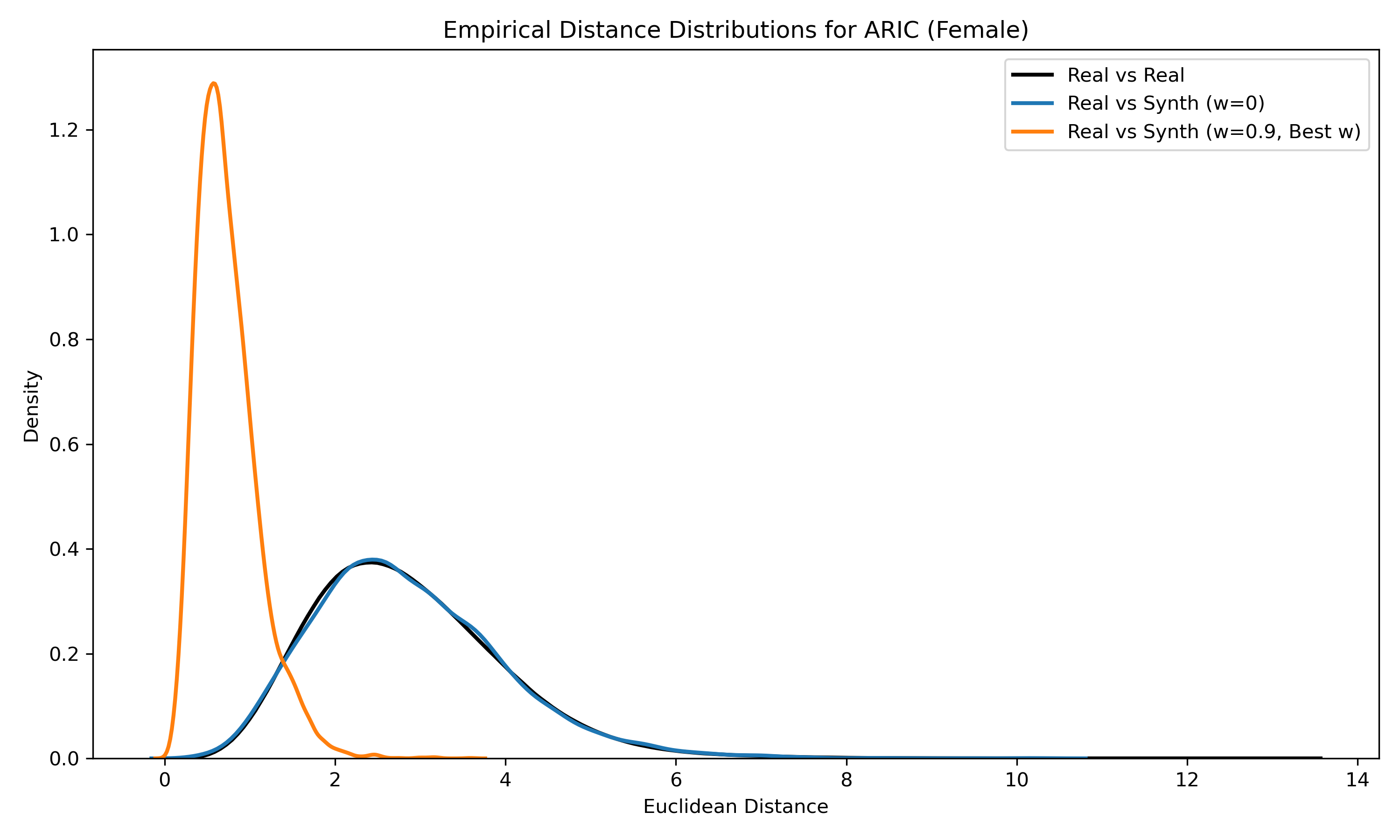}
        \caption{ARIC (Female)}
    \end{subfigure}
    \hfill
    \begin{subfigure}[b]{0.31\textwidth}
        \includegraphics[width=\linewidth]{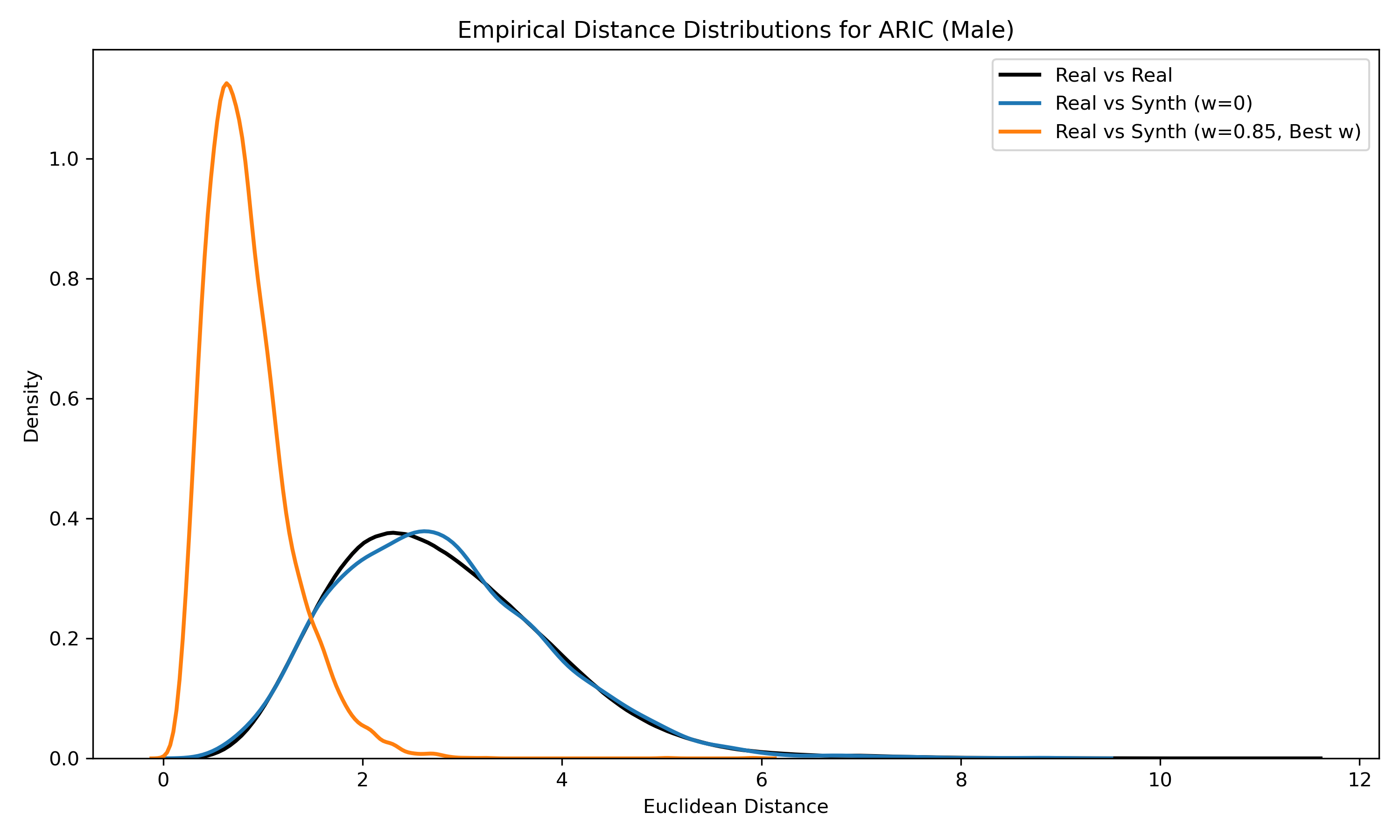}
        \caption{ARIC (Male)}
    \end{subfigure}
    \hfill
    \begin{subfigure}[b]{0.31\textwidth}
        \includegraphics[width=\linewidth]{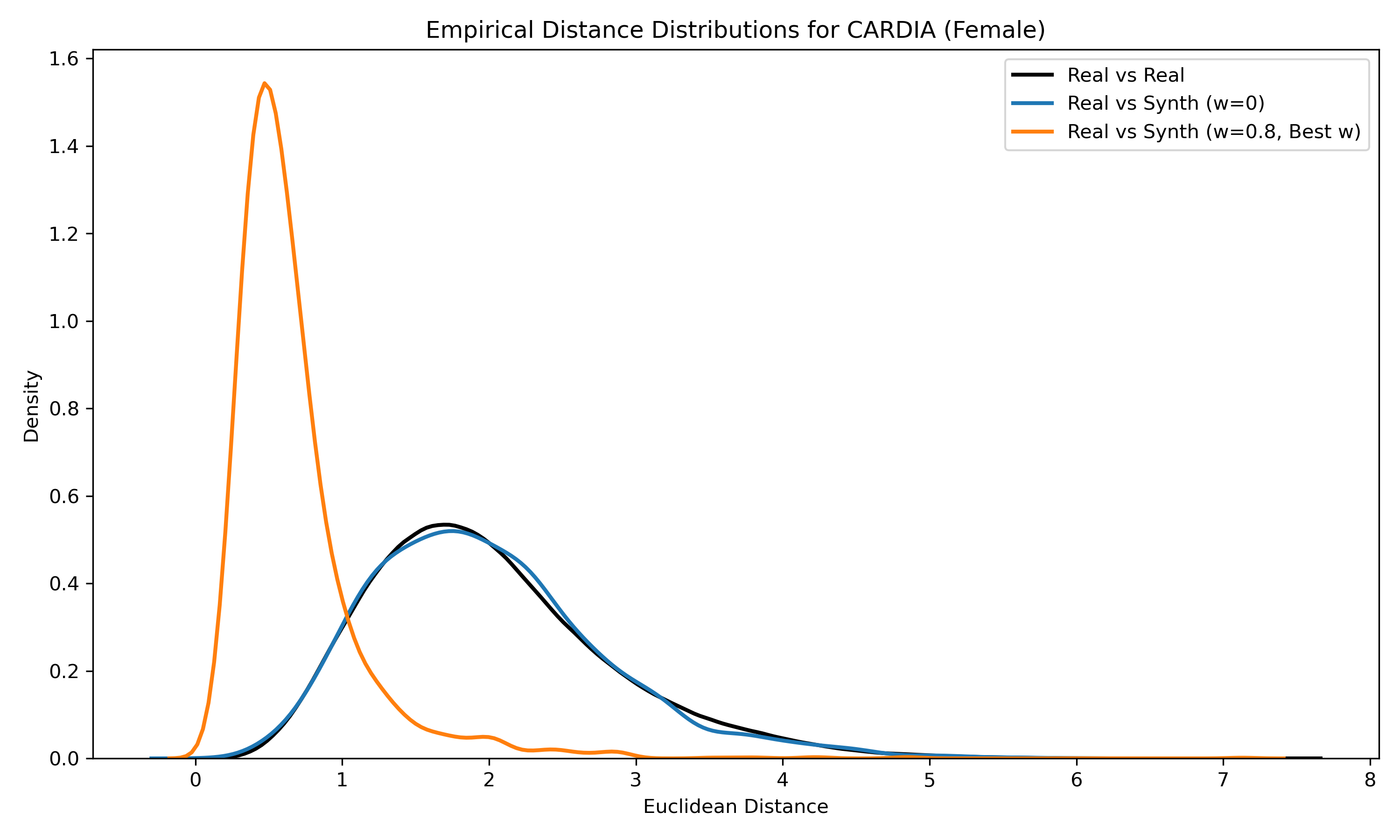}
        \caption{CARDIA (Female)}
    \end{subfigure}

    \vspace{0.5cm}

    \begin{subfigure}[b]{0.31\textwidth}
        \includegraphics[width=\linewidth]{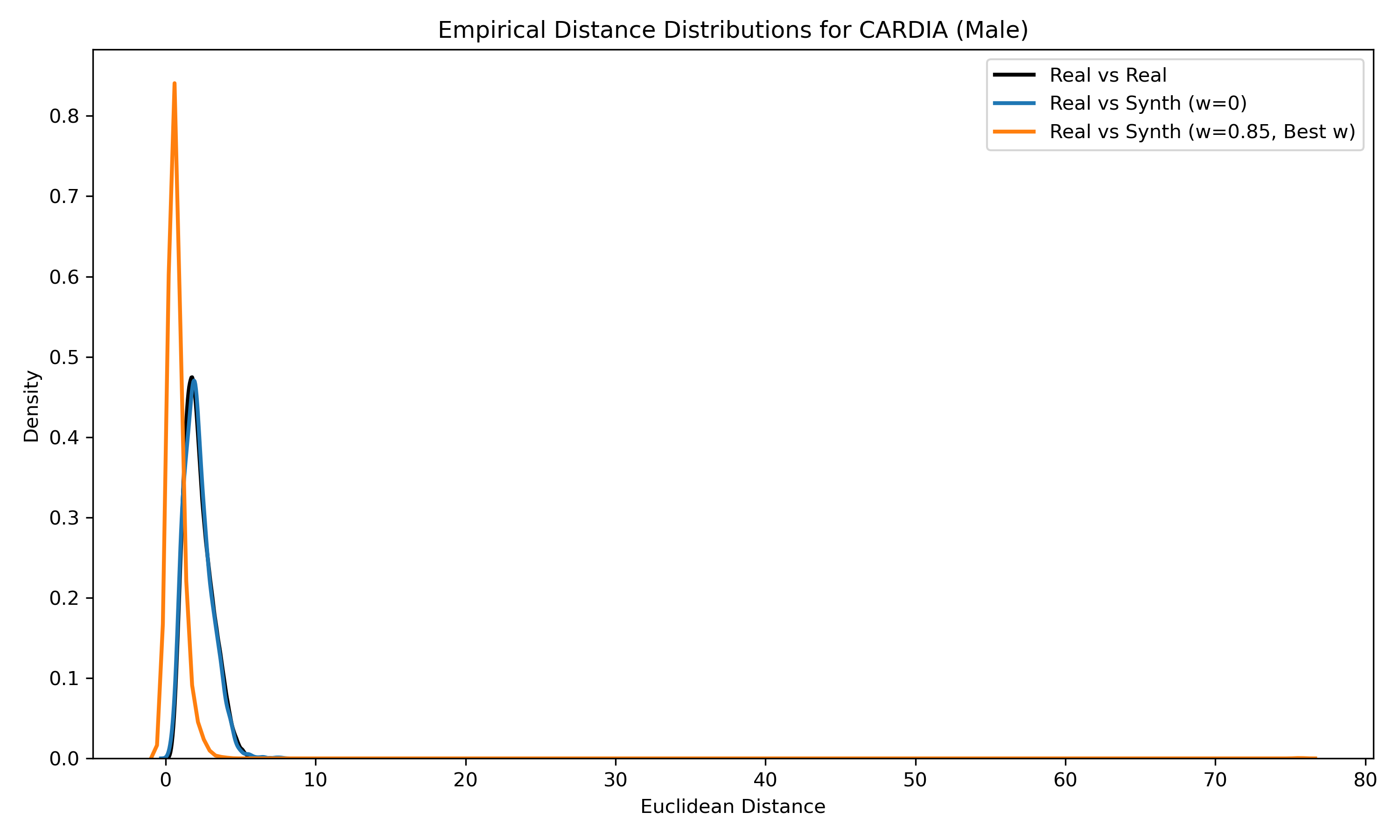}
        \caption{CARDIA (Male)}
    \end{subfigure}
    \hfill
    \begin{subfigure}[b]{0.31\textwidth}
        \includegraphics[width=\linewidth]{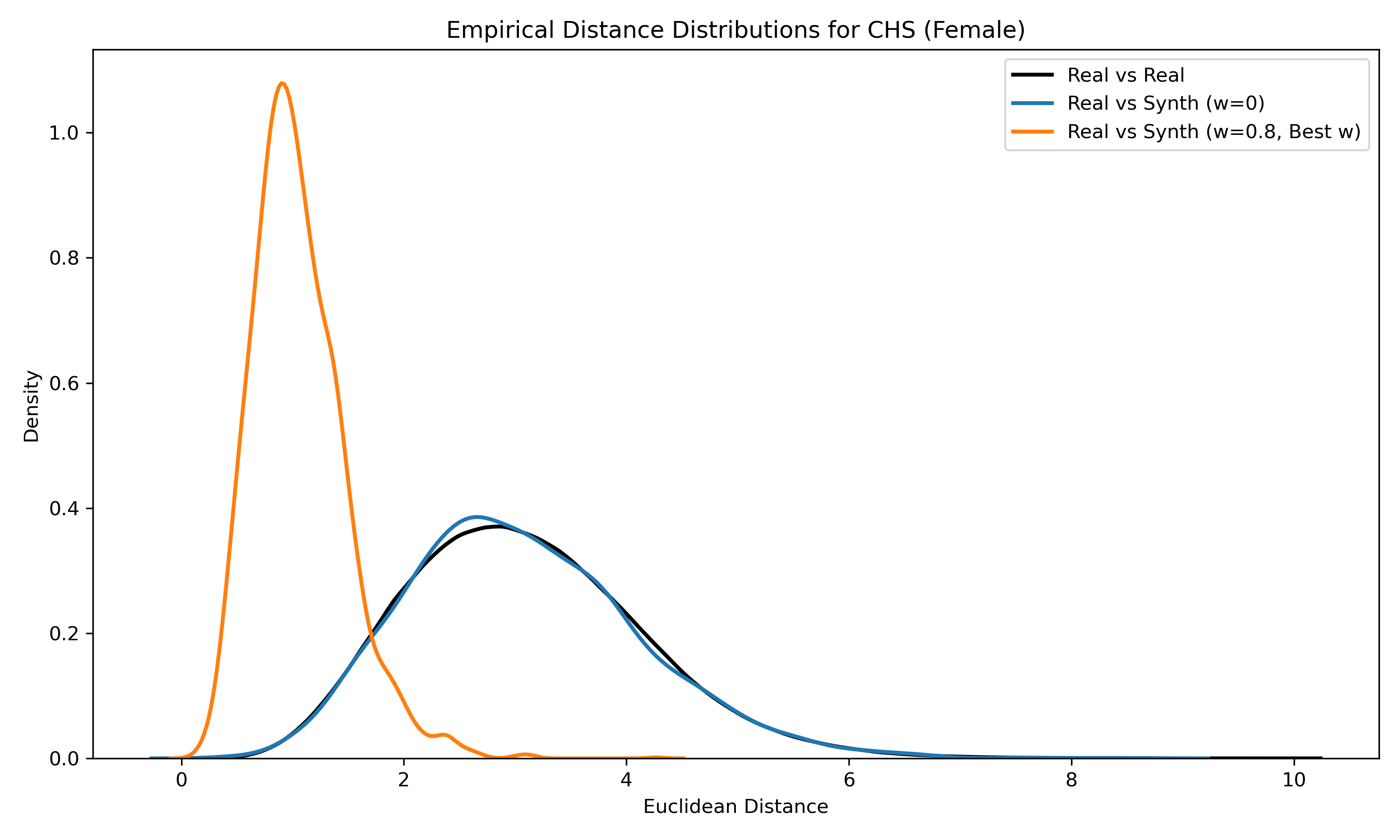}
        \caption{CHS (Female)}
    \end{subfigure}
    \hfill
    \begin{subfigure}[b]{0.31\textwidth}
        \includegraphics[width=\linewidth]{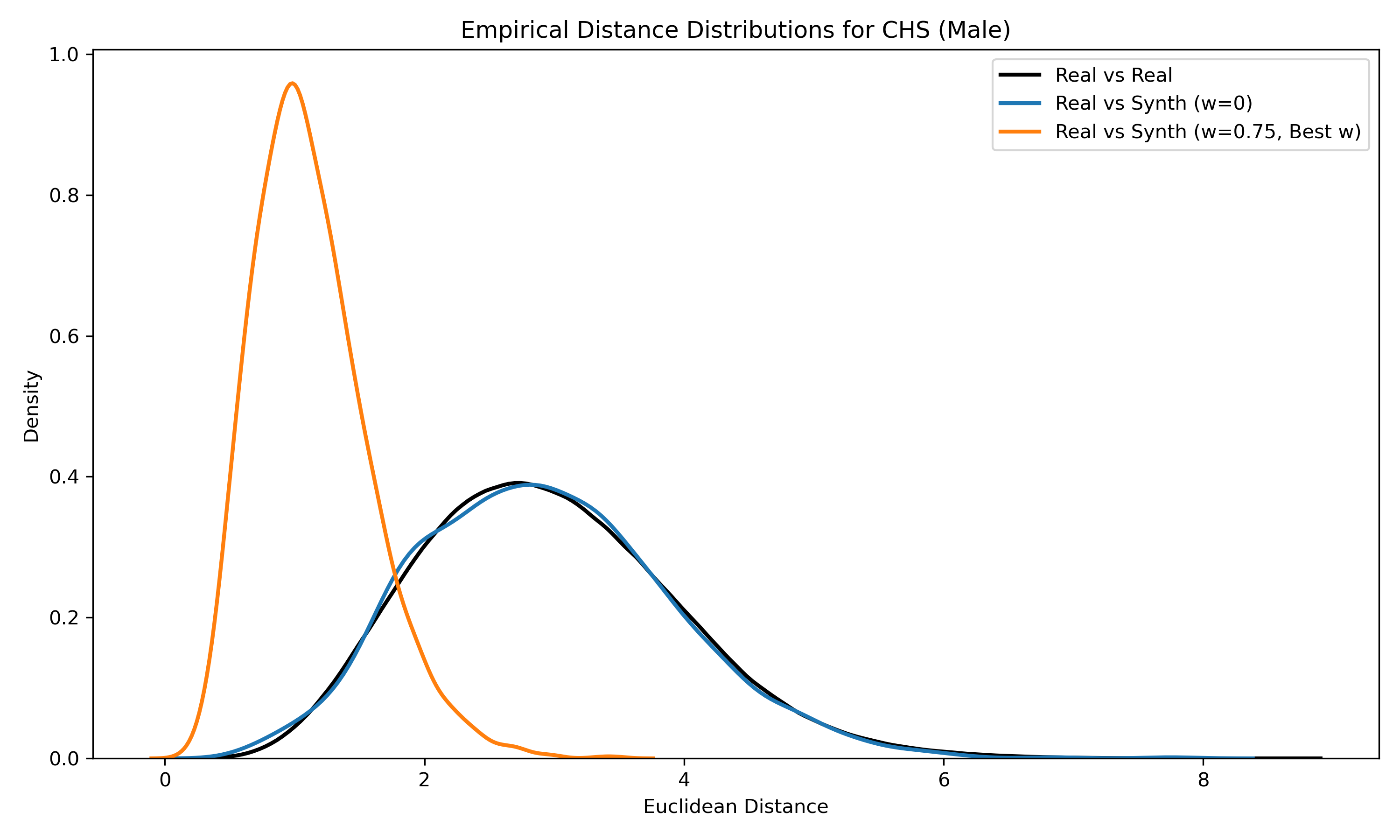}
        \caption{CHS (Male)}
    \end{subfigure}

    \vspace{0.5cm}

    \begin{subfigure}[b]{0.31\textwidth}
        \includegraphics[width=\linewidth]{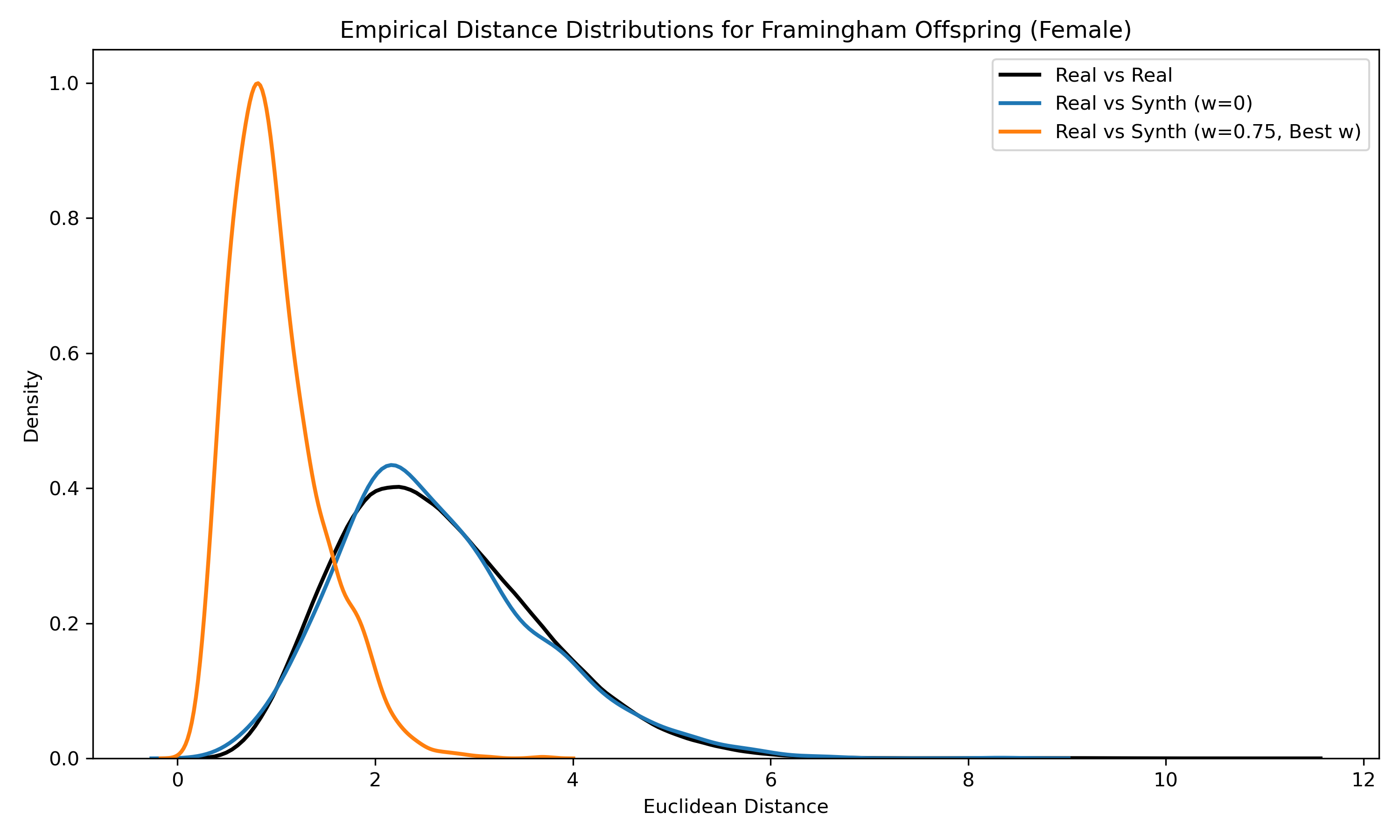}
        \caption{Framingham (Female)}
    \end{subfigure}
    \hfill
    \begin{subfigure}[b]{0.31\textwidth}
        \includegraphics[width=\linewidth]{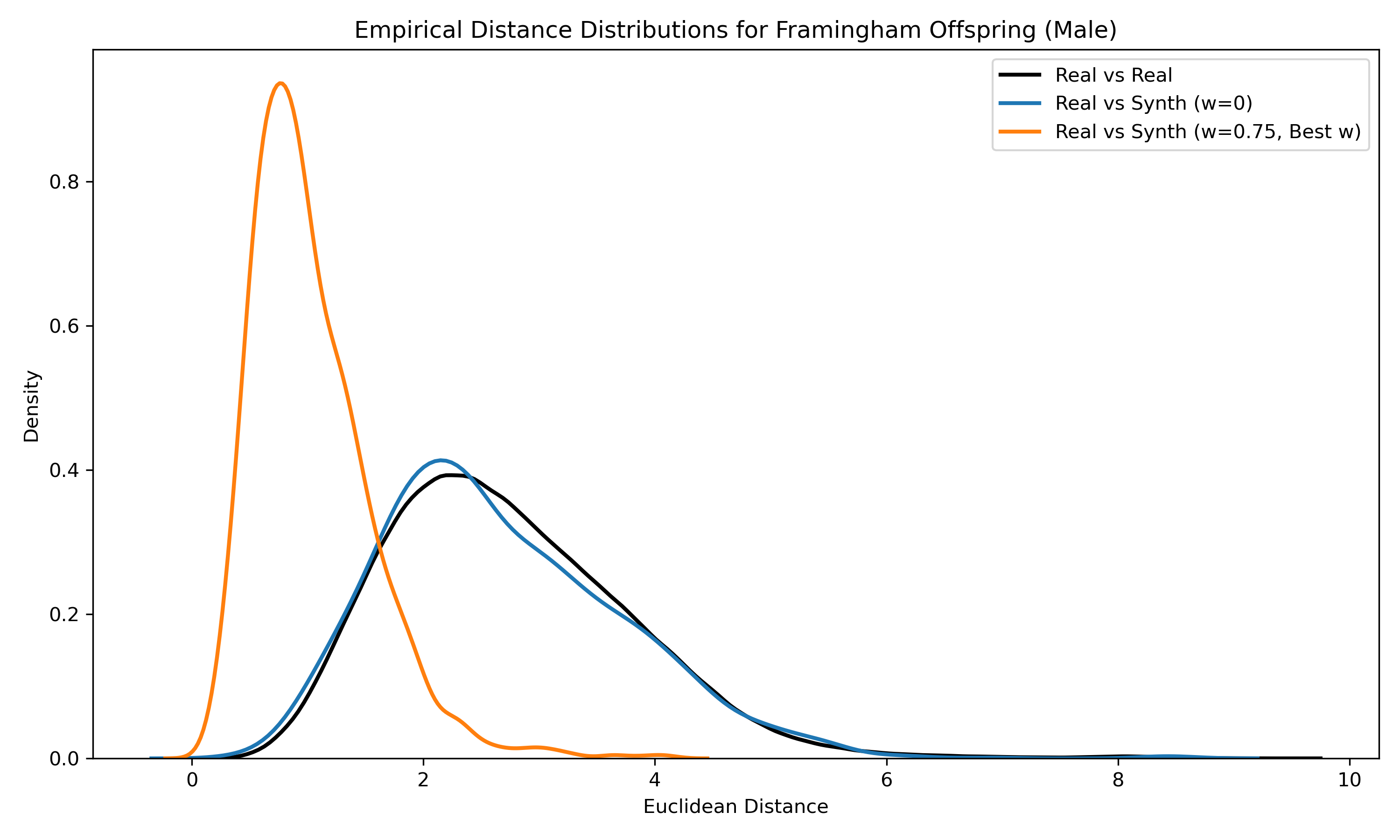}
        \caption{Framingham (Male)}
    \end{subfigure}
    \hfill
    \begin{subfigure}[b]{0.31\textwidth}
        \includegraphics[width=\linewidth]{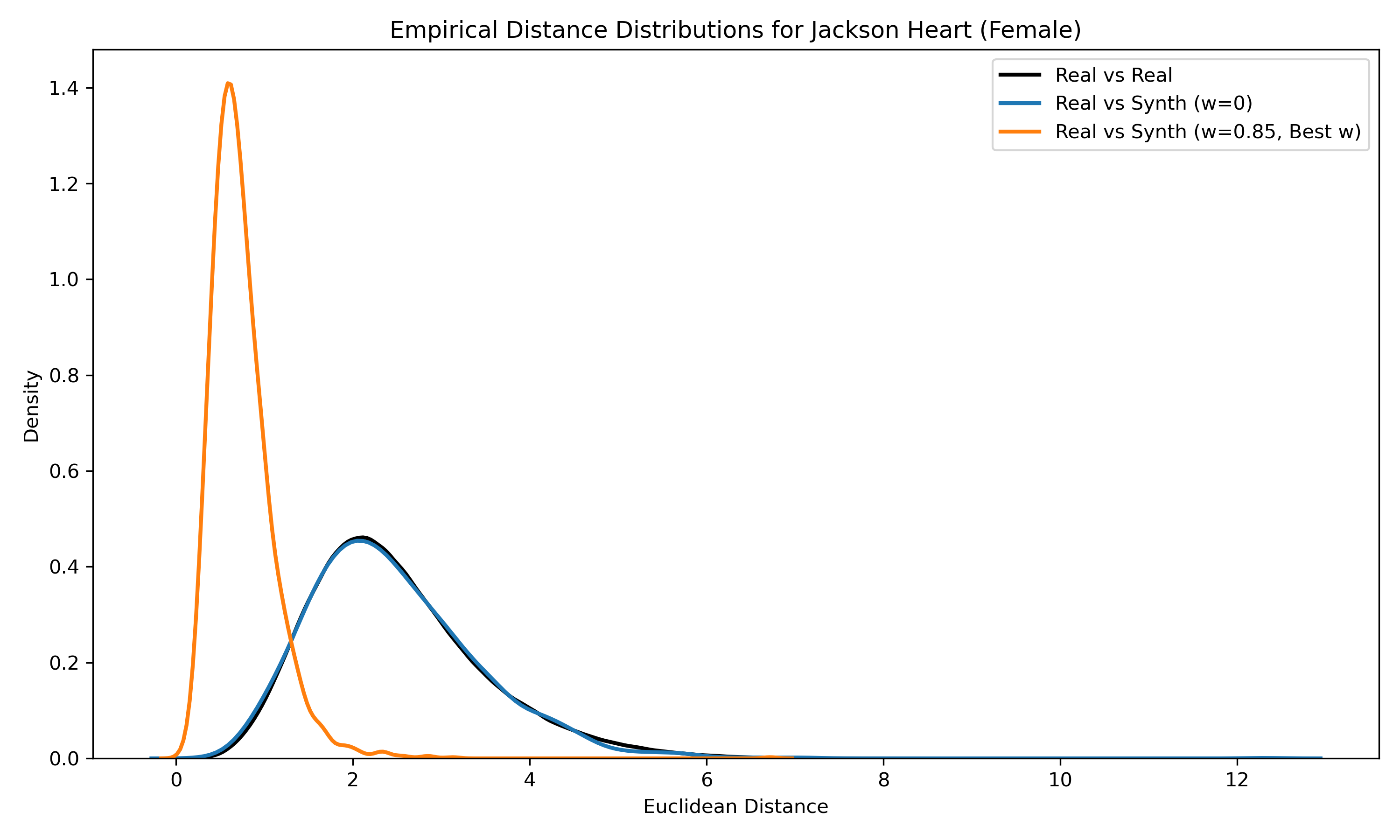}
        \caption{Jackson Heart (Female)}
    \end{subfigure}

    \vspace{0.5cm}

    \begin{subfigure}[b]{0.31\textwidth}
        \includegraphics[width=\linewidth]{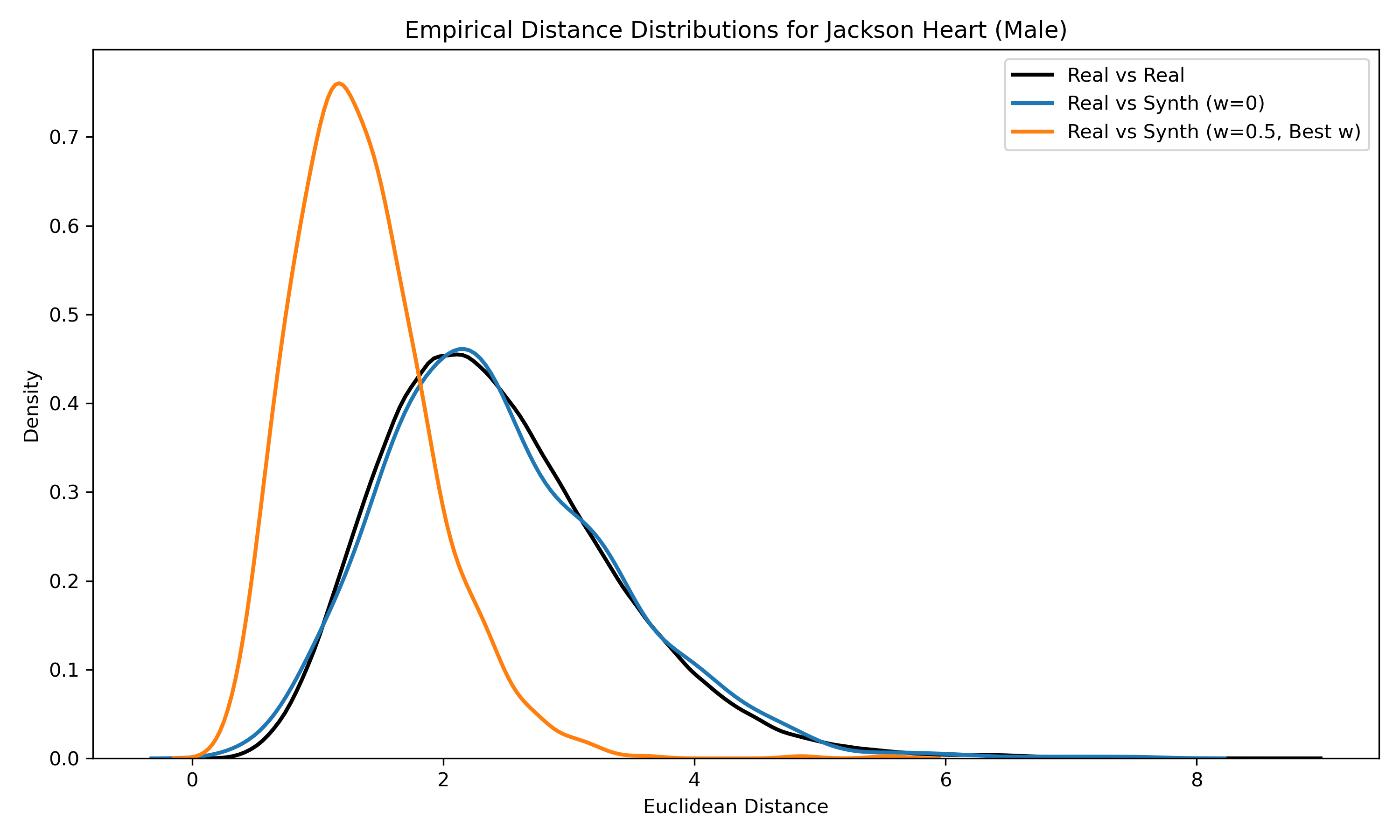}
        \caption{Jackson Heart (Male)}
    \end{subfigure}
    \hfill
    \begin{subfigure}[b]{0.31\textwidth}
        \includegraphics[width=\linewidth]{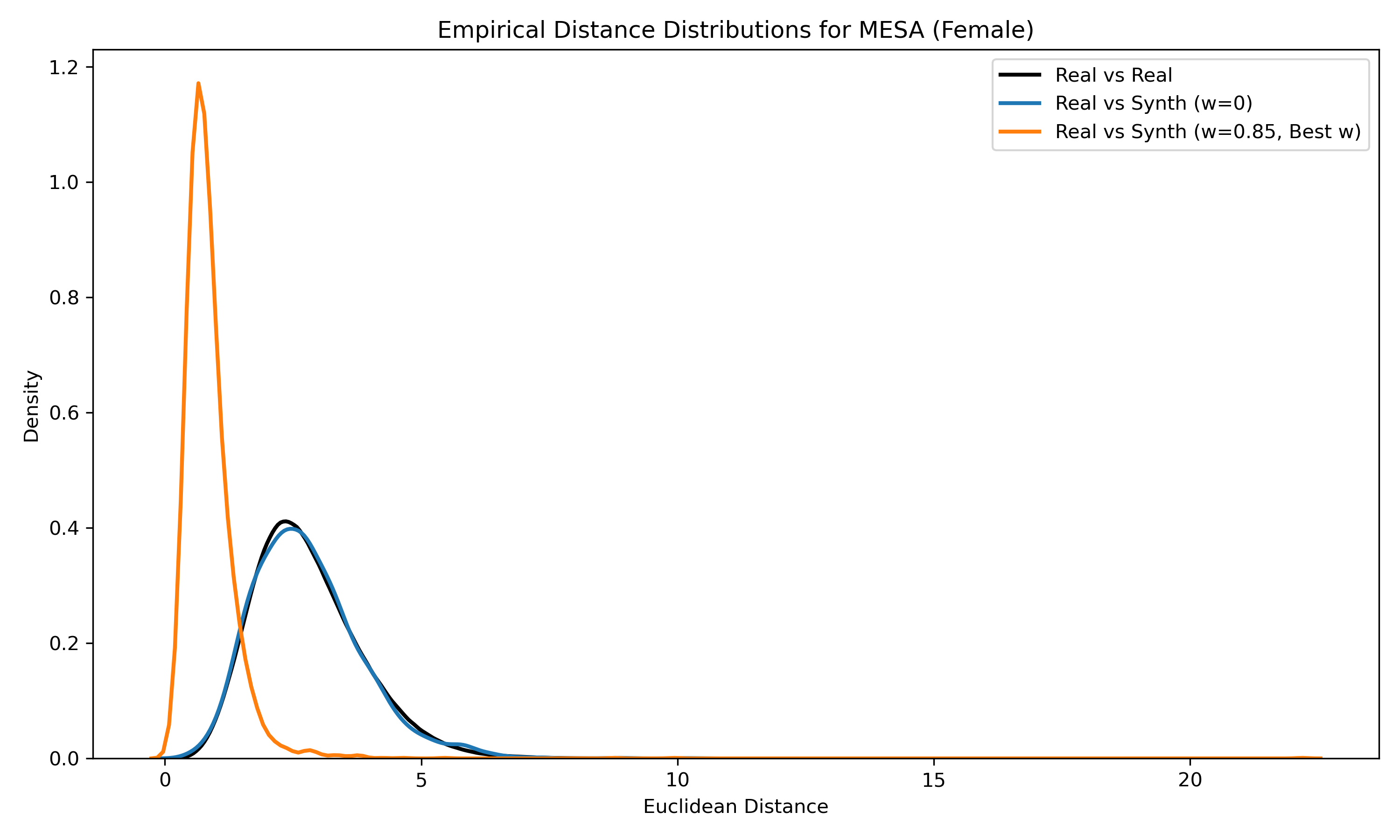}
        \caption{MESA (Female)}
    \end{subfigure}
    \hfill
    \begin{subfigure}[b]{0.31\textwidth}
        \includegraphics[width=\linewidth]{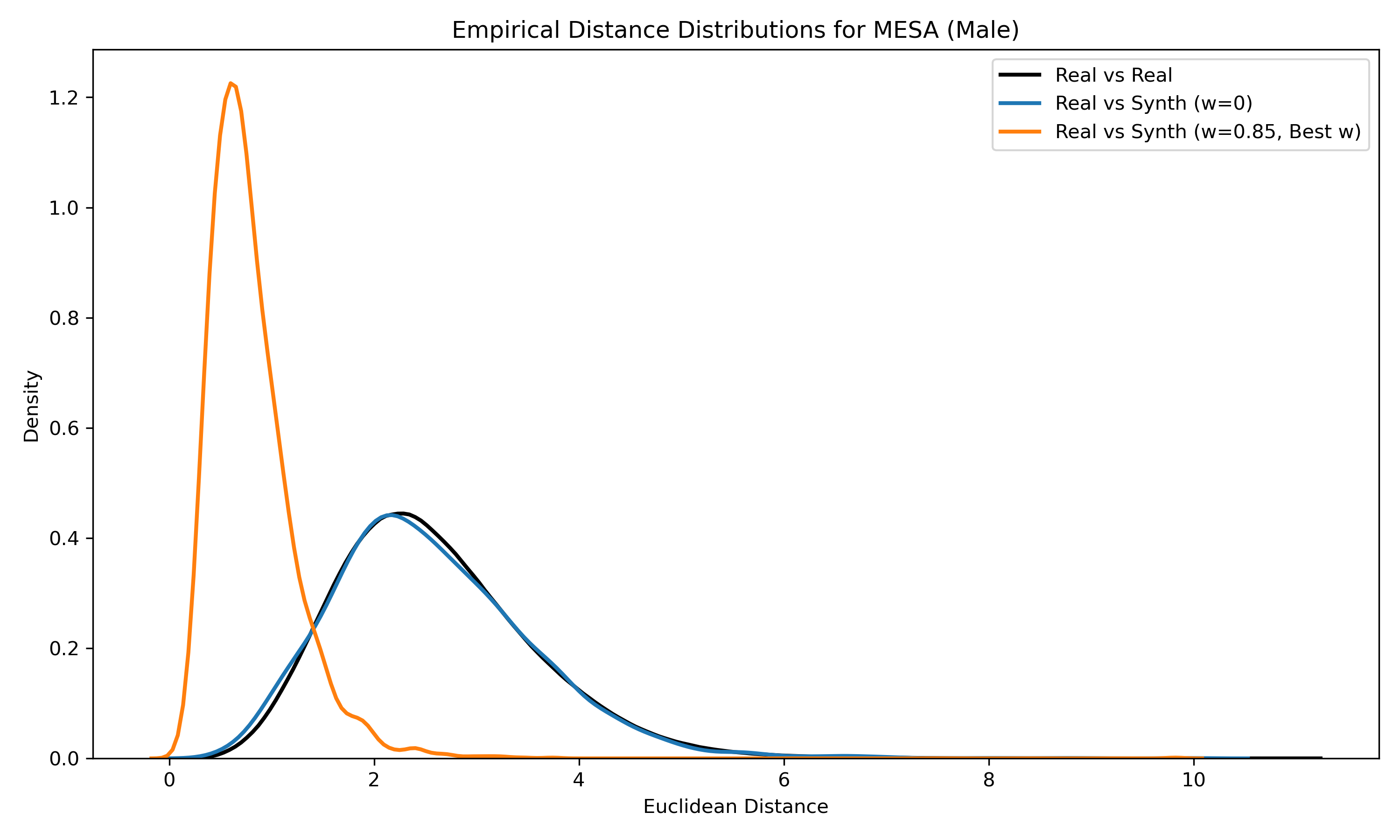}
        \caption{MESA (Male)}
    \end{subfigure}

    \caption{The empirical distribution of pairwise distances between two randomly selected real samples (black); the empirical distribution of pairwise distances between a real sample and its perturbed counterpart with $w=\hat{w}$ (orange); the empirical distribution of pairwise distances between a real sample and its perturbed counterpart with $w=0$ (blue) for all 12 sub-studies.}
    
    \label{fig:Distance_Dist_Privacy}
\end{figure}

To evaluate the privacy protection provided by selected $\hat{w},$ we additionally calculate the pairwise Euclidean Distances between real and synthetic samples and compare them with distances between two different real samples. If the distance between real and synthetic samples is very small, the risk of privacy leakage can be non-trivial as one may guess the real sample value from the released synthetic sample value.  The empirical distributions of distances between real and synthetic samples and distances of two randomly selected real samples are plotted in Figure \ref{fig:Distance_Dist_Privacy}, suggesting pair-wise distances between real and perturbed samples are indeed small, but not universally smaller than those between two distinct real observations. We calculate the empirical probability of $\|X_j-X_i\|_2<\|\tilde{X}_i-X_i\|_2$ for  $i\neq j$ to quantify the relative closeness of original and synthetic samples,
where $X_i$ is a real observation in a given study, and $\tilde{X}_i$ is its perturbed counterpart using $\hat{w}$ in the generated synthetic dataset. The results are reported as \texttt{Probability} in Table \ref{tab:Study_Stats} by sub-study.  These probabilities range from 0.9\% to 3.5\%, also suggesting sufficient perturbation to alleviate the privacy concern.  In Figure \ref{fig:Distance_Dist_Privacy}, we also plot the empirical distribution of distances between real and synthetic samples with $w=0,$ which provides 100\% privacy protection. As expected, this distribution is almost the same as the distribution of distances between two randomly selected real samples. Coupled with the fact that the reported \texttt{Probability} corresponding to $w=0$ is close to 50\% in Table \ref{tab:Study_Stats}, this supports the selection of the latter as a reference distribution to evaluate the privacy protection from selected $\hat{w}.$ 

\begin{figure}[H]
    \centering
    \includegraphics[width=0.85\textwidth]{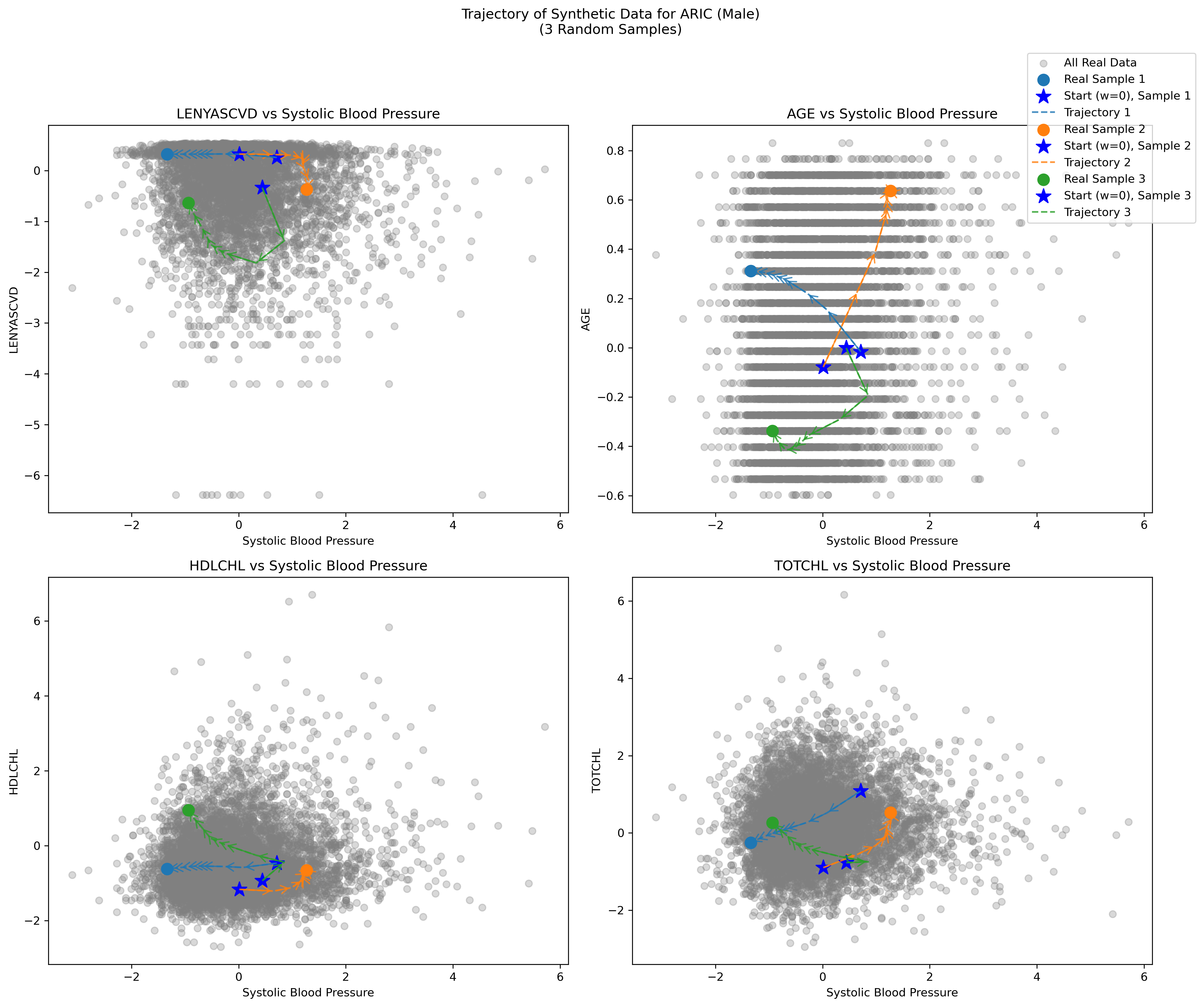}
   \caption{The evolution of $3$ randomly selected records as the perturbation weight $w$
increases from $0$ to $1$ in the ARIC–Male sub-study. All variables are standardized (a min-max logit transformation of years to CVD, and $Z$-score transformation of age, HDL, and total cholesterol). The scatter plots with SBP as the x-axis serve as a reference density to examine the effect of perturbation.  For each of the three individuals, a colored curve starts from its fully perturbed draw at $w=0$ (star) to its unperturbed value at $w=1$ (solid circle).}
    \label{fig:Synthetic_Trajectory_ARIC_Male}
\end{figure}

To further quantify the privacy protection with selected $\hat{w}$, for each real observation $X_i$ in a study, we calculate the distance $\|X_i-\tilde{X}_i\|_2$ and the distances between $X_i$ and other real observations:
$$D_i=\left\{ \|X_i-X_j\|_2 \mid j=1, \cdots, i-1, i+1, \cdots, n\right\}.$$
Then, we calculate the rank of $\|X_i-\tilde{X}_i\|_2$ among $D_i$, calculated as
\begin{equation} \label{eq:Med_Ranks}
r_i=\sum_{j \neq i} I\left( \|X_i-\tilde{X}_i\|_2 >  \|X_i-X_j\|_2  \right).
\end{equation}
The medians of $r_i$ for different $w$ are summarized in Table \ref{tab:Study_Stats}. For instance, among male ARIC participants, it suggests that, on average, when $w=\hat{w}=0.85,$ the perturbation moves the original observation beyond its $16$ nearest neighbors.  In other words, the perturbation is likely sufficient, since there are many distinct real samples closer to $X_i$ than its perturbed counterpart in the synthetic dataset. When $w=0,$ that is, when a perfect privacy protection is provided,  the median rank is close to 50\% of the sample size as expected. To visualize the effect of the proposed perturbation in synthetic data generation, Figure \ref{fig:Synthetic_Trajectory_ARIC_Male} shows trajectories for $3$ randomly selected observations from the ARIC (Male) sub-study as they transition from \( w = 0 \) to \( w = 1 \).

\begin{table}[H]

\centering
\begin{tabular}{lrrrr}
\hline
Study & \texttt{Size of Hidden Layer} & \texttt{Number of Hidden Layers} & \texttt{Number of Flows} \\
\hline
ARIC (M) & 128 & 1 & 5 \\
ARIC (F) & 64 & 1 & 4 \\
CARDIA (M) & 64 & 2 & 4 \\
CARDIA (F) & 128 & 1 & 4  \\
CHS (M) & 64 & 2 & 4  \\
CHS (F) & 64 & 1 & 5  \\
MESA (M) & 32 & 2 & 4  \\
MESA (F) & 32 & 2 & 5  \\
Framingham (M) & 64 & 1 & 5  \\
Framingham (F) & 64 & 1 & 4  \\
Jackson Heart (M) & 32 & 1 & 3  \\
Jackson Heart (F) & 32 & 2 & 3 \\
\hline
\end{tabular}
\caption{The configurations of tuning parameters selected via cross validation for MAF training across 12 sub-studies.}
\label{tab:MAFtuning}
\end{table}

\begin{figure}[H]
    \centering

    \begin{subfigure}[b]{0.48\textwidth}
        \includegraphics[width=\textwidth]{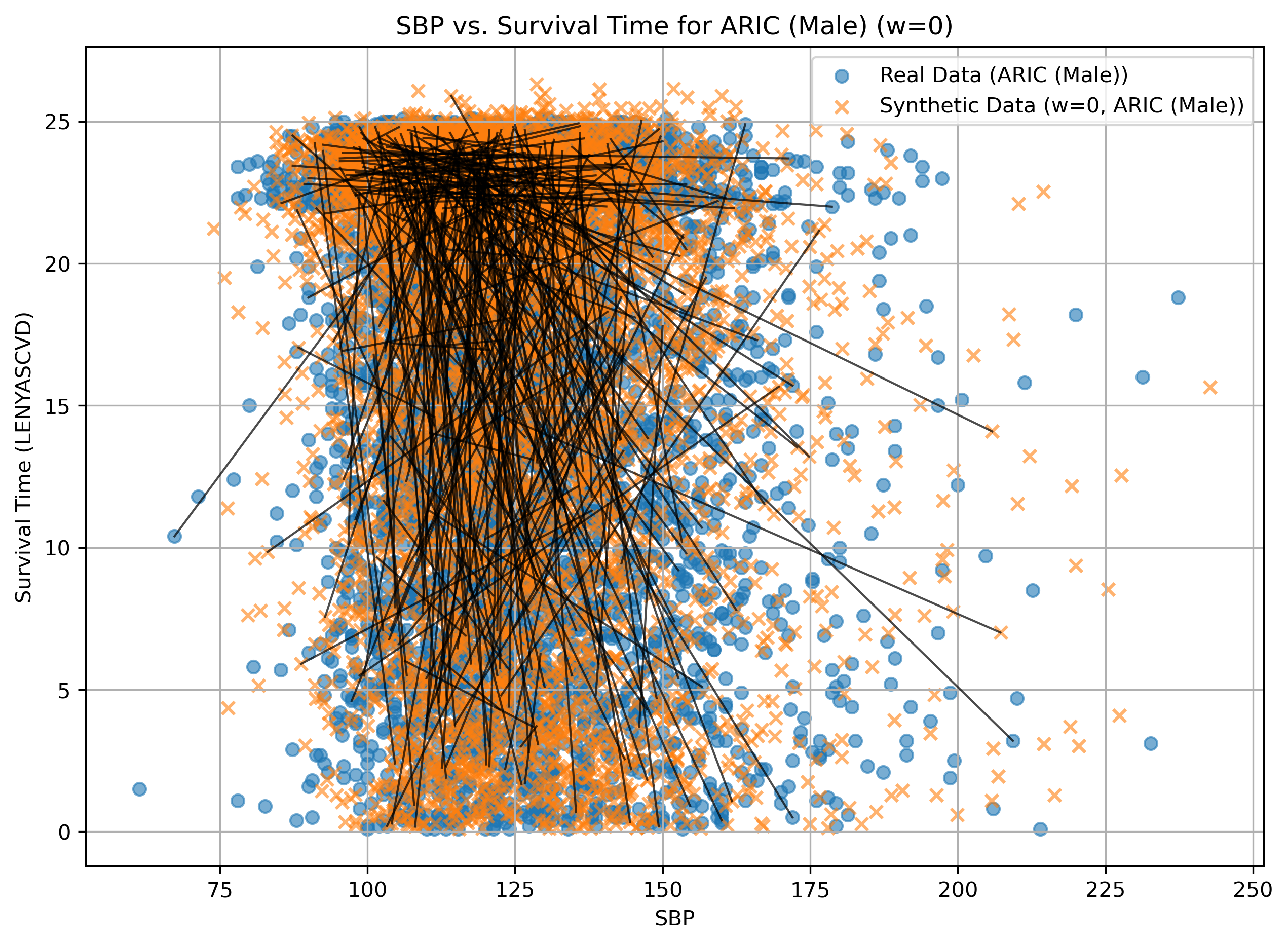}
        \caption{SBP vs. Survival Time (ARIC (Male), $w=0$), Randomly Sampled 300  Lines Connected}
        \label{fig:Sbp_Survival_w0}
    \end{subfigure}
    \hfill
    \begin{subfigure}[b]{0.48\textwidth}
        \includegraphics[width=\textwidth]{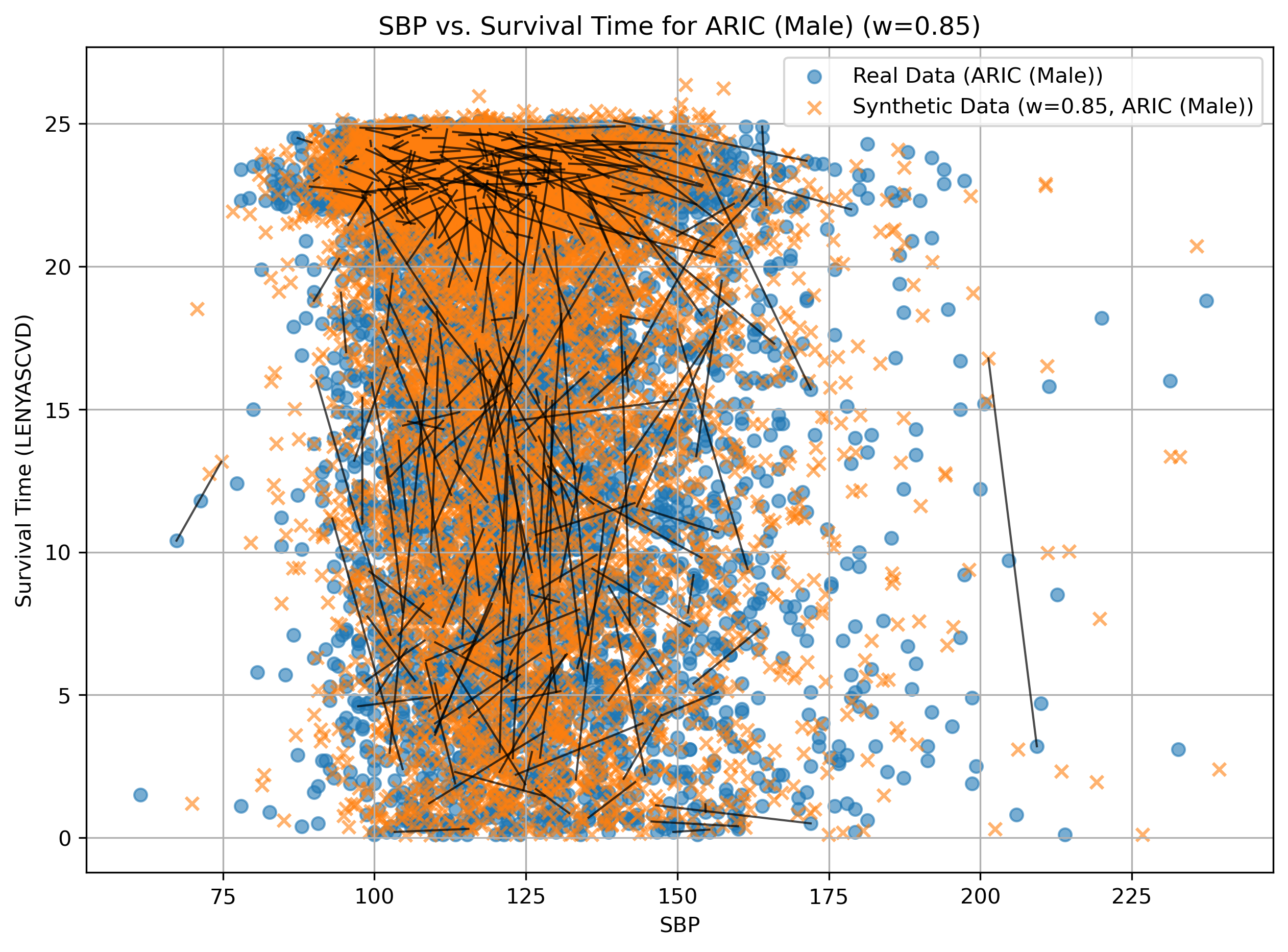}
        \caption{SBP vs. Survival Time (Aric (Male), $w=0.85$), Randomly Sampled 300 Lines Connected}
        \label{fig:Sbp_Survival_w09}
    \end{subfigure}

    \caption{Relationship between SBP (mmHg) and time to CVD (year) in the ARIC–Male sub-study. 
The left panel shows the high-privacy protection setting with $w=0$. The right panel
shows the setting with selected $\hat w=0.85$; blue circles are observed records and orange crosses are their perturbed
counterparts. To visualize record-level perturbation, $300$ randomly selected real–synthetic
pairs are linked by black line segments.}
    \label{fig:Sbp_Survival_Comparison}
\end{figure}

\begin{figure}[H]
    \centering

    \begin{subfigure}[b]{0.48\textwidth}
        \includegraphics[width=\textwidth, height=6cm]{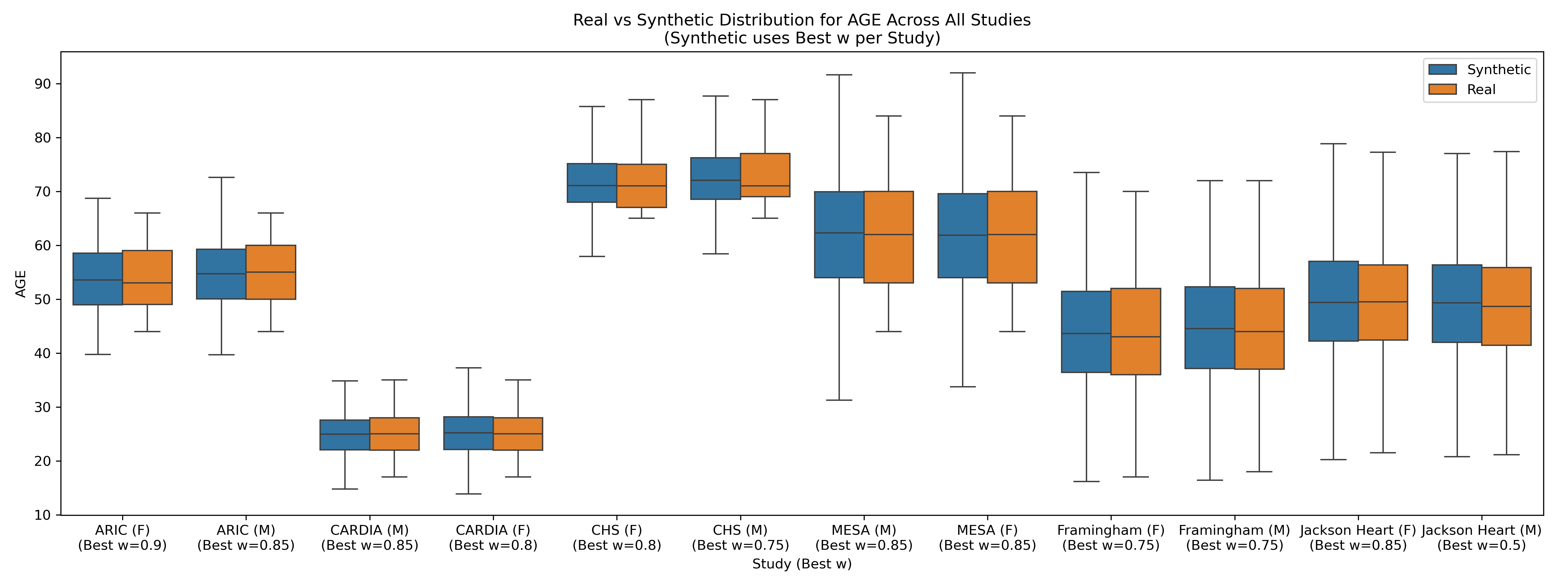}
        \caption{Age (year)}
    \end{subfigure}
    \hfill
    \begin{subfigure}[b]{0.48\textwidth}
        \includegraphics[width=\textwidth, height=6cm]{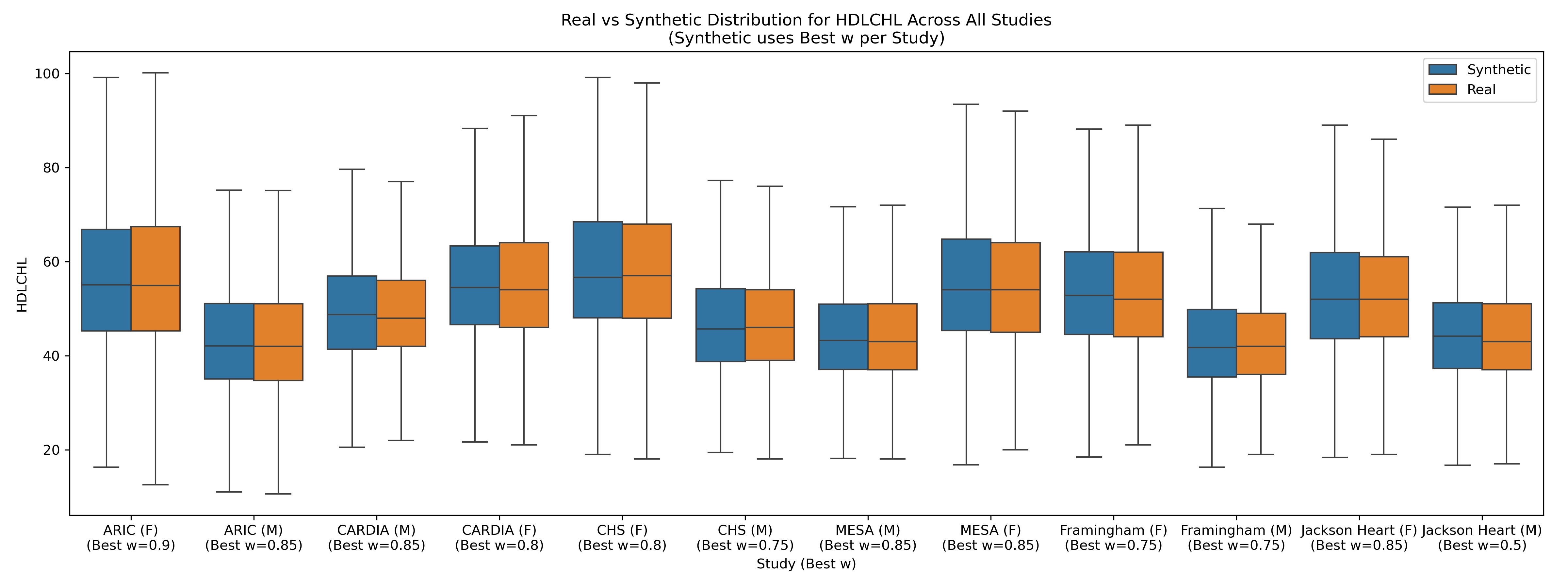}
        \caption{HDL (mg/dL)}
    \end{subfigure}

    \vspace{0.5cm}

    \begin{subfigure}[b]{0.48\textwidth}
        \includegraphics[width=\textwidth, height=6cm]{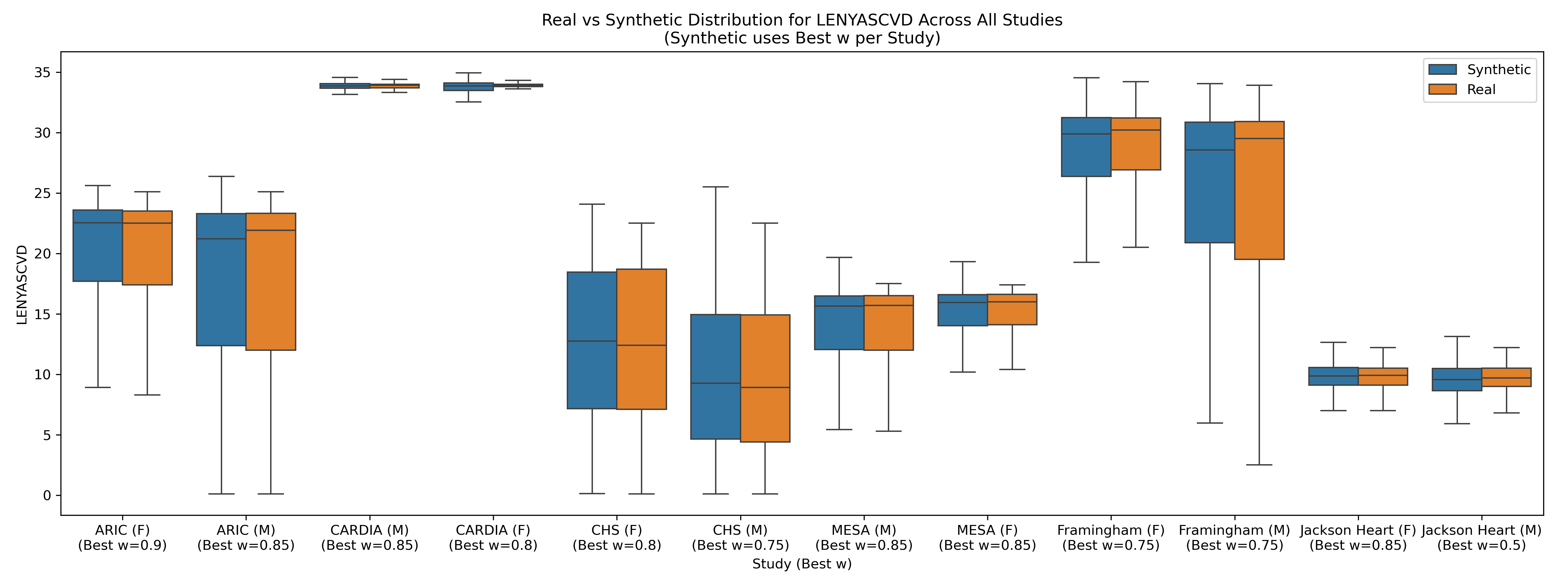}
        \caption{Time to CVD Event subject to Right Censoring (year)}
    \end{subfigure}
    \hfill
    \begin{subfigure}[b]{0.48\textwidth}
        \includegraphics[width=\textwidth, height = 6 cm]{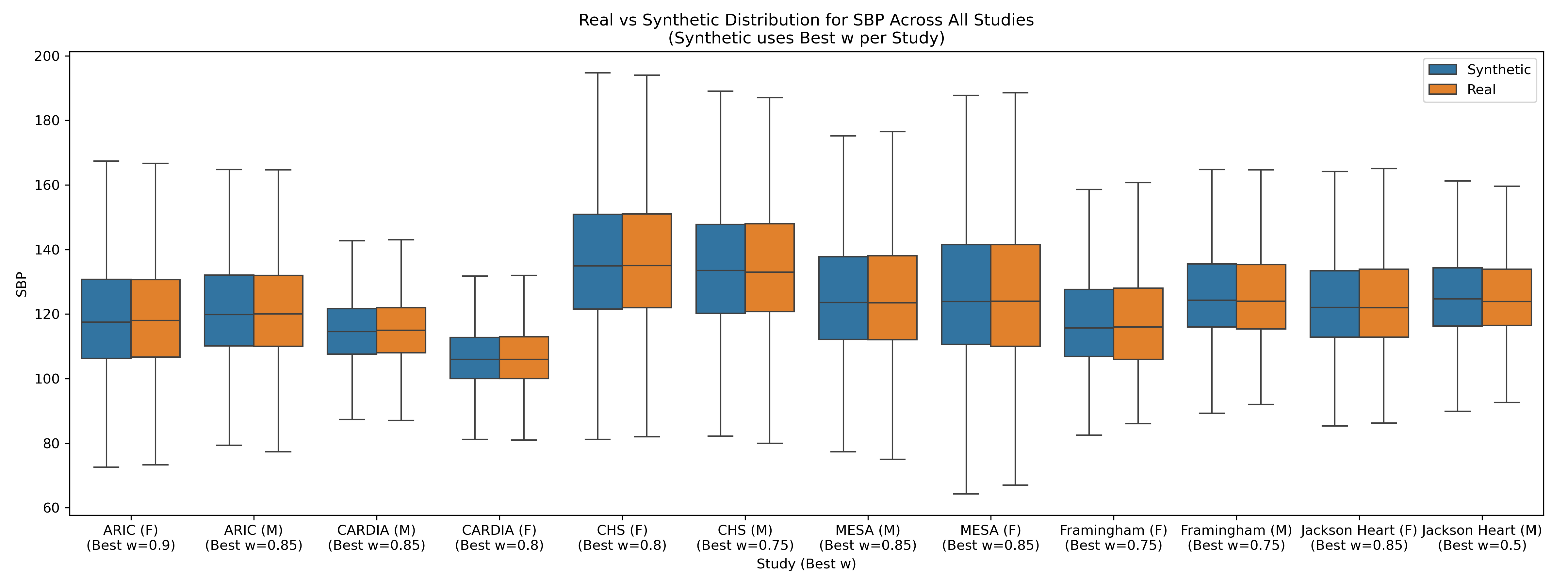}
        \caption{SBP (mmHg)}
    \end{subfigure}

    \vspace{0.5cm}

    \begin{subfigure}[b]{0.48\textwidth}
        \includegraphics[width=\textwidth, height=6cm]{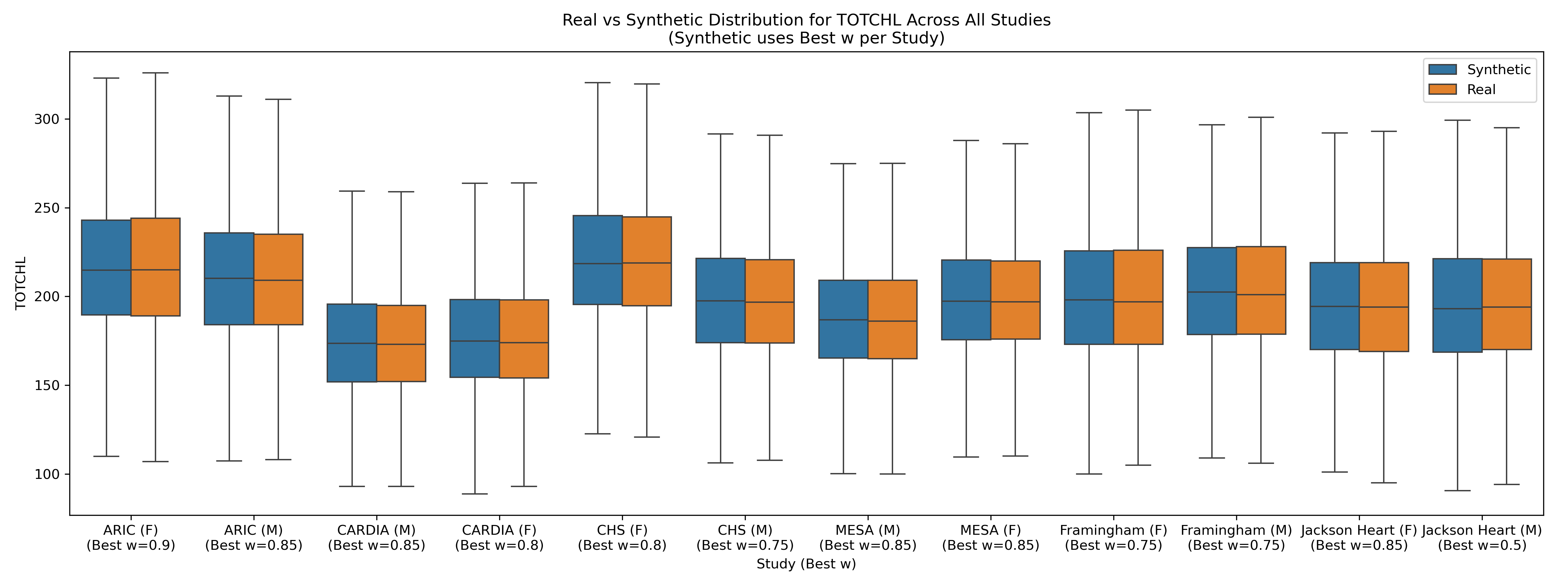}
        \caption{Total Cholesterol Level (mg/dL)}
    \end{subfigure}
    \hfill
    \begin{subfigure}[b]{0.48\textwidth}
    \end{subfigure}

    \caption{The boxplot of the synthetic data generated with the
selected $\hat w$ (blue) against that of the real data (orange). A close alignment of medians and quartiles between the two distributions suggests that the synthetic generator preserves the marginal distribution of each variable across 12 sub-studies.}
    \label{fig:Boxplot_Comparison}
\end{figure}

Figures \ref{fig:Sbp_Survival_Comparison} and \ref{fig:Boxplot_Comparison} further examine the fidelity of the synthetic data. Figure \ref{fig:Sbp_Survival_Comparison} compares real and synthetic joint distributions of SBP and survival time (time to CVD) in ARIC (Male) sub-study, where the synthetic data is generated using $w=0$ and $w = 0.85$.  Note that lines are drawn between the original and perturbed data points in each plot to highlight the effect of perturbation. These connecting lines reveal substantial movement for both $w = 0$ and $w = 0.85$, demonstrating that the synthetic data differ meaningfully from the original data on an individual level, which is crucial for ensuring privacy. At the same time, the overall joint distribution is well retained, indicating that the statistical utility is preserved. 

Figure \ref{fig:Boxplot_Comparison} presents boxplots of the real versus synthetic distributions for key variables across all studies, using \( \hat{w} \) for each. These visual comparisons indicate that the synthetic data maintain close alignment with real data distributions, suggesting high quality of the trained MAF models.

\subsection{Synthetic Data Analysis}
\begin{figure}[H]
    \centering
   \begin{subfigure}[b]{\textwidth}
       \includegraphics[width=\textwidth]{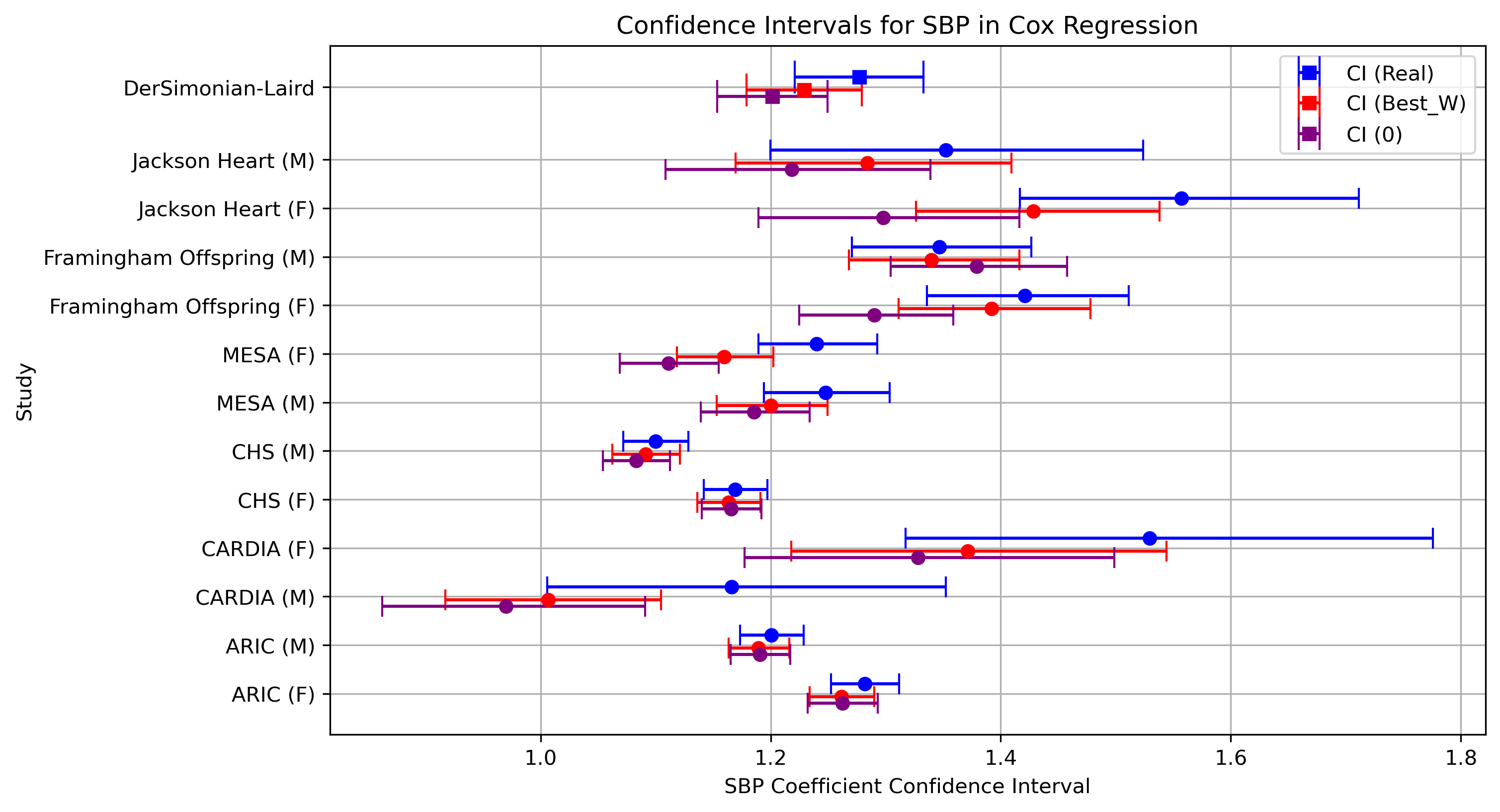}
       \caption{Unadjusted Hazard Ratio Corresponding to 10 mmHg Difference in SBP}
       \label{fig:SBP_Only}
   \end{subfigure}
    \hfill
    \begin{subfigure}[b]{\textwidth}
       \includegraphics[width=\textwidth]{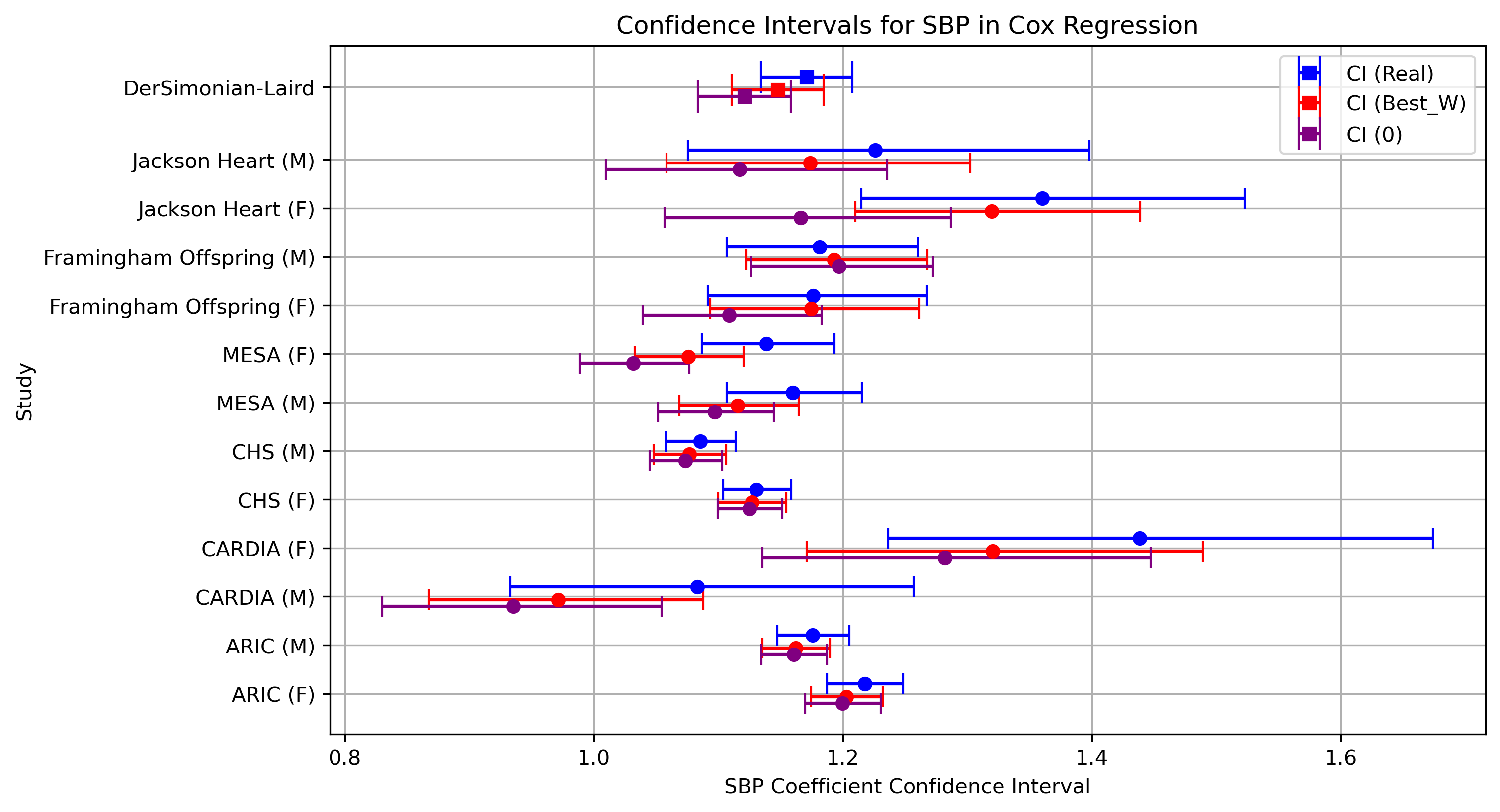}
       \caption{Adjusted Hazard Ratio Corresponding to 10 mmHg Difference in SBP (adjusted for Age, Total Cholesterol Level and HDL)}
       \label{fig:SBP_ALL}
    \end{subfigure}

    \caption{The forest plots for estimated hazard ratio associated with an increase of 10 mmHg in SBP; point estimates and associated 95\% CI from individual sub-studies and meta analysis are reported; estimates based on real data are colored blue; estimates based on synthetic data generated with $w=\hat{w}$ are colored red; and estimates based on synthetic data generated with $w=0$ are colored purple.}
    \label{fig:SBP_Meta}
\end{figure}

For generated synthetic data of each sub-study, we also fit two Cox Regression models and obtain relevant parameter estimates and their standard errors, which are then aggregated to generate a single estimator for the hazard ratio via the same meta analysis above. The results are reported in Figure \ref{fig:SBP_Meta}. The results based on synthetic data are very consistent with those based on original data. For example, after adjusting for age, total cholesterol level, and HDL, the final estimated HR based on synthetic data generated using $\hat{w}$ was $1.148$ (95\% CI: $1.111$ to $1.185$) for 10 mmHg difference in SBP, suggesting every 10 mmHg higher in baseline SBP is associated with a $14.8\%$  increase in CVD hazard. This result is fairly close to that based on original data: HR of $1.171$ (95\% CI: $1.134$ to $1.207$). On the other hand, when $w=0,$ the HR estimate is further way from that based on original data. Specifically, the adjusted HR estimate is $1.121$ (95\% CI: $1.084$ to $1.158$).  The results of unadjusted analysis follow a similar pattern. Overall, analyses results based on original data, synthetic data with $w=0$, and synthetic data with $w=\hat{w}$ all suggest a statistically significant association between SBP level and CVD risk, with the results based on synthetic data generated with $w=\hat{w}$ better reproducing those based on original data. This trend is also observed for individual studies, especially studies with a large sample size.

\begin{table}[H]
\centering
\caption{Selected \( w \) for 12 sub-studies.}
\label{tab:Best_w_Per_Study}
\begin{tabular}{lc}
\toprule
\textbf{Study} & \textbf{Best \( w 
 = \hat{w}\)} \\
\midrule
ARIC (M)                   & 0.85 \\
ARIC (F)                 & 0.90 \\
CARDIA (M)                 & 0.85 \\
CARDIA (F)               & 0.80 \\
CHS (M)                    & 0.75 \\
CHS (F)                  & 0.80 \\
MESA (M)                   & 0.85 \\
MESA (F)                 & 0.85 \\
Framingham Off. (M)        & 0.75 \\
Framingham Off. (F)      & 0.75 \\
Jackson Heart (M)          & 0.50 \\
Jackson Heart (F)        & 0.85 \\
\bottomrule
\end{tabular}
\end{table}

\begin{table}[H]
\centering
\caption{The selected $w = \hat{w}$ across 12 sub-studies and privacy protection at $w=0$. The median refers to the the median ranks (\ref{eq:Med_Ranks}), and the probability refers to the empirical probability  (\ref{eq:Probability4w}). A higher median rank and empirical probability for a study indicates stronger privacy protection. $n$ refers to the sample size of all $12$ sub-studies.}
\label{tab:Study_Stats}
\begin{tabular}{llccc}
\toprule
\textbf{Study} & \textbf{$w$} & \textbf{Median} & \textbf{Probability} & \textbf{$n$} \\
\midrule
ARIC (M)                 & 0.00 & 3389 & 0.503 & 6762 \\
                            & 0.85 & 16   & 0.015 &  \\
ARIC (F)               & 0.00 & 4043 & 0.504 & 8101 \\
                            & 0.90 & 10   & 0.009 &  \\
CARDIA (M)              & 0.00 & 1156 & 0.513 & 2297 \\
                            & 0.85 & 9    & 0.027 &  \\
CARDIA (F)            & 0.00 & 1370 & 0.508 & 2743 \\
                            & 0.80 & 15   & 0.023 &  \\
CHS (M)                 & 0.00 & 1184 & 0.479 & 2413 \\
                            & 0.75 & 18   & 0.029 &  \\
CHS (F)               & 0.00 & 1581 & 0.492 & 3254 \\
                            & 0.80 & 15   & 0.022 &  \\
MESA (M)                & 0.00 & 1539 & 0.506 & 3176 \\
                            & 0.85 & 7    & 0.014 &  \\
MESA (F)              & 0.00 & 1757 & 0.512 & 3552 \\
                            & 0.85 & 9    & 0.018 &  \\
Framingham (M)     & 0.00 & 745  & 0.489 & 1608 \\
                            & 0.75 & 11   & 0.034 &  \\
Framingham (F)   & 0.00 & 858  & 0.498 & 1766 \\
                            & 0.75 & 12   & 0.035 &  \\
Jackson Heart (M)       & 0.00 & 619  & 0.515 & 1196 \\
                            & 0.50 & 64   & 0.127 &  \\
Jackson Heart (F)     & 0.00 & 925  & 0.525 & 1913 \\
                            & 0.85 & 4    & 0.021 &  \\
\bottomrule
\end{tabular}
\end{table}


\begin{comment}
\begin{figure}[h!]
    \centering
    \begin{subfigure}[b]{0.48\textwidth}
        \includegraphics[width=\textwidth, height=6cm]{Images/Difference_Rank_Plot.png}
        \caption{L2 Distance Rank Distribution}
        \label{fig:L2_Distance}
    \end{subfigure}
    \hfill
    \begin{subfigure}[b]{0.48\textwidth}
        \includegraphics[width=\textwidth, height=6cm]{Images/Roc_Curve.png}
        \caption{ROC Curve}
        \label{fig:ROC_Curve}
    \end{subfigure}
    \caption{Synthetic Data Quality Analysis}
    \label{fig:Synthetic_Data_Quality}
\end{figure}
\end{comment}

By our Latent Noise Injection method and performing inference over $K$ synthetically generated  studies, we create a dual benefit: individual level privacy protection and valid statistical inference by retaining similar statistical properties to support meta analytic conclusions.

\subsection{Summary}

Our empirical evaluation demonstrates that the MAF model effectively generates synthetic data that preserves the distribution of the original data, and the Latent Noise Injection method can provide reasonable privacy protection when the perturbation size is appropriately selected. With sufficient sample size, the synthetic data closely aligns with the real data and can reproduce results of statistical analysis, such as Cox regression analysis, based on real data reasonably well. When sample size is smaller, the difference in analysis results between synthetic and real data can be greater as expected. In general, the meta analysis results incorporating multiple studies with different sample sizes are even more reproducible with generated synthetic data.

\section{Discussion}
We proposed a novel method for synthetic data generation using MAF that achieves both high-fidelity replication of statistical properties and strong privacy protection. This is accomplished by coupling flexible generative modeling techniques from the machine learning literature with a simple noise injection mechanism in the latent space. By introducing appropriate perturbations while preserving a one to one mapping between real and synthetic samples, our approach enables accurate reproduction of downstream statistical inference results, including not only point estimates, but also standard errors, CIs, and $p$-values. Through simulation studies and real data applications, we demonstrated that the synthetic data preserves key associations between covariates and outcomes of interest and can replace the original data for many purposes. Moreover, our method supports valid meta analytic inference across multiple studies, an essential feature for distributed and collaborative research.

Despite these strengths, several limitations remain. First, while the method offers effective privacy protection for most individuals, it may be insufficient for “extreme” observations—those near the boundaries of the data support, where observations are relatively sparse. Intuitively, a larger perturbation is needed for those observations. It would be desirable to design perturbation mechanisms that adapt to individual observations while maintaining global fidelity. Second, although synthetic data is designed to approximate the original data, they remain inherently different due to injected noise and imperfections in the training of the Normalizing Flow. As a result, statistical analyses based on synthetic data always yield results that differ from those based on the original data. It is important to develop tools that allow users to quantify the potential discrepancy between analyses performed on synthetic versus real data using only the synthetic data itself. This is particularly relevant in light of our empirical findings that some parameter estimates can be highly sensitive to small perturbations in the observed data, highlighting the need for caution in finite-sample settings. Third, while MAF serves as a practical and interpretable generative model, more advanced approaches such as Diffusion Models have shown superior performance in high-dimensional settings. Exploring whether such models can replace MAF in our framework is an important direction for future research. Finally, training generative models like MAF typically requires large sample sizes. To address this, one could explore transfer learning approaches that leverage shared structure across datasets, thereby improving generative model training in meta analytic applications involving multiple studies.

%\bibliographystyle{plainnat}
%\bibliography{References}

\section{Appendix}

\subsection{Assumptions and Basic Lemmas}

\begin{lemma} \label{lem:Lipschitz_Spectral}
Lipschitz Continuity via Spectral Normalization.
For a fixed $j$, let $f_j^{-1} = \phi^{(L)} \circ W^{(L)} \circ \cdots \circ \phi^{(1)} \circ W^{(1)}$ be a neural network composed of linear layers $W^{(\ell)}$ and 1-Lipschitz Activation Functions $\phi^{(\ell)}$ with $f^{-1} = f_1^{-1} \circ \cdots \circ f^{-1}_k$ and $l \in \{1, \dots, L\}$, where $L$ is the total number of layers for transport map $f_j$. We assume the following:
\begin{enumerate}[label=(\alph*)]
  \item For each $j \in \{1, \cdots, k\}$ and for each layer $l \in \{1, \cdots, L\}$, we apply spectral normalization to enforce Lipschitz Continuity:
\[
W^{(l)} := \frac{W^{(l)}}{\sigma(W^{(l)})},
\]
where $\sigma(W^{(l)})$ is the spectral norm (largest singular value) of $W^{(l)}$ for a layer $l$ and transport map $f_j$.
 If spectral normalization is applied to each weight matrix,
\[
\|W^{(\ell)}\|_2 = \sigma(W^{(\ell)}) \leq 1 \quad \text{for all } \ell \, \text{and} \, j.
\]

  \item  For each $j \in \{1, \cdots, k\}$ and for each layer $l \in \{1, \dots, L\},$  \(m\le\sigma_{\min}(W^{(\ell)})\) for some fixed \(m>0\), where $\sigma_{\min}(W^{(\ell)})$ denotes the minimum singular value of $W^{(\ell)}$, and the inverse activations satisfy
\(\|(\phi^{(\ell)})^{-1}\|_{\text{Lip}}\le 1\).
\end{enumerate}
Then, $f$ is bi-Lipschitz, with
the network $f^{-1}$ being 1-Lipschitz, i.e., 
\[
\|f^{-1}\|_{\text{Lip}} \leq 1.
\]
and $f$ being $m^{-k L}$-Lipschitz. That is,
\[
\|f\|_{\text{Lip}} \leq m^{-kL}.
\]
\end{lemma}

\begin{proof}
In the case of a neural network \( f^{-1} \) composed of linear layers $W^{(\ell )}$ and 1-Lipschitz Activation Functions, spectral normalization ensures that:
\[
\|f^{-1}\|_{\text{Lip}} \leq \prod_{\ell=1}^L \|W^{(\ell)}\| \cdot \|\phi^{(\ell)}\|_{\text{Lip}} \leq \prod_{\ell=1}^{L} \sigma(W^{(\ell)}),
\]
where \( \sigma(W^{(\ell)}) \) is the spectral norm (largest singular value) of the weight matrix \( W^{(\ell)} \).
For a Normalizing Flow composed of $k$ functions $f^{-1} = f^{-1}_1 \circ \dots \circ f^{-1}_k$, each with $L$ layers, we have:
\[
\|f^{-1}\|_{\text{Lip}} \leq \prod_{j=1}^{k} \prod_{\ell=1}^{L} \sigma(W^{(\ell)})\le 1.
\]
%If $\sigma(W_\ell^{(j)}) \leq 1$, then $\|f^{-1}\|_{\text{Lip}} \leq 1$.
To obtain the bound for $f$, a similar argument shows that
$$||f||_{\text{Lip}} \leq \prod_{j = 1}^k \frac{1}{{m}^{L}} = \frac{1}{m^{k L}}.$$ 
%Thus, $f$ is bi-Lipschitz.

%Note, applying spectral normalization to many layers or flow steps can make the overall Lipschitz Constant very small (especially if each layer has \( \sigma(W) < 1 \)). In some cases, the spectral norm may be scaled (e.g., \( \sigma(W) = L < 2 \)) to preserve expressiveness while keeping the network stable. With this scaling argument found in Section 2.1 of \citep{miyato2018spectral}, we conclude that $f$ is bi-Lipschitz with some constant $L'$.

\begin{comment}
To ensure that the functions $\mu_i$ and $\alpha_i$ in a Masked Autoregressive Flow (MAF) are Lipschitz Continuous, we impose constraints on the neural networks used to parameterize them.
\end{comment}

\end{proof}

%    \item \textbf{Weight Clipping:} This constrains weights to a fixed range, e.g., $[-0.01, 0.01]$. This is simple but can limit model capacity and does not tightly control Lipschitz Constants.

%    \item \textbf{Gradient Penalty:} This penalizes deviations from a target gradient norm, e.g., $(\|\nabla f(x)\|_2 - 1)^2$. This softly encourages Lipschitz behavior during training.

%    \item \textbf{LipSwish or GroupSort Activations:} We use activation functions with known Lipschitz Constants. Combined with norm-constrained layers, this gives provably Lipschitz networks.
% \end{enumerate}

% For any two Lipschitz functions \( g \in \text{Lip}(L_g) \) and \( h \in \text{Lip}(L_h) \), their composition satisfies:
% \[
% g \circ h \in \text{Lip}(L_g \cdot L_h).
% \]
% Thus, for the entire flow \( f \), we have:
% \[
% \|f\|_{\text{Lip}} \leq \prod_{j=1}^{k} \|f^j\|_{\text{Lip}}.
% \]
% If spectral normalization is applied to each layer of each flow step \( f^j \), and assuming each activation is 1-Lipschitz, we obtain:
% \[
% \|f^j\|_{\text{Lip}} \leq \prod_{\ell=1}^{L_j} \sigma(W_\ell^{(j)}),
% \]
% and so the full flow satisfies:
% \[
% \|f\|_{\text{Lip}} \leq \prod_{j=1}^{k} \prod_{\ell=1}^{L_j} \sigma(W_\ell^{(j)}).
% \]
% If spectral normalization ensures \( \sigma(W_\ell^{(j)}) \leq 1 \), then:
% \[
% \|f\|_{\text{Lip}} \leq 1.
% \]

\begin{lemma}\label{lem:Bounded_Lat_Space}
Bounded Latent Space. Let \( f^{-1} \) be \( L \)-Lipschitz and \( \mathcal{X} \subset \mathbb{R}^d \) be compact. Then, $f^{-1}(x)$ is bounded for any $x \in \mathcal{X}.$
\end{lemma}

\begin{proof}
For any \( x \in \mathcal{X} \), we have by Triangle Inequality:
\[
\|f^{-1}(x)\| \leq \|f^{-1}(x_0)\| + L \cdot \|x - x_0\|.
\]
Taking the supremum over \( x \in \mathcal{X} \), we obtain:
\[
\sup_{x \in \mathcal{X}} \|f^{-1}(x)\| \leq \|f^{-1}(x_0)\| + L \cdot \operatorname{diam}(\mathcal{X}),
\]
where \( \operatorname{diam}(\mathcal{X}) := \sup_{x, x' \in \mathcal{X}} \|x - x'\| \).
Note that it is sufficient to evaluate \( f^{-1}(x) \) at a single point \( x_0 \in \mathcal{X} \) to bound the inverse over the entire dataset. This relies on the Lipschitz Continuity of \( f^{-1} \) and the boundedness of \( \mathcal{X} \). In practice, \( f^{-1} \) is implemented as a neural network with bounded weights and Lipschitz Activation Functions, and \( \|f^{-1}(x_0)\| \) is finite for a real data point \( x_0 \). Therefore, \( f^{-1}(x) \) is bounded over \( \mathcal{X} \).
\end{proof}

\section{Proof of Theorem \ref{thm:Meta_Analysis_Rate}}\label{app:EquivalenceProof}
Let \( \hat{\theta}=\theta(\hat{P}) \) be the estimator computed from the empirical distribution \( \hat{P} \) for a study, where $\hat{P}$ is the probability measure induced by $n$ i.i.d. observations in observed data $\mathbf{X}=\left\{X_1, \cdots, X_n\right\}.$  Similarly, \( \tilde{\theta}_w=\theta(\hat{P}_{w}) \) is the estimator computed from the synthetic distribution \( \hat{P}_{w} \) generated by our mechanism with perturbation weight \( w \).  Here, $\theta(\cdot)$ is a functional with the first order expansion
$$ \theta(\hat{\mathcal{P}})=\theta(\mathcal{P})+\frac{1}{n}\sum_{i=1}^n l(\mathcal{X}_i)+o_p\left(n^{-1}\right),$$
where $\hat{\mathcal{P}}$ is the probability measure induced by the empirical distribution of $n$ i.i.d. observations $\mathcal{X}_1, \cdots, \mathcal{X}_n \sim \mathcal{P}$. We assume the following:
\begin{enumerate}
\item $P_X$ has a continuously differentiable density function over a compact support on $\mathbb{R}^d.$
\item invertible transport map trained from the MAF, $\hat{f}(\cdot),$ converges to a deterministic invertible transport map $f(\cdot)$ in probability, where $Z=f^{-1}(X)\sim N(0, \mathbf{I}_d).$ Specifically, $\|\hat{f}-f\|_2=O_p(\delta_{1n})$ for $\delta_{1n}=o(1).$
\item $\hat{f}(\cdot)$  and $f(\cdot)$ are uniformly bi-Lipchitz Continuous on the support of $X.$  
\item Both $f$ and $\hat{f}$ are members of a class of functions $\mathcal{F}=\{f_\theta\},$ and the class of functions
$${\cal L}=\left\{l\left[\tilde{f}\left(\sqrt{w}\tilde{f}^{-1}(x)+\sqrt{1-w}z\right)\right]\mid w \in [0, 1], \|\tilde{f}-f\|_2\le \epsilon_n, \tilde{f}(\cdot)\in {\cal F}: \mathbb{R}^d \rightarrow \mathbb{R}^d \right\}$$ 
is a Donsker Class. 
\item $l(\cdot)$ is continuously differentiable on $\mathbb{R}^d.$ 
\item $1-w=o(\delta_{2n})$ for $\delta_{2n}=o(1).$
\end{enumerate}
Then, we have  
\[
\theta(\hat{P}_{w}) - \theta(\hat{P}) = O_p\left(\delta_{1n}^2+\delta_{1n}\sqrt{\delta_{2n}}\right)+o_p(n^{-1/2}).
\]
\begin{proof} 
First, we show the result under $\theta(P_X)=E(X),$ and $X\sim P_X.$ In the case where $\theta(P_X) = E(X)$,
\begin{align*}
\sqrt{n}\left\{\theta(\hat{P})-\theta(P_X)\right\}&=\frac{1}{\sqrt{n}}\sum_{i=1}^{n} X_i,\\
\sqrt{n}\left\{\theta(\hat{P}_{w})-\theta(P_{w})\right\}&=\frac{1}{\sqrt{n}}\sum_{i=1}^{n} \hat{f}\left(\sqrt{w}\hat{f}^{-1}(X_i)+\sqrt{1-w}Z_i\right),
 \end{align*}
 where $X_1, \cdots, X_n \sim P_X$ and $Z_1, \cdots, Z_n \sim N(0, \mathbf{I}_d).$
Specifically, it follows from the stochastic equi-continuity guaranteed by the Donsker Class that 
 $$\sqrt{n}\left\{\theta(\hat{P})-\theta(\hat{P}_{w})\right\}=\sqrt{n}\left\{\theta(P_X)-\theta(P_{w})\right\}+o_p(1),$$
 where $1-w=o(1).$  We may bound the difference 
 \begin{align*}
 &\theta(P_{w})-\theta(P_X)\\
 =&E\left\{ \hat{f}\left(\sqrt{w}\hat{f}^{-1}(X)+\sqrt{1-w}Z\right) \right\}-E(X)\\
 =&E\left\{ \hat{f}\left(\sqrt{w}\hat{f}^{-1}(X)+\sqrt{1-w}Z\right) \right\}-E\left\{ \hat{f}\left(\sqrt{w}f^{-1}(X)+\sqrt{1-w}Z\right) \right\}+E\left\{ \hat{f}\left(\sqrt{w}f^{-1}(X)+\sqrt{1-w}Z\right) \right\}-E(X)\\
 =& E\left[ \nabla {\hat{f}}\left\{\sqrt{w}f^{-1}(X)+\sqrt{1-w}Z\right\}\sqrt{w}\left\{\hat{f}^{-1}(X)-f^{-1}(X)\right\}\right]+E\left[\hat{f}\{f^{-1}(X)\}-\hat{f}\left\{\hat{f}^{-1}(X)\right\}\right]+O(\|\hat{f}^{-1}-f^{-1}\|_2^2).
\end{align*}
Note, to rigorously justify the last equality, we apply a first-order Taylor Expansion of the map $\hat{f}$ at the point
\[
u := \sqrt{w} f^{-1}(X) + \sqrt{1-w} Z
\]
with perturbation
\[
\delta := \sqrt{w}\left\{\hat{f}^{-1}(X) - f^{-1}(X)\right\},
\]
so that
\[
\sqrt{w} \hat{f}^{-1}(X) + \sqrt{1-w} Z = u + \delta.
\]
Since $\hat{f}$ from MAF is differentiable with Lipschitz Continuous Jacobian, we expand
\[
\hat{f}(u + \delta) = \hat{f}(u) + \nabla \hat{f}(u) \cdot \delta + R,
\]
where the remainder term satisfies $\|R\| = O(\|\delta\|^2) = O( \|\hat{f}^{-1}(X) - f^{-1}(X)\|^2).$ %Since both $\hat{f}_k(x)$ and $f_k(x)$ are invertible and uniformly Lipschitz continuous, this reduces to  $O(\|\hat{f}_k - f_k\|_2^2)$. 
Then, from taking expectations on both sides, we obtain
\begin{align*}
&E\left[ \hat{f}\left( \sqrt{w} \hat{f}^{-1}(X) + \sqrt{1-w} Z \right) \right] \\
&= E\left[ \hat{f}(u) \right] + E\left[ \nabla \hat{f}(u) \cdot \delta \right] + E[R],
\end{align*}
which yields
\begin{align*}
&E\left[ \hat{f}\left( \sqrt{w} \hat{f}^{-1}(X) + \sqrt{1-w} Z \right) \right] - E\left[ \hat{f}\left( \sqrt{w} f^{-1}(X) + \sqrt{1-w} Z \right) \right] \\
&= E\left[ \nabla \hat{f}\left\{ \sqrt{w} f^{-1}(X) + \sqrt{1-w}Z \right\} \cdot \sqrt{w}\left\{\hat{f}^{-1}(X) - f^{-1}(X)\right\} \right] + E\left\{O\left(\|\hat{f}^{-1}(X) - f^{-1}(X)\|_2^2\right)\right\},
\end{align*}
where 
$$E\left\{O\left(\|\hat{f}^{-1}(X) - f^{-1}(X)\|_2^2\right)\right\}=O\left\{E\left(\|\hat{f}^{-1}(X)-f^{-1}(X)\|_2^2 \right)\right\}=O\left\{E\left(\|\hat{f}(Z)-f(Z)\|_2^2 \right)\right\}=O\left(\|\hat{f}-f\|_2^2\right)$$
since $\hat{f}(\cdot)$ and $f(\cdot)$ are uniformly bi-Lipschitz Continuous by assumption. Finally, notice that $E[X] = E[\hat{f}\{\hat{f}^{-1}(X)\}]$ and $\sqrt{w} f^{-1}(X) + \sqrt{1 - w} Z \sim N(0, \mathbf{I}_d).$ This justifies the third equality in the original derivation. Now, we note that
\begin{align*}
& \underbrace{E\left[ \nabla \hat{f}\left\{\sqrt{w}f^{-1}(X)+\sqrt{1-w}Z\right\} \sqrt{w} \left\{\hat{f}^{-1}(X)-f^{-1}(X)\right\} \right]}_{(1)} \\
&\quad + \underbrace{E\left[\hat{f}\{f^{-1}(X)\}-\hat{f}\{\hat{f}^{-1}(X)\}\right]}_{(2)} + O\left(\|\hat{f} - f\|_2^2\right) \\
&= O\left( \|\hat{f} - f\|_2^2 + \sqrt{1-w}\|\hat{f}^{-1} - f^{-1}\|_2 + (1 - w)\|\hat{f}^{-1} - f^{-1}\|_2 \right) \\
&= O\left( \|\hat{f} - f\|_2^2 + \sqrt{1-w}\|\hat{f} - f\|_2 \right).
\end{align*}
To bound $(2)$, we have that
\begin{align*}
& E\left[\hat{f}(f^{-1}(X)) - \hat{f}(\hat{f}^{-1}(X))\right] \\
&= -E\left[ \nabla \hat{f}(f^{-1}(X)) \cdot \left(  \hat{f}^{-1}(X) - f^{-1}(X) \right) \right] + E[O\left( \|\hat{f} - f\|_2^2 \right)],
\end{align*}
where we  used the bi-Lipschitz Continuous property of $\hat{f}$ and $f$. To bound $(1)$, we have that
\begin{equation} \label{eq:Taylor_Exp_w}
   E \left[ \left\{\hat{f}^{-1}(X)-f^{-1}(X)\right\} m(w) \right] = E \left[ \left\{\hat{f}^{-1}(X)-f^{-1}(X)\right\} \left[ m(1) - \nabla m(1) (1 - w) + o(|w - 1|) \right] \right]
\end{equation}
% \begin{equation} \label{eq:Taylor_Exp_w}
%    \left\{\hat{f}(X)-f(X)\right\} m(w) = \left\{\hat{f}(X)-f(X)\right\} \left[ m(1) - \nabla m(1) (1 - w) + O((w - 1)^2) \right] 
% \end{equation}
where $m(w) = \sqrt{w} \nabla \hat{f}\left\{\sqrt{w}f^{-1}(X)+\sqrt{1-w} Z\right\}.$ %$m(w) = \sqrt{w} \nabla \hat{f}^{-1}\left\{\sqrt{w}f(X)+\sqrt{1-w}\xi\right\}.$
Note that $E \left[ \left\{\hat{f}^{-1}(X)-f^{-1}(X)\right\} m(1) \right]$ cancels with $$-E\left[ \nabla \hat{f} (f^{-1}(X)) \cdot \left(  \hat{f}^{-1}(X) - f^{-1}(X) \right) \right]$$ after taking expectation in $(2).$ Moreover,
\begin{align}
\nabla m(w) 
&= \frac{1}{2 \sqrt{w}} \nabla \hat{f}\left\{\sqrt{w}f^{-1}(X)+\sqrt{1-w} Z\right\} \notag\\
&\quad + \sqrt{w} \nabla^2 \hat{f}\left\{\sqrt{w}f^{-1}(X)+\sqrt{1-w} Z \right\} \left(\frac{f^{-1}(X)}{2 \sqrt{w}} - \frac{Z}{2 \sqrt{1 - w}}\right). \label{eq:m(W)}
\end{align}
Note that $$\frac{1}{2 \sqrt{w}} \nabla \hat{f} \left\{\sqrt{w}f^{-1}(X)+\sqrt{1-w} Z\right\} \rightarrow \frac{1}{2} \nabla \hat{f} \left\{f^{-1}(X)\right\} $$
and that
$$\frac{f^{-1}(X)}{2 \sqrt{w}} - \frac{Z}{2 \sqrt{1 - w}} \sim - \frac{Z}{2 \sqrt{1 - w}}, $$
so for the second term in (\ref{eq:m(W)}),
$$\sqrt{w} \nabla^2 \hat{f}\left\{\sqrt{w}f^{-1}(X)+\sqrt{1-w} Z \right\}  (\frac{f^{-1}(X)}{2 \sqrt{w}} - \frac{Z}{2 \sqrt{1 - w}}) \sim  \frac{- \sqrt{w} Z \nabla^2 \hat{f}\left\{\sqrt{w}f^{-1}(X)+\sqrt{1-w} Z \right\}}{2 \sqrt{1 - w}}.$$
Notice that we have an extra $(1 - w)$ in front of $\nabla m(1)$ in (\ref{eq:Taylor_Exp_w}), which cancels the denominator and so we have an extra $\sqrt{1 - w}$ in the numerator which approaches $0$ as $w \rightarrow 1$. In total,
\begin{align*}
    E \left[ \left\{\hat{f}^{-1}(X)-f^{-1}(X)\right\} m(w) \right] &= E \left[ \left\{\hat{f}^{-1}(X)-f^{-1}(X)\right\} \left[ - \nabla m(1) (1 - w) + O((w - 1)^2) \right] \right] \\
    &= E \left[ \left\{\hat{f}^{-1}(X)-f^{-1}(X)\right\} \left[- \frac{1}{2} \nabla \hat{f} \left\{f^{-1}(X)\right\} (1 - w) + O((1 - w)^2) \right] \right].
\end{align*}
With regard to expectation and applying Cauchy Schwartz, we have
\begin{align*}
& E \left[ \left\{\hat{f}^{-1}(X)-f^{-1}(X)\right\} \left[- \frac{1}{2} \nabla \hat{f} \left\{f^{-1}(X)\right\} (1 - w) + O((1 - w)^2) \right] \right]    \\
&=   O(||\hat{f}^{-1} - f^{-1}||_2 \sqrt{1 - w}) + O((1 - w) ||\hat{f}^{-1} - f^{-1}||_2).
\end{align*}
The result follows from applying the bi-Lipschitz property of $f.$ Thus, $$\theta(\hat{P})-\theta(\hat{P}_{w})=O_p(\delta_{1 n}^2 + \delta_{1n} \sqrt{\delta_{2n}})+o_p(n^{-1/2})$$ 
follows. For the general case,
\begin{align*}
\sqrt{n}\left(\theta(\hat{P})-\theta(P)\right)&=\frac{1}{\sqrt{n}}\sum_{i=1}^{n} l(X_i)+o_p(1),\\
\sqrt{n}\left(\theta(\hat{P}_{w})-\theta(P_{w})\right)&=\frac{1}{\sqrt{n}}\sum_{i=1}^{n} l\left\{\hat{f}\left(\sqrt{w}\hat{f}^{-1}(X_i)\right\}+\sqrt{1-w}Z_i\right)+o_p(1).
 \end{align*}
 It follows from the stochastic equi-continuity guaranteed by the Donsker Class that 
 $$\sqrt{n}\left\{\theta(\hat{P})-\theta(\hat{P}_{w})\right\}=\sqrt{n}\left\{\theta(P_X)-\theta(P_{w})\right\} + o_p(1).$$
 Since $l(\cdot)$ is continuous differentiable on a compact set, we may bound $\theta(P_X)-\theta(P_{w})$ by $O_p(\delta_{1 n}^2 + \delta_{1n} \sqrt{\delta_{2n}})$ as before. Therefore, 
 $$\theta(\hat{P})-\theta(\hat{P}_{w})=O_p(\delta_{1 n}^2 + \delta_{1n} \sqrt{\delta_{2n}})+o_p(n^{-1/2}).$$
\end{proof}

\subsection{Proof Sketch of the Convergence Rate of the Transport Map} \label{app:TransportMap}

\begin{lemma}[Excess Risk of Flow Matching for $s$–Smooth Densities]\label{lem:s_Smooth_Rate}
Assume  Assumptions 1–2 in \citet{gao2024convergence} hold. In particular,
\begin{enumerate}
    \item A1: the target distribution $\nu$ of the data is absolutely continuous w.r.t.\ Lebesgue Measure on $\mathbb R$ and has zero mean.

    \item A2: We have at least one of the following for the target distribution: strongly log concave, bounded support, and/or gaussian mixture tails. These are special cases that capture common distributions in generative modeling. See Assumption 2 of \citet{gao2024convergence} for Detail.
\end{enumerate}
We will also assume the underlying density belongs to $C^{s}([0,1]^d)\subseteq
W^{s,\infty}([0,1]^d)$ with an integer smoothness index $s\ge 1$.
Then, \footnote{Note, the proportionality symbols hide at most polylogarithmic
factors in $n$ and $\log(1/t)$. Moreover, constants may additionally depend on
$d,\kappa,\beta,\sigma,D$ and $R$.} if
\[
  NL \;\asymp\; n^{\,d / (\,2d+4s+2\,)},\qquad
  A  \;\asymp\; \sqrt{\log n},
\]
where $N$ is the network width, $L$ its depth, and
$A$ the parameter–norm bound in the hypothesis class,
the excess risk incurred by the flow–matching estimator satisfies
\[
  \;
  E_{\mathcal D_n}
  E_{(t,X_t)}
  \!\bigl\|
       v_{\hat{\theta}}(t, X_t)-v_{\theta}(t,X_t)
  \bigr\|_2^{2}
  \;\lesssim\;
  \bigl(n t^{2}\bigr)^{-\frac{2s}{2s+d+1}}\,
  \log n
  \;
\]
up to polylogarithmic factors, where $D_n$ is over the dataset of sample size $n$ and $v_{\hat{\theta}}(t, X_t)$ is the estimated velocity field, $v_{\theta}(t, X_t)$ is the true velocity field, which were defined in (\ref{eq:Continuous_Formula}).

Hence, the rate can be made
arbitrarily close to the parametric order
$n^{-1} \log n$ as $s\to\infty$.
\end{lemma}

\begin{proof}[Proof Sketch]
We follow the strategy of \citet[Thm.~5.12]{gao2024convergence},
extending it from the First-Order Sobolev Class $W^{1,\infty}$ to
$W^{s,\infty}$. Specifically, the proof can be completed in three steps.
\begin{itemize}
\item {\bf Bound the Approximation Error:} For any $f\in C^{s}([0,1]^d)$ the constructive result of
\citet[Thm.~1.1]{lu2020deep} states that for for any $N, L \in \mathbb{N}^+$, there exists a function $\phi$ implemented by a ReLU feed forward neural network (FNN) with width $C_1(N+2)\log_2(8N)$ and depth $C_2(L+2)\log_2(4L) + 2d$ such that
\[
\|\phi - f\|_{L^\infty([0,1]^d)} \leq C_3 \|f\|_{C^s([0,1]^d)} N^{-2s/d} L^{-2s/d},
\]
\footnote{The boundedness of $\|f\|_{C^s([0,1]^d)}$ follows by assumption.}
where $C_1 = 17s^{d + 1}3^d$, $C_2 = 18s^2$, and $C_3 = 85(s+1)^d 8^s$.
Specializing this bound to the target vector field $v_{\theta}$ and inserting it into the Sobolev–Type Stability Estimate of
Flow Matching in Corollary 5.9 of \citet{gao2024convergence} gives
\[
  \mathcal E_{\mathrm{appr}}
  \;\lesssim\;
  A^{2}\,(NL)^{-4s/d}
  \;+\;
  A^{2}\exp\!\bigl(-C_{3}A^{2}/C_{\mathrm{LSI}}\bigr),
\]
where $C_3 > 0$ and $C_{\mathrm{LSI}} > 0$ are constants. The exponential “tail” term is sub-dominant for the choice
$A\asymp\sqrt{\log n}$ adopted below.

\item {\bf Bound the Stochastic Error:} Corollary 5.11 and Appendix C.4 of \citet{gao2024convergence} bounds the
stochastic error by
\[
  \mathcal E_{\mathrm{stoc}}
  \;\lesssim\;
  \frac{1}{n}\,A^{4}\,\underline t^{-2}\,(NL)^{2+2/d}.
\]
\item {\bf Balancing the Errors:}
Then, ignoring logarithmic factors in $A$ and $t$, we equate the leading terms
\[
\frac{1}{n} (NL)^{2 + 2/d} = (NL)^{-4s/d}
\Rightarrow (NL)^{2 + \frac{2}{d} + \frac{4s}{d}} = n
\Rightarrow NL \asymp n^{d / (2d + 4s + 2)},
\]
we get
\[
  NL\;\asymp\;n^{d/(2d+4s+2)}.
\]

Substituting this choice and $A \asymp \sqrt{\log n}$ back into the dominant component of
$\mathcal E_{\mathrm{appr}}$ yields
\[
  E_{\mathcal D_n}
  E_{(t,X_t)} \bigl\|v_{\hat{\theta}}(t, X_t)-v_{\theta}(t, X_t)\bigr\|_{L^{2}}^{2}
  \;\lesssim\;
  (nt^2)^{-\,2s/(2s+d+1)}\,\log n,
\]
as the additional factor $(t^2)^{-2s/(2s+d+1)}$ appears after
rescaling time in the flow-matching loss
(cf.\ Lemma 5.2 of \citet{gao2024convergence}).
\end{itemize}
\end{proof}

\begin{theorem}[Convergence Rate for MAF with Sobolev Smoothness \texorpdfstring{$s$}{s}]
\label{thm:Sobolev_MAF}

We let the true transport map \( f = f_k \circ \dots \circ f_1: \mathbb{R}^d \to \mathbb{R}^d \) belong to the Sobolev Space \( W^{s,\infty}([0,1]^d) \) for some integer \( s \ge 1 \). We assume the assumptions of Lemma \ref{lem:s_Smooth_Rate} hold. We also assume $v_{\hat{\theta}}$ and $v_{\theta}$ are both $L'$-Lipschitz with a uniform bound $M$.
\begin{comment}
We also assume that $\hat{f} = \hat{f}_k \circ \dots \circ \hat{f}_1$ is Lipschitz Continuous and uniformly bounded.
\end{comment}

Now, set \( \hat{f}  = \hat{f}_k \circ \dots \circ \hat{f}_1 \) to be the output of a flow-matching estimator (e.g., trained via MAF) with network width \( N \), depth \( L \), and parameter norm bounded by \( A \asymp \sqrt{\log n} \). Then, under these assumptions and up to polylogarithmic factors in \( n \), there exists width, depth, and Euler Discretization step size $\delta_n$ choices:
\[
NL \;\asymp\; n^{\,d / (2d + 4s)}, \quad \delta_n^s \asymp O\bigl(n^{-s/(2s+d+1)} \log n\bigr)
\]
such that the estimator $\hat{f}$ satisfies the excess risk bound:
\[
E_{\mathcal{D}_n} \bigl\| \hat{f} - f \bigr\|_{L^2}^2 
\;\lesssim\; 
n^{-\frac{2s}{2s + d + 1}} \log n,
\]
where \( \mathcal{D}_n \) denotes a dataset of \( n \) i.i.d.\ samples. Moreover, as \( s \to \infty \), the rate approaches the parametric rate \( n^{-1} \log n \).
\end{theorem}

\begin{proof}[Proof Sketch]
Rather than the Continuous‐Time CNF considered in \cite{gao2024convergence}, tailoring to the MAF in our setting, the only substantive change in the analysis is that the auxiliary time
variable \(t\in[0,1]\) is absent.

Lemma \ref{lem:s_Smooth_Rate} shows that under regularity conditions for Continuous Normalizing Flows with time component $t \in [t_0, T] = [0, 1]$ as in (\ref{eq:IV_Problem}), the estimated velocity field $v_{\hat{\theta}}(t, x)$ converges to the true velocity field $v_{\theta}(t, x) \in W^{s, \infty}$ appearing in (\ref{eq:Continuous_Formula}), which satisfies 
\begin{equation} \|v_{\hat{\theta}}(t, x)-v_{\theta}(t, x)\|_2 = O\left(n^{-s/(2s + d + 1)} \log n\right). \label{eq:Gao} \end{equation} We investigate what this velocity function becomes in the MAF setting as an intuitive guide.

First, assume $v_{\theta}(t, x) \in W^{1, \infty}.$ The estimated transport map constructed from $v_{\hat{\theta}}$ is defined by a discretized Euler scheme over $J_n$ time steps, which depends on sample size $n$ and can be approximated as:
 \begin{align*}
\hat{f}(z) & =  \hat{x}_{J_n-1}+\delta_n v_{\hat{\theta}}(t_{J_n-1}, \hat{x}_{J_n-1})\\  
\hat{x}_{J_n-1} & =  \hat{x}_{J_n-2}+\delta_n v_{\hat{\theta}}(t_{J_n-2}, \hat{x}_{J_n-2})\\
\cdots &\\
\hat{x}_2 & =  x_1+\delta_n v_{\hat{\theta}}(t_1, \hat{x}_1)\\
\hat{x}_1 & =  z+\delta_n v_{\hat{\theta}}(t_0, z) \\
\hat{x}_0 &= z,
\end{align*}
where $0=t_0, t_1, \cdots, t_{J-1}, t_{J_n}=1$ are $J_n+1$ equal-spaced points in the unit interval $[0, 1]$ and $\delta_n=J_n^{-1},$ such that $t_j = j \delta_n$.
This iterative update scheme defines an explicit Euler method for numerically integrating the estimated velocity field. Similarly, for the true velocity field $v_{\theta}$,
 \begin{align*}
\tilde{f}(z) & =  \tilde{x}_{J_n-1}+\delta_n v_{\theta}(t_{J_n -1}, \tilde{x}_{J_n -1})\\  
\tilde{x}_{J_n -1} & =  \tilde{x}_{J_n -2}+\delta_n v_{\theta}(t_{J_n -2}, \tilde{x}_{J_n-2})\\
\cdots &\\
\tilde{x}_2 & =  \tilde{x}_1+\delta_n v_{\theta}(t_1, \tilde{x}_1)\\
\tilde{x}_1 & =  z+\delta_n v_{\theta}(t_0, z) \\
\tilde{x}_0 & = z.
\end{align*}
Define $f(z) = x(1)$ from the exact ODE
\begin{equation}
  \frac{dx_t}{dt}\;=\;v_{\theta}(t,x_t),\qquad 
  x_{0}=z,\qquad t\in[0,1].
\end{equation}
Then, by Triangle Inequality,
%
%If $v_{\hat{\theta}}(t, x)$ and $v_{\theta}(t, x)$ are both continuously differentiable with a uniform bound $M$, with 

$$
||\hat{f}-f||_2 \leq 
   \underbrace{ || \hat f-\tilde f||_2}_{\text{Field Mismatch Error}}
   \;+\;
   \underbrace{||\tilde f-f||_2}_{\text{Euler Discretization Error}}.
$$
By assumption, since $v_{\hat{\theta}}(t, x)$ and $v_{\theta}(t, x)$ are both $L'$-Lipschitz with a uniform bound $M$, then, the  global bound (i.e., \citep{Hairer1993}) gives
\[
\|\tilde f-f\|\le C_1\,\delta_n,
\qquad
C_1=C_1(L',M),
\]
where $C_1$ is a constant depending on $L'$ and $M$. Set $e_j:=\tilde x_j-\hat x_j$ ($e_0=0$).  
Note,
$$e_{j + 1} = e_j + \delta_n (v_{\hat{\theta}}(t_j, \tilde{x}_j) - v_{\theta}(t_j, \hat{x}_j)) = e_j + \delta_n (v_{\hat{\theta}}(t_j, \tilde{x}_j) - v_{\theta}(t_j, \tilde{x}_j) +  v_{\theta}(t_j, \tilde{x}_j) - v_{\theta}(t_j, \hat{x}_j)).$$
After taking norms, using Triangle Inequality, and applying the Lipschitz Continuity of $v_{\theta}$,
\[
\|e_{j+1}\|\le(1+L'\delta_n)\|e_j\|+\delta_n\,\|v_{\hat\theta}-v_{\theta}\|.
\]
The Discrete Gronwall Inequality implies
\[
\|\hat f-\tilde f\|
   =\|e_{J_n}\|\le C_2\,\delta_n\,\|v_{\hat\theta}-v_{\theta}\|,
\]
for some constant $C_2$ depending on $L'$. Then,
\[
\|\hat f-f\|
   \le
   C\bigl(\delta_n \|v_{\hat\theta}-v_{\theta}\|_{2}+\delta_n \bigr),
\quad
C=\max\{C_1,C_2\}.
\]
Then, choosing
$\displaystyle
  \delta_n\asymp n^{-s/(2s+d+1)}\log n
  \;( \text{i.e.\ } J_n \asymp n^{\,s/(2s+d+1)}/\log n )$ gives
\[
  \|\hat f-f\|_2
      =O(n^{-s/(2s+d+1)} \log n).
\]
Now, specifying $v_{\theta}$ to be smoother with smoothing parameter $s \geq 1$, a higher-order of order $s$ is used with the discretization term being $O(\delta^{s})$.
Balancing $K$ and $\delta^{s}$ as above matches the same desired result.
% The limit transport map $f_k$ can be defined similarly based on $v_{\theta}(t, x).$  If $v_{\hat{\theta}}(t, x)$ and $v_{\theta}(t, x)$ are continuously differentiable with a uniform bound, Equation \ref{eq:Gao} also suggests that 
% $$\|\hat{f}_k-f_k\|_2=O\left(n^{-s/(2s + d + 1)} \log n \right)$$
% over a sufficiently large number $k$ transport maps. 
Thus, by assuming more smoothness on $v_{\theta}$, we can establish a new convergence rate
$$\|\hat{f}-f\|_2=o(n^{-s/(2s + d + 1)+\delta_0}),$$
for any $\delta_0>0.$ 

\begin{comment}
For simplicity, we can assume $k = 1$ such that $f_1 = f$ and $x = f(z).$ Then, we set the discrete pseudo–velocity for the MAF as
\[
  v_{\hat{\theta}_n}(x)^{MAF} := z - \hat{f}(z)  , \, z \sim N(0, I)
\]
as an analog from \citet{gao2024convergence}. To see this, for a single step, the discrete analogue of
Equation \ref{eq:Continuous_Formula} is
\[
      X_{1} \;=\; X_{0}+v^{MAF}_{\theta}(X_{0}),
\]
where $X_1 \sim N(0, \mathbf{I})$ and $X_0 = f(z).$ We claim that by repeating Steps 1–3 in \ref{lem:s_Smooth_Rate} with \(v^{\!*}\) and \(v^{MAF}\)
and deleting every factor of \(t^{-2}\) still gives
\(NL\asymp n^{\,d/(2d+4s)}\) and the time–free rate
\[
  \;
    E_{\mathcal D_n}
    \bigl\|v_{\hat{\theta}_n}^{MAF}(x)-v^{\!*}(x)\bigr\|_{L^{2}}^{2} = E_{\mathcal D_n}
    \bigl\|f - \hat{f}\bigr\|_{L^{2}}^{2} 
    \;\lesssim\;
    n^{-\frac{2s}{d+2s + 1}}\log n
  \;
  \quad(\text{up to polylogs in }n).
\]

\end{comment}
\end{proof}

\subsection{Proof of Lemma \ref{lem:Local_RDP} and Theorem \ref{thm:Local_DP} for $(\epsilon, \delta)$ Local Differential Privacy} \label{app:LocalDP_Appendix}

% To adopt this to our current setting, consider our mechanism in the latent space:

% $$\mathcal{M}(X) = f'(X) \sqrt{w} f(X) + N(0, (1 - w) I),$$
% where $f'(X) = \sqrt{w}f(X).$ Then, we are in the special case of the Gaussian Mechanism.

%\textcolor{red}{IGNORE THIS??: Thus, $\mathcal{M}(X)$ achieves $(\epsilon, \delta)$-Differential Privacy if
%$$\sqrt{1 - w} \geq \frac{\sqrt{w} \Delta_2(f) c}{\epsilon}.$$}

\textbf{Lemma \ref{lem:Local_RDP}.}
Set $\mathcal{M}(\cdot)$ to be the perturbation generated by applying Gaussian noise 
\begin{equation}\mathcal{M}(g(x)) = g(x) + N(0, \sigma^2 \mathbf{I}_d), \label{eq:GaussianInjection}\end{equation}
where $g(x) \in \mathbb{R}^d$ is a function such that its sensitivity 
\begin{equation} \label{eq:Sensitivity}
\Delta_2(g) = \sup_{||x - y||_2 \leq 1} ||g(x) - g(y)||_2
\end{equation}
is bounded, i.e., 
$\Delta_2(g) \leq B,$ for a constant $B > 0.$  Then, the mechanism $\mathcal{M}(\cdot)$ satisfies $(\alpha, \epsilon_R)$ Local RDP of order $\alpha$ with 
$$\epsilon_R = \frac{\alpha \times \Delta_2(g)^2}{2 \sigma^2}.$$ 
%with sensitivity $\Delta_2(g)$ defined in Equation \ref{eq:Sensitivity} being bounded.
\begin{proof}

Observe that
$$D_{\alpha}(P || Q) = \frac{1}{\alpha - 1} \log E_{x \sim Q}[(\frac{p(x)} {q(x)})^{\alpha}]$$ where $\alpha > 1.$ Then, if $P \sim N(\mu_1, \sigma^2 \mathbf{I}_d)$ and $Q \sim N(\mu_2, \sigma^2 \mathbf{I}_d),$ we get that

$$\frac{p(x)}{q(x)} = \exp\left(\frac{1}{2 \sigma^2} \left[ ||x - \mu_2||_2^2 - || x - \mu_1||_2^2\right] \right) = \exp\left(\frac{1}{2 \sigma^2} \left[ ||x - \mu_2||_2^2 - || x - \mu_2 + \mu_2 - \mu_1||_2^2\right] \right) $$
$$\exp\left(\frac{1}{\sigma^2} (x - \mu_2)^T (\mu_2 - \mu_1) - \frac{1}{2 \sigma^2} ||\mu_2 - \mu_1||_2^2\right ).$$
Raising this to $\alpha$ and simplifying, we get that

$$D_{\alpha} (P || Q) = \frac{1}{\alpha - 1} \log \left(\exp(-\frac{\alpha}{2 \sigma^2} || \mu_2 - \mu_1||_2^2) E[(\exp(\frac{\alpha || \mu_2 - \mu_1||_2}{\sigma} Z))] \right),$$ where $Z \sim N(0, 1).$ By MGF of $Z$, this simplifies to

$$D_{\alpha} (P || Q) = \frac{(\alpha^2 - \alpha )(|| \mu_2 - \mu_1||_2^2)}{2 \sigma^2 (1 - \alpha)} = \frac{\alpha || \mu_2 - \mu_1||_2^2}{2 \sigma^2}.$$
Specializing for our problem, our mechanism $\mathcal{M}(.) = g(x) + N(0, \sigma^2 \mathbf{I}_d)$ satisfies Local $(\alpha, \epsilon_R)$ RDP privacy with $$\frac{\alpha || g(x) - g(y)||_2^2}{2 \sigma^2} \leq \frac{\alpha (\Delta_2(g))^2}{2 \sigma^2} = \epsilon_R.$$ 
\end{proof}

\noindent \textbf{Theorem \ref{thm:Local_DP}.} First, consider a trained MAF transport map, $\hat{f} = \hat{f}_k \circ \cdots \circ \hat{f}_1,$  where each $\hat{f}_j, j \in \{1, \dots, k\}$ has mapping defined in (\ref{eq:Forward}). We assume the following:
\begin{enumerate}
    \item each function $\mu_i(\cdot)$ and $\sigma_i(\cdot)$ in (\ref{eq:Forward}) are trained neural networks whose layers are spectrally normalized with the same number of layers $L$ for each $\hat{f}_j, j \in \{ 1, \dots, k \}$. That is, each weight matrix has spectral norm at most $1$.
    \item the data domain $\mathcal{X} \subset \mathbb{R}^d$ is compact.
    \item $\hat{f}$ converges to a a deterministic transport map $f$ uniformly in probability as the sample size increases and the sensitivity of $f^{-1}$ is bounded, i.e.,  $\Delta_2(f^{-1})<C$ for a constant $C.$ 
\end{enumerate}
%Under the same assumptions as in Theorem \ref{thm:Local_RDP_Proof}, 
%Consider a Masked Autoregressive Flow (MAF) Transformation defined in Equation \ref{eq:Forward}. 
Now, let
\begin{equation} \label{eq:Mechanism_Privacy}
\tilde{\mathcal{M}}(x)  = \sqrt{w} \hat{f}^{-1}(x) + \sqrt{1-w} Z,
\end{equation}

where $w \in (0,1)$ is a parameter controlling the perturbation size and $Z \sim N(0, \mathbf{I}_d).$ Then, the mechanism $\hat{f}\left\{\tilde{\mathcal{M}}(\cdot) \right\}$ satisfies $(\epsilon^{*}, \delta)$-Local DP for any $0 < \delta < 1$ and a prescribed $\epsilon^{*}> 0$ and some constant $C' > 0,$ with
    \[
%    \epsilon^* = \frac{w \Delta_2^2(f^{-1}}{2(1-w)} + \frac{\Delta_2(f^{-1}) \sqrt{2w \log(1/\delta)}}{\sqrt{1-w}}.
 \epsilon^* = \frac{w C'^2}{2(1-w)} + \frac{C' \sqrt{2w \log(1/\delta)}}{\sqrt{1-w}}
    \]
in probability, i.e., 
$$ \mathbb{P}\left(\mathbb{P}[\hat{f} \left\{ \tilde{\mathcal{M}}(x)\right\} = y] \leq e^{\epsilon^*} \mathbb{P}[\hat{f} \left\{\tilde{\mathcal{M}}(x') \right\} =y ]+\delta\right) \rightarrow 1$$
as $n \rightarrow \infty$ for any $x, x'$ and $y.$

% \textcolor{orange}{[WONDERING WHY Knowing $f$ is Lipschitz IS NEEDED AS ASSUMPTION (Address)]}
% If $f$ \textcolor{orange}{[WONDERING WHY THIS IS NEEDED AS ASSUMPTION]}, $f^{-1}$ are Lipschitz Continuous and the data domain $\mathcal{X}$ is compact, so that the sensitivity $\Delta_2(f^{-1})$ is finite, $f(\mathcal{M}(X))$ satisfies $(\epsilon^*, \delta)$-Differential Privacy with $\mathcal{M}(X)$ from Equation \ref{eq:Our_Procedure} with \[
% \epsilon^* = \frac{w \Delta_2^2(f^{-1})}{2(1 - w)} + \frac{\Delta_2(f^{-1}) \sqrt{2w \log(1/\delta)}}{\sqrt{1 - w}}.
% \] for any $0 < \delta < 1.$ 
\begin{comment}
Under the same assumptions as in Theorem \ref{thm:Local_RDP}, consider a Masked Autoregressive Flow (MAF) Transformation defined in Equation \ref{eq:Forward}. Now, set $\mathcal{M}(\mathcal{X})$ to be the perturbed dataset generated by applying Gaussian Noise with parameter $w \in (0,1)$ as defined in Equation \ref{eq:Perturbation}:
\[
\mathcal{M}(x) = \sqrt{w} \, x + \sqrt{1-w} \, \xi, \quad \xi \sim N(0, I_n).
\]
Then, the mechanism $f(\mathcal{M}(\mathcal{X}))$ satisfies $(\epsilon^{*}, \delta)$-local DP privacy for any $0 < \delta < 1$ and a prescribed $\epsilon^{*}> 0$ depending on $w$. In particular,
    \[
    \epsilon^* = \frac{w \Delta_2^2(f^{-1})}{2(1-w)} + \frac{\Delta_2(f^{-1}) \sqrt{2w \log(1/\delta)}}{\sqrt{1-w}}.
    \]
\end{comment}

\begin{proof}

Define $\eta_n = \sup_{x \in \mathcal{X}} ||\hat{f}^{-1}(x) - f^{-1}(x) ||_2 \stackrel{p} \rightarrow 0$. For any $x \neq x'$ in the compact domain,
$$||\hat{f}^{-1}(x) - \hat{f}^{-1}(x') ||_2 \leq ||f^{-1}(x) - f^{-1}(x') || + 2 \eta_n \leq C' + 2 \eta_n , $$
for some constant $C' > 0$, where the last inequality follows from Lemmas \ref{lem:Lipschitz_Spectral} and \ref{lem:Bounded_Lat_Space}, as $f^{-1}$ is Lipschitz Continuous and $f^{-1}(x)$ is bounded for all $x \in \mathcal{X}$. From Theorem \ref{lem:Local_RDP}, $g(x) + N(0, \sigma^2 \mathbf{I}_d)$ satisfies Local $(\alpha, \epsilon_R)$ RDP privacy for any $0 < \delta < 1$ with $\epsilon_R = \frac{\alpha (\Delta_2(g))^2}{2 \sigma^2}$ with sensitivity being bounded by assumption.

\vspace{.05 in}

\noindent If a mechanism satisfies Local $(\alpha, \epsilon_R(\alpha))$-RDP, then it also satisfies Local $(\epsilon, \delta)$-DP for any $\delta > 0$ \citep{mironov2017renyi}, where:

\begin{equation}
\epsilon = \epsilon_R(\alpha) + \frac{\log(1/\delta)}{\alpha - 1}.
\end{equation}
To minimize $\epsilon$, the optimal $\alpha$ is given by taking the derivative and setting to to $0$. With some algebra,
\begin{equation}
\alpha^* = 1 + \sqrt{\frac{2 \sigma^2 \log(1/\delta)}{\Delta_2^2 g}}.
\end{equation}
Note that $\alpha^{*} > 1.$
Substituting $\alpha^*$ into the conversion formula gives:
\begin{equation}
\epsilon^* = \frac{\Delta_2^2 (g)}{2 \sigma^2 } + \frac{\Delta_2 (g) \sqrt{2 \log (1/\delta)}}{\sigma} .
\end{equation}
\begin{comment}
 Note, that for large noise $\sigma^2$, this simplifies to:
\begin{equation}
\epsilon \approx \frac{\Delta_2 (g)}{\sigma} \sqrt{2 \log(1/\delta)}.
\end{equation}
\end{comment}
Now, replacing $g$ with $\sqrt{w} \hat{f}^{-1}$ and $\sigma^2 = 1 - w,$ we get $\tilde{\mathcal{M}}(x)$ satisfies Local $(\epsilon^{*}, \delta)$ Differential Privacy for any $\delta \in (0, 1)$ with $$\epsilon^* = \frac{w (C' + 2 \eta_n)^2}{2(1-w)} + \frac{(C' + 2 \eta_n) \sqrt{2w \log(1/\delta)}}{\sqrt{1-w}},$$
since 
$$\frac{w \Delta_2^2 (\hat{f}^{-1})}{2 ( 1 - w)} + \frac{ \Delta_2(\hat{f}^{-1}) \sqrt{2 w \log(1/\delta)}}{\sqrt{1 - w}} \leq \frac{w (C' + 2 \eta_n)^2}{2(1-w)} + \frac{(C' + 2 \eta_n) \sqrt{2w \log(1/\delta)}}{\sqrt{1-w}}.$$
As $n \rightarrow \infty$, $\tilde{\mathcal{M}}(x)$ satisfies Local $(\epsilon^*, \delta)$ Local DP, replacing $C' + 2\eta_n$ with $C'$.

Under this condition, by the Post Processing Theorem, which states that if $\tilde{\mathcal{M}}(x)$ is Locally $(\epsilon^{*}, \delta)$-Differentially Private, any function applied to the output also preserves the same privacy guarantees \citep{dwork2014algorithmic}, we conclude that by applying the MAF back to the observed space, the mechanism $\hat{f}(\tilde{\mathcal{M}}(x))$ is also Locally $(\epsilon^{*}, \delta)-$Differentially Private with $\Delta_2(\hat{f}^{-1}) \leq C' + 2\eta_n$. Then, from Appendix \ref{app:TransportMap}, since $\hat{f}$ converges to $f$ in $L_2,$ it also converges in probability as $n \rightarrow \infty.$ Then, from (\ref{eq:Mechanism_Privacy}), sending $n \rightarrow \infty$ the mechanism $\hat{f}(\tilde{\mathcal{M}}(x))$ satisfies Local $(\epsilon^{*}, \delta)$ DP for any $0 < \delta < 1$ with
\[
%    \epsilon^* = \frac{w \Delta_2^2(f^{-1}}{2(1-w)} + \frac{\Delta_2(f^{-1}) \sqrt{2w \log(1/\delta)}}{\sqrt{1-w}}.
 \epsilon^* = \frac{w C'^2}{2(1-w)} + \frac{C' \sqrt{2w \log(1/\delta)}}{\sqrt{1-w}}.
    \]
This completes the proof.
\end{proof}

Without using Local RDP, we may directly prove the following theorem on the Local DP of our proposed mechanism. 
\begin{theorem} \label{thm:Local_DP2}
First, consider a trained MAF transport map, $\hat{f} = \hat{f}_k \circ \cdots \circ \hat{f}_1,$  where each $\hat{f}_j, j \in \{1, \dots, k\}$ has mapping defined in (\ref{eq:Forward}). We assume the following:
\begin{enumerate}
    \item each function $\mu_i(\cdot)$ and $\sigma_i(\cdot)$ in (\ref{eq:Forward}) are trained neural networks whose layers are spectrally normalized with the same number of layers $L$ for each $\hat{f}_j, j \in \{ 1, \dots, k \}$. That is, each weight matrix has spectral norm at most $1$.
    
    \item the data domain $\mathcal{X} \subset \mathbb{R}^d$ is compact.
    \item  $\hat{f}$ converges to a a deterministic transport map $f$ uniformly in probability as the sample size increases and the sensitivity of $f^{-1}$ is bounded, i.e.,  $\Delta_2(f^{-1}) \leq C$ for a constant $C$.
\end{enumerate}
Let 
\[
\hat{f}(\tilde{\mathcal{M}}(x))= \hat{f}\left\{\sqrt{w} \hat{f}^{-1}(x) + \sqrt{1-w} Z\right\}, \quad Z \sim N(0, \mathbf{I}_d).
\]
Then, with $\hat{\Delta} = \|\hat{f}^{-1}(x) - \hat{f}^{-1}(y)\|_2$ and $\Delta = \|f^{-1}(x) - f^{-1}(y)\|_2$, the mechanism $\hat{f} \left\{ \tilde{\mathcal{M}}(\cdot) \right\}$ satisfies $(\epsilon, \delta)$-Local DP Privacy in probability for any $0 < \delta < \frac{1}{2}$ if for sufficiently small $w$ 
\begin{equation} \label{eq:Priv_Condition}
w < \frac{1}{1 + \frac{\Delta^2}{\epsilon^2}\left(\epsilon -  2\log(2\delta) \right)},
\end{equation}
\begin{comment}
 $$w < \left[ 1+\frac{2 \Delta_2(f^{-1})^2}{\epsilon^2}\log\left(\frac{1.25}{\delta}\right)\right]^{-1}$$    
\end{comment}
i.e., 
$$ \mathbb{P}\left(\mathbb{P}[\hat{f} \left\{ \tilde{\mathcal{M}}(x)\right\} = y] \leq e^{\epsilon^*} \mathbb{P}[\hat{f} \left\{\tilde{\mathcal{M}}(x') \right\} =y ]+\delta\right) \rightarrow 1$$
as $n \rightarrow \infty$ for any $x, x'$ and $y$ under (\ref{eq:Priv_Condition}) with sufficiently small $w$.
\begin{comment}
 $$w < \left[ 1+\frac{2 \Delta_2(f^{-1})^2}{\epsilon^2}\log\left(\frac{1.25}{\delta}\right)\right]^{-1}$$    
\end{comment}

\end{theorem}

\begin{proof} First, suppose we are in $\mathbb{R}$. Now, consider the input space $\tilde{\mathcal{M}}(x)$: $\sqrt{w} \hat{f}^{-1}(x)+\sqrt{1-w} Z$, where $Z \sim N(0, 1).$ Note that 
$\sqrt{w} \hat{f}^{-1}(x)+\sqrt{1-w} Z \sim N(\sqrt{w} \hat{f}^{-1}(x),  1-w)$ and $\hat{f}^{-1}(x)-\hat{f}^{-1}(x')=\hat{\Delta},$ where $x, x' \in \mathcal{X}$, and $\hat{\Delta} \leq C' + 2 \eta_n$ similar to the setup for the proof of Theorem \ref{thm:Local_DP} with some constant $C' > 0$. We first need to show that
$$\mathbb{P}(\log\frac{\mathbb{P}[\tilde{\mathcal{M}}(x) = y]}{\mathbb{P}[\tilde{\mathcal{M}}(x') =y ]} > \epsilon) \leq \delta$$
for any $0 < \epsilon, \delta < 1.$ Note that
\begin{align*}
&\log\left(\frac{\mathbb{P}[\tilde{\mathcal{M}}(x) = y]}{\mathbb{P}[\tilde{\mathcal{M}}(x') =y ]} \right) \\
=& \frac{-(y-\sqrt{w} \hat{f}^{-1}(x))^2+(y-\sqrt{w}\hat{f}^{-1}(x'))^2}{2(1-w)}\\
=& \hat{\Delta} \frac{2\sqrt{w} (\sqrt{w} \hat{f}^{-1}(x)+\sqrt{1-w}Z) -w(\hat{f}^{-1}(x)+\hat{f}^{-1}(x'))}{2(1-w)}\\
=& \hat{\Delta} \frac{2\sqrt{w(1-w)}Z +w\hat{\Delta}}{2 (1-w)}.
\end{align*}
Then, we would like the following:
\begin{align*}
& \mathbb{P}\left(\hat{\Delta} \frac{2\sqrt{w(1-w)}Z +w\hat{\Delta}}{2(1-w)}>\epsilon \right)\\
=& \mathbb{P}\left(Z>\frac{\epsilon \sqrt{1-w}}{\sqrt{w}\hat{\Delta}}-\frac{\sqrt{w}\hat{\Delta}}{2\sqrt{1-w}} \right)\\
\le & \frac{1}{2}\exp\left\{-\frac{1}{2}\left[\frac{\epsilon \sqrt{1-w}}{\sqrt{w}\hat{\Delta}}-\frac{\sqrt{w}\hat{\Delta}}{2\sqrt{1-w}}\right]^2 \right\}<\delta,
\end{align*}
if
\[
w < \frac{1}{1 + \frac{\hat{\Delta}^2}{\epsilon^2}\left(\epsilon -  2\log(2\delta) \right)}
\]
\begin{comment}
$$w < \left[ 1+\frac{2\Delta^2}{\epsilon^2}\log\left(\frac{1.25}{\delta}\right)\right]^{-1}$$
for sufficiently small $w$ and any $0 < \epsilon < 1$ and \(0 < \delta < 1/2\).
\end{comment}
To see the sufficient condition, we are given the quantity:
\[
E(w) = \frac{\epsilon^2(1-w)}{w\hat{\Delta}^2} - \epsilon + \frac{w\hat{\Delta}^2}{4(1-w)},
\]
and we want to find conditions on \(w\) such that
\[
\frac{1}{2} \exp\left\{ -\frac{1}{2} \left( \frac{ \epsilon \sqrt{1-w} }{\sqrt{w}\hat{\Delta}} - \frac{ \sqrt{w}\hat{\Delta} }{2\sqrt{1-w}} \right)^2 \right\} < \delta.
\]
Multiplying both sides by 2 and taking the natural logarithm, we get:
\[
-\frac{1}{2}E(w) < \log(2\delta),
\]
\[
E(w) > -2\log(2\delta).
\]
Substituting the expression for \(E(w)\) yields:
\[
\frac{\epsilon^2(1-w)}{w\hat{\Delta}^2} - \epsilon + \frac{w\hat{\Delta}^2}{4(1-w)} > -2\log(2\delta).
\]
Thus,
\[
\frac{\epsilon^2(1-w)}{w\hat{\Delta}^2} + \frac{w\hat{\Delta}^2}{4(1-w)} > \epsilon - 2\log(2\delta).
\]
Note, the left-hand side consists of two positive terms. Thus, a sufficient condition is obtained by focusing on the dominant term \(\frac{\epsilon^2(1-w)}{w\hat{\Delta}^2}\), and requiring:
\[
\frac{\epsilon^2(1-w)}{w\hat{\Delta}^2} > \epsilon - 2\log(2\delta).
\]

\noindent
Dividing both sides by \(\epsilon^2/\hat{\Delta}^2\) and rearranging gives:
\[
\frac{1-w}{w} > \frac{\hat{\Delta}^2}{\epsilon^2}\left(\epsilon - 2\log(2\delta) \right).
\]
Thus,
\[
w < \frac{1}{1 + \frac{\hat{\Delta}^2}{\epsilon^2}\left(\epsilon -  2\log(2\delta) \right)}.
\]
\footnote{Since \(0<\delta<1/2\), \(\log(2\delta)<0\) and \(-\log(2\delta)>0\), making the right-hand side well-defined and positive. Similarly, we may also explicitly derive a solution isolating $w$ by solving a quadratic formula without any assumptions of $w \rightarrow 0$, only that $w \in [0, 1].$} Now, let us partition $R$ as $R = R_1 \cup R_2$, where
\[
R_1 = \left\{ Z \in R :\hat{\Delta} \frac{2\sqrt{w(1-w)}Z +w\hat{\Delta}}{2(1-w)} \leq\epsilon  \right\},
\quad
R_2 = \left\{ Z \in R : \hat{\Delta} \frac{2\sqrt{w(1-w)}Z +w\hat{\Delta}}{2(1-w)}>\epsilon \right\}.
\]
Now, fix any subset $S \subseteq R$, and set
\[
S_1 = \{ y \in S : \mathcal{M}(x) = y, \, \xi \in R_1 \},
\quad
S_2 = \{ y \in S: \mathcal{M}(x) = y, \, \xi \in R_2 \},
\]
where $\xi \sim N(0,1)$ is a noise realization in the mechanism. We have
\begin{align*}
\mathbb{P}[\tilde{\mathcal{M}}(x) \in S]
&= \mathbb{P}[\tilde{\mathcal{M}}(x) \in S_1] + \mathbb{P}[\tilde{\mathcal{M}}(x) \in S_2] \\
&\leq \mathbb{P}[\tilde{\mathcal{M}}(x) \in S_1] + \delta \\
&\leq e^{\varepsilon} \mathbb{P}[\tilde{\mathcal{M}}(x') \in S_1] + \delta \\
&\leq e^{\varepsilon} \mathbb{P}[\tilde{\mathcal{M}}(x') \in S] + \delta,
\end{align*}
where the second inequality follows because the probability mass over $R_2$ is at most $\delta$, and the third inequality follows from bounding the privacy loss over $R_1$ by $\varepsilon$.

Extending this to the multivariate setting, suppose $\hat{f}^{-1}(x), \hat{f}^{-1}(x') \in \mathbb{R}^d$ are neighboring points. Define $\hat{\Delta}_{\text{Vec}} = \hat{f}^{-1}(x) - \hat{f}^{-1}(x')$ and $\hat{\Delta} = \|\hat{f}^{-1}(x) - \hat{f}^{-1}(x')\|$. $\tilde{\mathcal{M}}(x)$ is $\sqrt{w}\hat{f}^{-1}(x) + \sqrt{1-w}Z$, where $Z \sim N(0, \mathbf{I}_d)$.
Thus,
\[
\tilde{\mathcal{M}}(x) \sim N(\sqrt{w} \hat{f}^{-1}(x), (1-w)\mathbf{I}_d), \quad \tilde{\mathcal{M}}(x') \sim N(\sqrt{w} \hat{f}^{-1}(x'), (1-w)\mathbf{I}_d).
\]
Note, the privacy loss at output $y$ is:
\[
L(y) = \log\left( \frac{p_{\tilde{\mathcal{M}}(x)}(y)}{p_{\tilde{\mathcal{M}}(x')}(y)} \right) = \frac{1}{2(1-w)} \left( \|y - \sqrt{w} \hat{f}^{-1}(x')\|^2 - \|y - \sqrt{w} \hat{f}^{-1}(x)\|^2 \right).
\]
Applying $\|a\|^2 - \|b\|^2 = (a-b)^\top (a+b),$
\[
\|y - \sqrt{w}\hat{f}^{-1}(x')\|^2 - \|y - \sqrt{w} \hat{f}^{-1}(x)\|^2 = (\sqrt{w}\hat{\Delta}_{\text{Vec}})^\top (2y - \sqrt{w}(\hat{f}^{-1}(x)+\hat{f}^{-1}(x'))).
\]
Substituting $y = \sqrt{w}\hat{f}^{-1}(x) + \sqrt{1-w} Z$,
\begin{align*}
2y - \sqrt{w}(\hat{f}^{-1}(x)+\hat{f}^{-1}(x')) &= 2(\sqrt{w}\hat{f}^{-1}(x) + \sqrt{1-w}Z) - \sqrt{w}(\hat{f}^{-1}(x)+\hat{f}^{-1}(x')) \\
&= \sqrt{w}(\hat{f}^{-1}(x) - \hat{f}^{-1}(x')) + 2\sqrt{1-w}Z \\
&= \sqrt{w}\hat{\Delta}_{\text{Vec}} + 2\sqrt{1-w} Z.
\end{align*}
Thus,
\[
\sqrt{w}\hat{\Delta}_{\text{Vec}}^\top (2y - \sqrt{w}(\hat{f}^{-1}(x)+\hat{f}^{-1}(x'))) = w\|\hat{\Delta}_{\text{Vec}}\|^2 + 2\sqrt{w(1-w)}\hat{\Delta}_{\text{Vec}}^\top Z
\]
with
\[
L(y) = \frac{1}{2(1-w)} \left( w\|\hat{\Delta}_{\text{Vec}}\|^2 + 2\sqrt{w(1-w)}\hat{\Delta}_{\text{Vec}}^\top Z \right).
\]
Now,
\[
Z' = \frac{\hat{\Delta}_{\text{Vec}}^\top Z}{\|\hat{\Delta}_{\text{Vec}}\|}, \quad \text{thus} \quad Z' \sim N(0,1).
\]
Hence,
\[
\hat{\Delta}_{\text{Vec}}^\top Z = \hat{\Delta} Z'.
\]
Substituting,
\[
L(y) = \frac{w\hat{\Delta}^2}{2(1-w)} + \frac{2\sqrt{w(1-w)}\hat{\Delta} Z'}{2(1-w)}.
\]
Simplifying,
\[
L(y) = \hat{\Delta} \times \frac{2\sqrt{w(1-w)}Z' + w\hat{\Delta}}{2(1-w)}.
\]
Thus, the final privacy loss for this multivariate Gaussian Mechanism is
\[
L(y) = \hat{\Delta} \times \frac{2\sqrt{w(1-w)}Z' + w\hat{\Delta}}{2(1-w)}, \quad Z' \sim N(0,1).
\]
So, the same proof goes through in the $\mathbb{R}^d$ case with
\[
w < \frac{1}{1 + \frac{\hat{\Delta}^2}{\epsilon^2}\left(\epsilon -  2\log(2\delta) \right)}
\]
\begin{comment}
 $$w < \left[ 1+\frac{2 \Delta_2(f^{-1})^2}{\epsilon^2}\log\left(\frac{1.25}{\delta}\right)\right]^{-1}$$    
\end{comment}
for sufficiently small $w$, but with $\hat{\Delta}$ in the Multivariate Setting being with respect to $L_2$ Norm. Repeating our argument above, by the Post Processing Theorem \citep{dwork2014algorithmic}, we conclude that by applying the MAF back to the observed space, the mechanism $\hat{f}(\tilde{\mathcal{M}}(X))$ is also Locally $(\epsilon^{*}, \delta)-$Differentially Private with high probability when replacing $\hat{\Delta}$ with $\Delta$ as $n \rightarrow \infty$.

Repeating our argument above again, from Appendix \ref{app:TransportMap}, since $\hat{f}$ converges to $f$ in $L_2,$ it also converges in probability as $n \rightarrow \infty.$ Then, from (\ref{eq:Mechanism_Privacy}), the mechanism $\hat{f}(\tilde{\mathcal{M}}(x))$ satisfies Local $(\epsilon^{*}, \delta)$ DP with high probability as $n \rightarrow \infty$ if
\[
w < \frac{1}{1 + \frac{\Delta^2}{\epsilon^2}\left(\epsilon -  2\log(2\delta) \right)}
\]
with sufficiently small $w$ for any $0 < \epsilon < 1$ and \(0 < \delta < 1/2\) .

\end{proof}

\begin{remark}
In Theorems \ref{thm:Local_DP} and \ref{thm:Local_DP2}, the sensitivity $\Delta_2(f^{-1})$ is assumed to be bounded from above. This is a consequence of Lemma \ref{lem:Lipschitz_Spectral} and Lemma \ref{lem:Bounded_Lat_Space}.
\end{remark}

\vspace{.05 in}

\end{document}